%% file: main.tex
\PassOptionsToPackage{dvipsnames}{xcolor}

\documentclass[10pt]{article} 
\usepackage[preprint]{tmlr}

\input{math_commands.tex}

\usepackage{hyperref}
\usepackage{url}

\usepackage{booktabs}
\usepackage{colortbl}
\usepackage{adjustbox}
\usepackage{subcaption}
\usepackage{multirow}
\usepackage{wrapfig}
\usepackage{algorithm}
\usepackage{algpseudocode}
\usepackage{dsfont} 

\title{Revisiting Active Learning in the Era of \\Vision Foundation Models}

\author{\name Sanket Rajan Gupte\thanks{Equal contribution.} \email sanketg@stanford.edu \\
      \addr Department of Computer Science\\
      Stanford University
      \AND
      \name Josiah Aklilu\footnotemark[1] \email josaklil@stanford.edu \\
      \addr Department of Biomedical Data Science\\
      Stanford University
      \AND
      \name Jeffrey J. Nirschl \email jnirschl@stanford.edu\\
      \addr Department of Pathology\\
      Stanford University
      \AND
      \name Serena Yeung-Levy \email syyeung@stanford.edu\\
      \addr Department of Biomedical Data Science\\
      Stanford University\\
      }

\def\month{06}  
\def\year{2024} 
\def\openreview{\url{https://openreview.net/forum?id=u8K83M9mbG}} 

\newcommand{\uncertainty}{{\textbf{Uncertainty}}\:}
\newcommand{\entropy}{{\textbf{Entropy}}\:}
\newcommand{\margins}{{\textbf{Margins}}\:}
\newcommand{\coreset}{{\textbf{Coreset}}\:}
\newcommand{\bald}{{\textbf{BALD}}\:}
\newcommand{\powerbald}{{\textbf{PowerBALD}}\:}
\newcommand{\probcover}{{\textbf{ProbCover}}\:}
\newcommand{\typiclust}{{\textbf{Typiclust}}\:}
\newcommand{\badge}{{\textbf{BADGE}}\:}
\newcommand{\alfamix}{{\textbf{Alfa-Mix}}\:}
\newcommand{\dropout}{\textcolor{LimeGreen}{\textbf{DropQuery}}\:}

\begin{document}

\newcolumntype{A}{>{\columncolor{Tan!50}}c}
\newcolumntype{B}{>{\columncolor{CornflowerBlue!25}}c}
\newcolumntype{D}{>{\columncolor{LimeGreen!50}}c}
\renewcommand{\algorithmicrequire}{\textbf{Input:}}
\renewcommand{\algorithmicensure}{\textbf{Output:}}

\maketitle

\begin{abstract}
Foundation vision or vision-language models are trained on large unlabeled or noisy data and learn robust representations that can achieve impressive zero- or few-shot performance on diverse tasks. Given these properties, they are a natural fit for \textit{active learning} (AL), which aims to maximize labeling efficiency. However, the full potential of foundation models has not been explored in the context of AL, specifically in the low-budget regime. In this work, we evaluate how foundation models influence three critical components of effective AL, namely, 1) initial labeled pool selection, 2) ensuring diverse sampling, and 3) the trade-off between representative and uncertainty sampling. We systematically study how the robust representations of foundation models (DINOv2, OpenCLIP) challenge existing findings in active learning. Our observations inform the principled construction of a new simple and elegant AL strategy that balances uncertainty estimated via dropout with sample diversity. We extensively test our strategy on many challenging image classification benchmarks, including natural images as well as out-of-domain biomedical images that are relatively understudied in the AL literature. We also provide a highly performant and efficient implementation of modern AL strategies (including our method) at \url{https://github.com/sanketx/AL-foundation-models}. 
\end{abstract}

\section{Introduction}

Foundation models \citep{dinov2, openclip, foundation} have become ubiquitous in computer vision. The advent of large vision and vision-language models, pretrained on web-scale data corpora, has led to the significant enhancement of visual representation spaces. Despite differences in pretraining (e.g. contrastive frameworks with language supervision as in \cite{openclip} or vision-only instance discrimination objectives as in \cite{dinov2}), the shared theme among vision foundation models are robust learned representations that can be effectively leveraged for state-of-the-art performance on multiple downstream tasks. These learned visual features have direct implications for machine learning paradigms designed to mitigate data paucity challenges, most notably deep active learning. 

Active learning (AL) \citep{coreset, badge, bald, powerbald, alges, typiclust, probcover, alfamix, wassersteinal} is a machine learning framework that addresses the scarcity of labeled data within a limited label budget. It achieves this by iteratively requesting labels from an external oracle, such as a subject matter expert or clinician. Stated differently, the challenge is determining the most beneficial labels to query when constrained by a limited labeling budget to maximize model performance. This approach is especially pertinent in practical computer vision applications that necessitate the creation of domain-specific datasets for custom tasks. A well-designed AL strategy can optimize resource allocation and substantially lower acquisition costs in situations where labeled data are scarce and labor-intensive to produce — such as annotating high-resolution histopathology images for tumor classification.

Previous research has tested the benefit of using pretrained representations during AL by leveraging representations of unlabeled data to 1) address the cold-start problem by selecting a good initial pool of candidates for labeling \cite{typiclust, makefirstchoice, oninitpools}, 2) improve query functions by selecting points that maximize the coverage of the representation space \cite{probcover, coreset} and, 3) incorporating unlabeled samples in the training process via semi-supervised learning \cite{consistency, rethinking}. However, these studies limit their focus to supervised pretraining on ImageNet or self-supervised pretraining on available unlabeled data, which often proves impractical in many real-world scenarios. For instance, biomedical imaging datasets exhibit distribution shifts from natural images and may not have sufficient samples for effective self-supervised representation learning.
 
Foundation models offer a compelling solution to these challenges. Image embeddings extracted from these models are semantically organized in the representation space, as evidenced by impressive zero- and few-shot performance with simple linear models, even on fine-grained classification datasets. Evidence also suggests they are robust to domain shifts and generalize well to out-of-distribution data. Since these embeddings "just work" out of the box, building dataset-specific feature extractors is no longer necessary. For these reasons, we envision that future research in AL will build upon the ever-increasing capabilities \cite{scalingto22, evaclip} of Vision and Vision-Language foundation models to formulate efficient and scalable acquisition strategies. However, there is limited research \cite{plex} on the impact of foundation models on AL. Our work seeks to re-examine the pillars of effective AL strategies in this context, specifically in the low-budget regime in which only a handful of images per class can be annotated. We hope our study's compelling results and novel insights will spur the use of foundation models as a "batteries included" launchpad for future AL research.

\textbf{Our key contributions are the following:} 

\begin{enumerate}
    \item We investigate the impact of large-scale vision foundation models on the four pillars of active learning strategies, specifically in the low-budget regime, and contrast our findings with established results. Our analysis systematically explores the following dimensions of a successful AL strategy:
    \begin{itemize}
        \item We study the impact of initial pool selection and highlight differences with existing research on the cold-start problem.
        \item We explore the importance of querying diverse samples and demonstrate that with a few simple modifications, poorly performing AL query strategies can match those that explicitly incorporate sample diversity.
        \item We compare representative sampling with uncertainty-based sampling and counter the existing notion that uncertainty-based query methods are ineffective in the low-budget regime.
        \item We demonstrate the difficulties of leveraging semi-supervised learning as a complementary approach for building accurate models with limited labeled data.
    \end{itemize}
    \item As a direct consequence of the results of our investigation, we construct a simple, performant, and scalable AL strategy, \dropout. This method, built atop the rich semantic features generated by powerful foundation models, leverages an intelligent strategy for initial pool selection, utilizes an uncertainty-based criterion to generate a pool of query candidates, and selects a diverse subset of candidates for annotation. Our strategy outperforms the current state-of-the-art on a diverse collection of datasets, including fine-grained image classification, out-of-distribution biomedical images, and image classification at scale, all of which are relatively understudied in AL.
\end{enumerate}

\section{Background and Related Works}
Traditional research in AL for image classification has largely focused on the development of acquisition functions to query samples for labeling by an oracle. These strategies have spanned uncertainty-based sampling approaches \citep{shannon}, Bayesian methods \citep{bald, powerbald}, strategies to maximize sample diversity, query by committee strategies \citep{diverseensembles}, and techniques to estimate the largest update to the model parameters \citep{badge}.

Uncertainty-based methods supply a single measure of uncertainty (epistemic or aleatoric) for unlabeled samples so as to rank them for acquisition. \entropy selects instances with maximum predictive entropy, \uncertainty selects instances with the lowest confidence, and \margins selects instances with the lowest margin between the highest and second-highest predicted class probabilities. The Bayesian approaches to AL \citep{bald, batchbald, balentacq} leverage Bayesian neural networks to model uncertainty and the potential informativeness of unlabeled instances. Bayesian Active Learning by Disagreement or \bald \citep{bald} aims to query instances that maximize the mutual information gain between model parameters and predictive outputs. Additionally, the \powerbald \citep{powerbald} query challenges the traditional top-$B$ acquisition approach by noting the correlation of queried instances \textit{between} each iteration of AL, which is alleviated by introducing stochasticity in the query. 

The prior works that have placed an emphasis on diversity sampling argue that the classifier should learn good decision boundaries in the representation space early. In the lens of these diversity-based queries, the AL paradigm can be re-framed as \textit{"what are the diverse instances that are representative of the data distribution?"}. The \coreset algorithm \citep{coreset} can be approximated as solving the \textit{k}-Center greedy problem \citep{kcentergreedy}, where the distance between any instance and its nearest cluster center (the labeled instances from previous AL iterations) is minimized. Following the Coreset query, the recently developed \probcover query aims to avoid the selection of outlier instances and query more representative points. This is done by recasting the AL problem as a max coverage problem \citep{maxcoverage}. 

A hybrid technique, \badge \citep{badge}, uses an AL query that inspects the gradients of a neural network's final layer relative to the current model's prediction. This identifies unlabeled instances that could induce significant model updates. The underlying intuition is that samples prompting large gradient updates require substantial adjustments, signifying uncertainty in samples with new features. While this method yields diverse and highly uncertain samples, it is computationally intensive due to the storage of penultimate activations. \alfamix \citep{alfamix} solicits unlabeled samples with the greatest classification prediction variability when their representations are linearly interpolated with labeled instances, suggesting that the model needs to learn new features from unlabeled data points. \typiclust \citep{typiclust}, queries typical instances during the early AL stages leveraging the representation space. Clustering in a semantically meaningful feature space after self-supervised learning ensures maximum diversity sampling. Sampling dense regions within clusters aids in better classifier learning within feature space.

However, in most of the settings considered by these works, a model is trained from scratch at each iteration using a random initialization or an ImageNet pretrained backbone. The resulting models are often poorly calibrated \citep{oncalibration} and fail to provide robust estimates of uncertainty. These settings also require a relatively large labeling budget as training deep non-linear models with limited labels can be challenging \citep{nlpcoldstart}. Consequently, a number of such strategies have been shown to underperform random sampling in the low budget \citep{rethinking, typiclust}, or do not perform significantly better than random when strong regularization is applied \citep{towardsrobust}.

Recent methods designed for the low budget regime \citep{typiclust, probcover} take a more holistic approach to developing an effective AL strategy by leveraging pretrained model embeddings to cluster features and select representative points, which has been shown to improve the selection of the initial labeled pool. Other approaches aim to exploit unlabeled points using self-supervised or semi-supervised learning, or a combination of both \citep{onthemarginal, selfsupsemisup}. Their findings demonstrate that leveraging rich representations significantly improves AL performance in the context of a small labeling budget \citep{luth2023}.

Given these promising results, it is reasonable to conjecture that embeddings from large-scale vision transformers trained on billions of images would amplify the efficacy of these complementary components, constructing an effective strategy that surpasses a simpler baseline consisting of only a standalone acquisition function. To comprehensively investigate these facets of AL, we compare the impact of a simple pool initialization strategy that leverages the representation space's implicit structure to sample diverse and representative points. Further, we investigate different classes of query functions in this context to compare uncertainty-based approaches with typicality-based methods, as the latter have been shown to significantly outperform the former in the low-budget regime. Orthogonal to the query function, we investigate if semi-supervision via label propagation can be an effective aid in improving the performance of an active learner.

Although some recent work has explored the effects of pretraining on the AL query \citep{typiclust, alpretrain}, there has not been a rigorous evaluation of AL queries with respect to developments in foundation models and vision-language pretraining that have yielded highly expressive and rich visual representation spaces. In the following sections, we outline the critical components necessary for an effective AL query in the context of large foundation models.

\section{Investigating the pillars of effective active learning strategies}

Formally, we consider the batch AL setting: let $\mathcal{D} = \{\mathcal{D}_{\textbf{\textsubscript{pool}}} \cup \mathcal{D}_{\textbf{\textsubscript{test}}}\}$ be a dataset with a held-out test set and training pool $\mathcal{D}_{\textbf{\textsubscript{pool}}} = \{\mathcal{D}_U \cup \mathcal{D}_L \}$ consisting of unlabeled instances $\{\bm{x}_i\}$ for $i \in \{N_U\}$ and labeled instances $\{(\bm{x}_j, y_j)\}$ for $j \in \{N_L\}$. Let $f : \mathbb{R}^{H \times W \times 3} \rightarrow \mathbb{R}^d$ be a feature extractor such that $\bm{z}_i := f(\bm{x}_i) \in \mathbb{R}^d$ is a feature vector and $d$ is the embedding dimension. Let $g(\bm{z}; \bm{\theta}) : \mathbb{R}^{d} \rightarrow \mathbb{R}^K $ be a classifier parameterized by $\bm{\theta}$. The typical AL procedure consists of a series of sequential iterations beginning at $t = 0$ where the labeled pool $\mathcal{D}_L$ is the empty set. At each iteration, a querying function $a(\{\bm{z}_i : i \in \{N_U\}\}; \bm{\theta}_t)$, which given the current model parameters $\bm{\theta}_t$, receives as input all unlabeled instances, selects $B$ instances to be labeled by an external oracle (e.g. clinical expert). Once labels are acquired, these labeled instances are subsumed in $\mathcal{D}_L$ and removed from the unlabeled pool, and the model is retrained on this slightly larger labeled pool. We simulate the AL procedure by hiding labels $y_i$ until queried for a chosen instance $\bm{x}_i$.

In this work, we employ frozen vision-only or vision-language foundation models as the feature extractor $f(.)$ and a linear head as the classifier $g(.)$. We only require a single forward pass through the feature extractor to generate embeddings which are saved for use in our various experiments. This enables us to study a wide range of experimental settings for AL efficiently. Unless mentioned otherwise, we use a DINOv2 VIT-g14 \citep{dinov2} transformer as the feature extractor. Simple linear models trained on features extracted by this model achieve performance that is just shy of state-of-the-art, making them a compelling alternative to end-to-end fine-tuning of the backbone with scarce labels. Following the recommendations of \citep{towardsrobust}, we apply strong regularization to our models in the form of weight decay (1e-2) and aggressive dropout (0.75). Note that the benchmarks used to evaluate the various AL methods in this work were not used in the pretraining of the foundation models we investigated (we refer the reader to the data curation processes outlined by \cite{dinov2}).

In our experiments, we set a query budget $B$ of 1 sample per class per iteration. For instance, the CIFAR100 dataset contains images corresponding to 100 classes, so the query budget is set to 100 samples. Note that setting $B$ to 1 sample per class per iteration does not necessarily imply that exactly one sample is chosen from each class, since the class labels are unknown at the time of querying. We run our active learning loop for 20 iterations and average our results over 5 random seeds for each experiment. We analyze the following query functions: Random Sampling baseline, Entropy, Uncertainty, Margins, Core-Set (greedy k-centers), BALD, Power BALD, BADGE, ALFA-Mix, Typiclust, and ProbCover. The hyper-parameters for these query functions, if any, are taken from the original publication.

\subsection{The impact of initial pool selection}\label{init_pool}
We begin our study by measuring the impact of different initial pool selection strategies in AL, particularly in the \textit{cold-start} scenario where no data has been used for training the classifier. It has been shown in the recent literature \cite{typiclust,probcover,makefirstchoice} that the initial pool is crucial for establishing a good rudimentary classifier that enjoys performance gains throughout AL. \cite{typiclust} and \cite{probcover} also argue that the AL acquisition function need not be decoupled from the initial pool selection. Rather, an AL query should acquire representative samples so the classifier can learn good decision boundaries from the first query. Intuitively, in order for an AL query to be effective, it must rely on the current classifier's estimates of uncertainty or informativeness. 

To assess initialization strategies, we contrast AL query performance when the initial training pool for the active learner is randomly selected for labeling to a centroid-seeking approach following \cite{simplebaselinelowbudgetal}, where the initial pool consists of samples from $\mathcal{D}_U$ that are the closest to class centers after \textit{K}-means clustering in the feature space, where $B$ is the number of clusters. \typiclust and \probcover utilize different initialization strategies based on sample density and maximizing coverage, respectively, so we hold these queries true to their initialization approach. Table \ref{tab:delta_init_pool} compares the deltas ($\Delta$) in the test set performance of a linear classifier actively trained on foundation model features using a randomly initialized initial pool versus a centroid-based initial pool. Note that a random initial pool is suboptimal to methods that explicitly take advantage of semantically meaningful embedding spaces (i.e. \typiclust or \probcover). The tremendous performance gains in early AL iterations for uncertainty-based AL queries like \entropy sampling enable these methods to surpass queries like \typiclust and \probcover.

Table \ref{tab:delta_init_pool} also demonstrates the impact of initial pool selection on AL queries acting on foundation model features. Some queries, like \uncertainty, \entropy, and \coreset, enjoy performance gains throughout AL given a centroid-based initialization, a major boost to the random initialization. Interestingly, some queries like \alfamix and \badge have deltas that converge within 2-4 iterations, and in later iterations (8-16) we observe higher accuracy with a randomly selected initial pool. Since the \alfamix query crucially relies on interpolation in the feature space, we hypothesize that the separability of foundation model representation space renders differences between initial pooling strategies negligible. After a few iterations, the robust visual representations from foundation models enable the selection of highly representative samples, which help the classifier establish good class boundaries in few-shot.

The results from this experiment are in contrast to previous preliminary findings from \cite{oninitpools} that report no significant difference in initial pool selection strategies in the long run. While we also see diminishing returns as we progress to later iterations, we emphasize that the experimental results in \cite{oninitpools} also showed little to no difference caused by most initialization strategies, even in the very first iteration of AL. This observation does not align with our results in which we see stark differences caused by intelligent selection of the initial labeled pool in the beginning iterations.

However, this discrepancy can be resolved when taking into account the sizes of the labeling budgets. Our experimental setting is the very low-budget regime, and the number of labeled samples in the final iterations of our studies barely overlaps with the starting budgets studied in \cite{oninitpools}. We conclude that in the low-budget regime, where only a few examples per class have been acquired, initialization is crucial for AL performance, but in later iterations, as the number of labeled samples grows or as we transition to higher-budget regimes, it is not as significant and a randomly selected initial pool works just as well.

\begin{table}[!t]
    \centering
    \caption{Effect of initial pool selection on performance. Test set accuracy using our centroid-based initialization. In parentheses, we show the difference ($\Delta$) in performance when utilizing our centroid initialization vs. random initialization, where a positive ($\Delta$) shown in green indicates improvement over random initialization. We show AL iterations $t$ for datasets CIFAR100 \citep{cifar100}, Food101 \citep{food101}, ImageNet-100 \citep{imagenet100}, and DomainNet-Real \citep{domainnetreal} (from top to bottom) with DINOv2 ViT-g14 as the feature extractor $f$. For both the Typiclust and ProbCover queries, we report the test accuracy using their own respective initialization strategies for the cold start.}
    \vspace{0.5em}
    \begin{adjustbox}{width=1\textwidth}
    \begin{tabular}{c | c c c c c c c c c A B}
    \toprule 
        \rowcolor{gray!10} $t$ & Random & Uncertainty & Entropy & Margins & BALD & pBALD & Coreset & BADGE & Alfa-mix & Typiclust & ProbCover \\
    \midrule
    \multicolumn{12}{c}{CIFAR100} \\
    \midrule
        1 & 
        \cellcolor{LimeGreen!90} 72.4 (+24.4) & 
        \cellcolor{LimeGreen!90} 72.4 (+24.4) & 
        \cellcolor{LimeGreen!90} 72.4 (+24.4) & 
        \cellcolor{LimeGreen!90} 72.4 (+24.4) &  
        \cellcolor{LimeGreen!90} 72.4 (+24.4) &  
        \cellcolor{LimeGreen!90} 72.4 (+24.4) &  
        \cellcolor{LimeGreen!90} 72.4 (+24.4) &  
        \cellcolor{LimeGreen!90} 72.4 (+24.4) &  
        \cellcolor{LimeGreen!90} 72.4 (+24.4) &  
        64.6 & 
        62.3 \\
        
        2 & 
        \cellcolor{LimeGreen!50} 78.1 (+13.6) & 
        \cellcolor{LimeGreen!52} 76.7 (+14.0) & 
        \cellcolor{LimeGreen!68} 76.5 (+18.4) & 
        \cellcolor{LimeGreen!34} 79.1 (+9.2) &  
        \cellcolor{LimeGreen!42} 78.6 (+11.3) & 
        \cellcolor{LimeGreen!33} 80.3 (+9.0) & 
        \cellcolor{LimeGreen!49} 77.7 (+13.2) &  
        \cellcolor{LimeGreen!27} 79.8 (+7.3) &  
        \cellcolor{LimeGreen!10} 80.7 (+2.7) & 
        80.8 & 
        76.2 \\
        
        4 & 
        \cellcolor{LimeGreen!14} 82.5 (+3.7) & 
        \cellcolor{LimeGreen!22} 80.6 (+5.9) & 
        \cellcolor{LimeGreen!31} 78.8 (+8.4) & 
        \cellcolor{LimeGreen!5} 84.0 (+1.4) &  
        \cellcolor{LimeGreen!6} 82.1 (+1.7) & 
        \cellcolor{LimeGreen!5} 85.2 (+1.3) & 
        \cellcolor{LimeGreen!14} 81.9 (+3.7) &  
        \cellcolor{LimeGreen!2} 84.6 (+0.5) &  
        \cellcolor{Red!1} 83.7 (-0.1) & 
        86.8 & 
        81.9 \\

        8 & 
        \cellcolor{LimeGreen!1} 86.4 (+0.4) & 
        \cellcolor{LimeGreen!4} 85.3 (+1.1) & 
        \cellcolor{LimeGreen!10} 84.0 (+2.6) & 
        \cellcolor{LimeGreen!2} 88.3 (+0.4) &  
        \cellcolor{LimeGreen!4} 86.3 (+1.1) & 
        \cellcolor{LimeGreen!1} 88.8 (+0.4) & 
        \cellcolor{LimeGreen!1} 84.5 (+0.2) &  
        \cellcolor{Red!1} 88.8 (-0.1) &  
        \cellcolor{Red!5} 87.3 (-0.5) & 
        88.4 & 
        86.5 \\

        16 & 
        \cellcolor{Red!0} 89.2 (-0.0) & 
        \cellcolor{Red!0} 89.2 (-0.0) & 
        \cellcolor{LimeGreen!3} 88.6 (+0.8) & 
        \cellcolor{LimeGreen!1} 90.9 (+0.2) &  
        \cellcolor{LimeGreen!2} 89.0 (+0.6) & 
        \cellcolor{LimeGreen!0} 90.9 (+0.1) & 
        \cellcolor{Red!3} 88.0 (-0.3) &  
        \cellcolor{Red!2} 90.7 (-0.2) &  
        \cellcolor{Red!4} 90.4 (-0.4) & 
        89.3 & 
        89.1 \\
    \midrule
    \multicolumn{12}{c}{Food101} \\
    \midrule 
        1 & 
        \cellcolor{LimeGreen!90} 69.6 (+22.3) & 
        \cellcolor{LimeGreen!90} 69.6 (+22.3) & 
        \cellcolor{LimeGreen!90} 69.6 (+22.3) & 
        \cellcolor{LimeGreen!90} 69.6 (+22.3) & 
        \cellcolor{LimeGreen!90} 69.6 (+22.3) & 
        \cellcolor{LimeGreen!90} 69.6 (+22.3) & 
        \cellcolor{LimeGreen!90} 69.6 (+22.3) & 
        \cellcolor{LimeGreen!90} 69.6 (+22.3) & 
        \cellcolor{LimeGreen!90} 69.6 (+22.3) & 
        68.3 & 
        66.1 \\
        
        2 & 
        \cellcolor{LimeGreen!41} 74.5 (+10.1) & 
        \cellcolor{LimeGreen!76} 69.6 (+18.8) & 
        \cellcolor{LimeGreen!79} 69.4 (+19.5) & 
        \cellcolor{LimeGreen!42} 73.0 (+10.3) &  
        \cellcolor{LimeGreen!33} 69.7 (+18.9) & 
        \cellcolor{LimeGreen!27} 73.7 (+6.6) & 
        \cellcolor{LimeGreen!75} 69.9 (+18.5) &  
        \cellcolor{LimeGreen!33} 72.9 (+8.2) &  
        \cellcolor{LimeGreen!12} 77.1 (+3.0) & 
        79.1 & 
        72.3 \\
        
        4 & 
        \cellcolor{LimeGreen!8} 79.3 (+2.1) & 
        \cellcolor{LimeGreen!39} 73.2 (+9.7) & 
        \cellcolor{LimeGreen!55} 72.1 (+13.5) & 
        \cellcolor{LimeGreen!11} 78.1 (+2.7) &  
        \cellcolor{LimeGreen!40} 72.1 (+9.9) & 
        \cellcolor{LimeGreen!3} 79.0 (+0.8) & 
        \cellcolor{LimeGreen!35} 71.7 (+8.7) &  
        \cellcolor{LimeGreen!5} 78.7 (+1.2) &  
        \cellcolor{LimeGreen!4} 80.5 (+0.9) & 
        83.1 & 
        78.1 \\

        8 & 
        \cellcolor{LimeGreen!0} 83.9 (+0.1) & 
        \cellcolor{LimeGreen!9} 79.4 (+2.2) & 
        \cellcolor{LimeGreen!20} 77.9 (+4.9) & 
        \cellcolor{LimeGreen!1} 85.0 (+0.3) &  
        \cellcolor{LimeGreen!15} 78.0 (+3.7) & 
        \cellcolor{Red!3} 85.4 (-0.3) & 
        \cellcolor{LimeGreen!14} 76.8 (+3.5) &  
        \cellcolor{LimeGreen!1} 85.5 (+0.2) &  
        \cellcolor{Red!5} 85.1 (-0.5) & 
        86.0 & 
        81.9 \\

        16 & 
        \cellcolor{Red!2} 87.3 (-0.2) & 
        \cellcolor{LimeGreen!2} 85.6 (+0.4) & 
        \cellcolor{LimeGreen!7} 84.3 (+1.6) & 
        \cellcolor{Red!1} 89.4 (-0.1) &  
        \cellcolor{LimeGreen!6} 83.8 (+1.5) & 
        \cellcolor{LimeGreen!0} 89.4 (+0.0) & 
        \cellcolor{LimeGreen!3} 81.7 (+0.8) &  
        \cellcolor{LimeGreen!1} 89.5 (+0.1) &  
        \cellcolor{Red!3} 88.8 (-0.3) & 
        87.3 & 
        85.1 \\
    \midrule
    \multicolumn{12}{c}{ImageNet-100} \\
    \midrule
        1 & 
        \cellcolor{LimeGreen!90} 80.8 (+26.0) & 
        \cellcolor{LimeGreen!90} 80.8 (+26.0) & 
        \cellcolor{LimeGreen!90} 80.8 (+26.0) & 
        \cellcolor{LimeGreen!90} 80.8 (+26.0) & 
        \cellcolor{LimeGreen!90} 80.8 (+26.0) & 
        \cellcolor{LimeGreen!90} 80.8 (+26.0) & 
        \cellcolor{LimeGreen!90} 80.8 (+26.0) & 
        \cellcolor{LimeGreen!90} 80.8 (+26.0) & 
        \cellcolor{LimeGreen!90} 80.8 (+26.0) & 
        76.7 & 
        76.6 \\
        
        2 & 
        \cellcolor{LimeGreen!31} 85.4 (+8.9) & 
        \cellcolor{LimeGreen!82} 83.8 (+23.8) & 
        \cellcolor{LimeGreen!85} 82.5 (+24.7) & 
        \cellcolor{LimeGreen!25} 86.2 (+7.2) &  
        \cellcolor{LimeGreen!28} 86.6 (+7.9) & 
        \cellcolor{LimeGreen!14} 88.0 (+4.1) & 
        \cellcolor{LimeGreen!29} 84.8 (+8.3) &  
        \cellcolor{LimeGreen!18} 87.0 (+5.1) &  
        \cellcolor{LimeGreen!8} 87.3 (+2.4) & 
        89.5 & 
        89.1 \\
        
        4 & 
        \cellcolor{LimeGreen!7} 88.8 (+2.0) & 
        \cellcolor{LimeGreen!41} 86.2 (+11.8) & 
        \cellcolor{LimeGreen!59} 85.6 (+17.0) & 
        \cellcolor{Red!1} 88.9 (-0.1) &  
        \cellcolor{Red!3} 87.9 (-0.3) & 
        \cellcolor{Red!1} 90.6 (-0.1) & 
        \cellcolor{LimeGreen!3} 86.6 (+0.9) &  
        \cellcolor{Red!2} 89.6 (-0.2) &  
        \cellcolor{LimeGreen!0} 90.2 (-0.5) & 
        92.3 & 
        91.7 \\

        8 & 
        \cellcolor{LimeGreen!2} 91.6 (+0.6) & 
        \cellcolor{LimeGreen!10} 88.4 (+2.7) & 
        \cellcolor{LimeGreen!21} 87.8 (+6.0) & 
        \cellcolor{Red!7} 91.5 (-0.7) &  
        \cellcolor{Red!6} 89.2 (-0.6) & 
        \cellcolor{Red!7} 92.3 (-0.7) & 
        \cellcolor{Red!5} 88.6 (-0.5) &  
        \cellcolor{Red!3} 92.3 (-0.3) &  
        \cellcolor{Red!2} 93.2 (-0.2) & 
        93.3 & 
        92.7 \\

        16 & 
        \cellcolor{Red!3} 93.0 (-0.3) & 
        \cellcolor{LimeGreen!2} 91.0 (+0.5) & 
        \cellcolor{LimeGreen!3} 90.5 (+0.8) & 
        \cellcolor{Red!8} 93.5 (-0.8) &  
        \cellcolor{Red!1} 91.3 (-0.1) & 
        \cellcolor{Red!3} 94.1 (-0.3) & 
        \cellcolor{Red!7} 90.4 (-0.7) &  
        \cellcolor{Red!4} 93.8 (-0.4) &  
        \cellcolor{Red!1} 94.3 (-0.1) & 
        93.4 & 
        93.3 \\
    \midrule
    \multicolumn{12}{c}{DomainNet-Real} \\
    \midrule 
        1 & 
        \cellcolor{LimeGreen!90} 68.5 (+23.6) & 
        \cellcolor{LimeGreen!90} 68.5 (+23.6) & 
        \cellcolor{LimeGreen!90} 68.5 (+23.6) & 
        \cellcolor{LimeGreen!90} 68.5 (+23.6) & 
        \cellcolor{LimeGreen!90} 68.5 (+23.6) & 
        \cellcolor{LimeGreen!90} 68.5 (+23.6) & 
        \cellcolor{LimeGreen!90} 68.5 (+23.6) & 
        \cellcolor{LimeGreen!90} 68.5 (+23.6) &  
        \cellcolor{LimeGreen!90} 68.5 (+23.6) & 
        64.8 & 
        63.9 \\
        
        2 & 
        \cellcolor{LimeGreen!39} 71.9 (+10.1) & 
        \cellcolor{LimeGreen!63} 70.6 (+16.5) & 
        \cellcolor{LimeGreen!75} 70.1 (+19.7) & 
        \cellcolor{LimeGreen!39} 72.0 (+10.1) &  
        \cellcolor{LimeGreen!44} 71.7 (+11.5) & 
        \cellcolor{LimeGreen!30} 73.1 (+7.8) & 
        \cellcolor{LimeGreen!47} 71.1 (+12.2) &  
        \cellcolor{LimeGreen!28} 72.6 (+7.4) &  
        \cellcolor{LimeGreen!8} 73.6 (+2.1) & 
        73.0 & 
        73.8 \\
        
        4 & 
        \cellcolor{LimeGreen!11} 76.0 (+2.9) & 
        \cellcolor{LimeGreen!32} 72.9 (8.4) & 
        \cellcolor{LimeGreen!52} 72.2 (+13.5) & 
        \cellcolor{LimeGreen!12} 75.7 (+3.3) &  
        \cellcolor{LimeGreen!18} 73.9 (4.6) & 
        \cellcolor{LimeGreen!6} 77.1 (1.4) & 
        \cellcolor{LimeGreen!17} 73.7 (+4.5) &  
        \cellcolor{LimeGreen!5} 76.6 (+1.2) &  
        \cellcolor{Red!4} 76.2 (-0.4) & 
        74.8 & 
        77.8 \\

        8 & 
        \cellcolor{LimeGreen!1} 79.6 (+0.4) & 
        \cellcolor{LimeGreen!11} 77.1 (+3.0) & 
        \cellcolor{LimeGreen!24} 76.5 (+6.2) & 
        \cellcolor{LimeGreen!1} 80.1 (+0.2) &  
        \cellcolor{LimeGreen!5} 77.5 (+1.2) & 
        \cellcolor{LimeGreen!1} 81.0 (+0.3) & 
        \cellcolor{LimeGreen!4} 77.0 (+1.1) &  
        \cellcolor{LimeGreen!0} 80.8 (+0.1) &  
        \cellcolor{Red!3} 78.4 (-0.3) & 
        76.2 & 
        80.6 \\

        16 & 
        \cellcolor{LimeGreen!1} 82.2 (+0.1) & 
        \cellcolor{LimeGreen!4} 81.3 (+0.9) & 
        \cellcolor{LimeGreen!7} 80.7 (+1.7) & 
        \cellcolor{Red!0} 84.0 (-0.0) &  
        \cellcolor{LimeGreen!0} 81.2 (+0.1) & 
        \cellcolor{Red!0} 84.0 (-0.0) & 
        \cellcolor{LimeGreen!1} 80.5 (+0.1) &  
        \cellcolor{Red!1} 84.3 (-0.1) &  
        \cellcolor{Red!2} 79.5 (-0.2) & 
        78.0 & 
        82.3 \\
    \bottomrule
    \end{tabular}
    \end{adjustbox}
    \label{tab:delta_init_pool}
\end{table}

\subsection{On the importance of diversity in query selection}\label{diversity}

In line with the initial pool selection, we evaluate how partitioning the representation space via clustering is crucial for AL performance in subsequent iterations, even after the initial query. To conduct these experiments, we allow AL queries that use stand-alone measures of uncertainty (i.e. \uncertainty, \entropy, \margins, and \bald) to deviate from the top-$B$ acquisition pattern persistent in the AL literature. As noted by \cite{powerbald}, top-$B$ acquisition can potentially lead to a correlation between queried batches for some $t_i$ and $t_j$, which can hurt performance. 

We modify these queries to query the top-$(K \cdot B)$ samples based on their uncertainty metrics ($K=50$) and then cluster these samples by \textit{k}-means into $B$ clusters and select points closest to the cluster centroids. We also experiment with enabling dropout at inference time to add an element of stochasticity that disrupts the classifier's estimates of uncertainty, allowing for diverse sample selection. We report the results of our experiments in Table \ref{tab:delta_diversity}. To decouple the influence of the initial pool selection, all queries, including \typiclust and \probcover, use an identical randomly selected pool of initial samples. Our results show that by imposing simple diversity measures, uncertainty-based queries like \uncertainty, \entropy, \margins, and \bald surpass the performance of AL strategies that explicitly incorporate diversity in their queries.

\begin{table}[]
    \centering
    \caption{Marginal ($\Delta$) in performance for stand-alone uncertainty measures (i.e. softmax uncertainty, predictive entropy, margins sampling, or BALD) when using clustering ($\Delta_k$) or when using clustering with dropout during inference ($\Delta_{k+d}$). For fairness, we hold all queries to use a randomly selected initial pool. Power BALD, Coreset, BADGE, Alfa-mix, Typiclust, and ProbCover inherently enforce diversity in their respective queries, so we do not show $\Delta$s for these methods. The cells are color-coded according to the magnitude of $\Delta$ for better visualization.}
    \vspace{0.5em}
    \begin{adjustbox}{width=1\textwidth}
    \begin{tabular}{c | c c c | c c c | c c c | c c | c c c c c c}
    \toprule 
        \multirow{2}{*}{\centering $t$} & \multicolumn{3}{c|}{Uncertainty} & \multicolumn{3}{c|}{Entropy} & \multicolumn{3}{c|}{Margins} & \multicolumn{2}{c|}{BALD} & \multirow{2}{*}{\centering pBALD} & \multirow{2}{*}{\centering Coreset} & \multirow{2}{*}{\centering BADGE} & \multirow{2}{*}{\centering Alfa-mix} & \multirow{2}{*}{\centering Typiclust} & \multirow{2}{*}{\centering ProbCover} \\

        & Acc. & $\Delta_k$ & $\Delta_{k+d}$ & Acc. & $\Delta_k$ & $\Delta_{k+d}$
        & Acc. & $\Delta_k$ & $\Delta_{k+d}$ & Acc. & $\Delta_k$ 
        & & & & & & \\
    \midrule
    \multicolumn{18}{c}{CIFAR 100} \\
    \midrule
        2 & 
        79.6 & 
        \cellcolor{LimeGreen!55} +13.1 & 
        \cellcolor{LimeGreen!72} +17.0 & 
        79.4 & 
        \cellcolor{LimeGreen!70} +16.6 & 
        \cellcolor{LimeGreen!90} +21.4 &  
        80.4 & 
        \cellcolor{LimeGreen!33} +7.9 & 
        \cellcolor{LimeGreen!44} +10.5 &  
        76.9 & 
        \cellcolor{LimeGreen!40} +9.5 &  
        71.2 & 64.5 & 72.5 & 78.0 & 78.9 & 74.6 \\
        
        4 & 
        86.7 & 
        \cellcolor{LimeGreen!51} +12.0 & 
        \cellcolor{LimeGreen!51} +12.0 & 
        86.7 & 
        \cellcolor{LimeGreen!67} +15.9 & 
        \cellcolor{LimeGreen!69} +16.3 &  
        87.1 & 
        \cellcolor{LimeGreen!17} +4.1 & 
        \cellcolor{LimeGreen!19} +4.5 &  
        86.6 & 
        \cellcolor{LimeGreen!26} +6.2 & 
        83.9 & 78.2 & 84.1 & 83.9 & 86.4 & 82.2 \\

        8 & 
        89.9 & 
        \cellcolor{LimeGreen!23} +5.6 & 
        \cellcolor{LimeGreen!24} +5.7 & 
        89.8 & 
        \cellcolor{LimeGreen!35} +8.2 & 
        \cellcolor{LimeGreen!35} +8.4 &  
        89.8 & 
        \cellcolor{LimeGreen!8} +1.8 & 
        \cellcolor{LimeGreen!8} +1.9 &  
        89.3 & 
        \cellcolor{LimeGreen!17} +4.1 &  
        88.4 & 88.3 & 90.9 & 90.8 & 90.0 & 89.1 \\

        16 & 
        91.2 & 
        \cellcolor{LimeGreen!9} +2.2 & 
        \cellcolor{LimeGreen!8} +1.9 & 
        91.4 & 
        \cellcolor{LimeGreen!15} +3.5 & 
        \cellcolor{LimeGreen!15} +3.6 &  
        91.4 & 
        \cellcolor{LimeGreen!4} +0.9 & 
        \cellcolor{LimeGreen!3} +0.7 &  
        91.0 & 
        \cellcolor{LimeGreen!11} +2.7 & 
        90.8 & 88.3 & 90.9 & 90.8 & 90.0 & 89.1 \\ 
    \midrule
    \multicolumn{18}{c}{Food101} \\
    \midrule
        2 & 
        70.3 & 
        \cellcolor{LimeGreen!67} +16.1 & 
        \cellcolor{LimeGreen!81} +19.4 & 
        68.8 & 
        \cellcolor{LimeGreen!60} +14.3 & 
        \cellcolor{LimeGreen!79} +18.9 &  
        72.3 & 
        \cellcolor{LimeGreen!31} +7.4 & 
        \cellcolor{LimeGreen!40} +9.6 &  
        64.0 & 
        \cellcolor{LimeGreen!55} +13.2 & 
        67.1 & 51.5 & 64.8 & 74.1 & 77.0 & 73.8 \\
        
        4 & 
        81.8 & 
        \cellcolor{LimeGreen!57} +17.0 & 
        \cellcolor{LimeGreen!61} +18.3 & 
        80.0 & 
        \cellcolor{LimeGreen!59} +17.6 & 
        \cellcolor{LimeGreen!68} +20.2 &  
        81.7 & 
        \cellcolor{LimeGreen!24} +5.7 & 
        \cellcolor{LimeGreen!26} +6.2 &  
        77.5 & 
        \cellcolor{LimeGreen!52} +15.3 & 
        78.2 & 63.0 & 77.5 & 79.6 & 82.5 & 78.6 \\

        8 & 
        86.4 & 
        \cellcolor{LimeGreen!39} +9.3 & 
        \cellcolor{LimeGreen!38} +9.1 & 
        85.7 & 
        \cellcolor{LimeGreen!46} +11.1 & 
        \cellcolor{LimeGreen!53} +12.7 &  
        86.1 & 
        \cellcolor{LimeGreen!6} +1.5 & 
        \cellcolor{LimeGreen!6} +1.5 &  
        84.3 & 
        \cellcolor{LimeGreen!42} +10.0 &  
        85.6 & 73.4 & 85.3 & 85.6 & 85.8 & 82.3 \\

        16 & 
        88.9 & 
        \cellcolor{LimeGreen!19} +4.5 & 
        \cellcolor{LimeGreen!15} +3.6 & 
        88.7 & 
        \cellcolor{LimeGreen!25} +5.9 & 
        \cellcolor{LimeGreen!25} +6.1 &  
        89.0 & 
        \cellcolor{LimeGreen!1} +0.1 & 
        \cellcolor{Red!4} -0.4 &  
        87.9 & 
        \cellcolor{LimeGreen!23} +5.6 & 
        89.4 & 80.8 & 89.4 & 89.2 & 87.4 & 85.3 \\
    \midrule
    \multicolumn{18}{c}{ImageNet-100} \\
    \midrule
        2 & 
        88.5 & 
        \cellcolor{LimeGreen!64} +21.6 & 
        \cellcolor{LimeGreen!84} +28.5 & 
        88.4 & 
        \cellcolor{LimeGreen!62} +21.1 & 
        \cellcolor{LimeGreen!90} +30.5 &  
        88.5 & 
        \cellcolor{LimeGreen!25} +8.5 & 
        \cellcolor{LimeGreen!28} +9.5 &  
        86.7 & 
        \cellcolor{LimeGreen!24} +8.0 & 
        83.9 & 76.5 & 81.8 & 84.9 & 89.1 & 87.0 \\
        
        4 & 
        90.6 & 
        \cellcolor{LimeGreen!46} +15.6 & 
        \cellcolor{LimeGreen!23} +16.2 & 
        90.4 & 
        \cellcolor{LimeGreen!62} +20.9 & 
        \cellcolor{LimeGreen!64} +21.8 &  
        91.2 & 
        \cellcolor{LimeGreen!7} +2.3 & 
        \cellcolor{LimeGreen!7} +2.2 &  
        90.2 & 
        \cellcolor{LimeGreen!6} +2.0 &  
        90.6 & 85.7 & 89.8 & 90.7 & 92.4 & 91.8 \\

        8 & 
        93.3 & 
        \cellcolor{LimeGreen!23} +7.6 & 
        \cellcolor{LimeGreen!23} +7.6 & 
        92.0 & 
        \cellcolor{LimeGreen!33} +11.3 & 
        \cellcolor{LimeGreen!30} +10.3 &  
        93.3 & 
        \cellcolor{LimeGreen!4} +1.3 & 
        \cellcolor{LimeGreen!3} +1.1 &  
        91.4 & 
        \cellcolor{LimeGreen!5} +1.6 & 
        92.9 & 89.0 & 92.6 & 93.4 & 93.3 & 92.7 \\

        16 & 
        94.1 & 
        \cellcolor{LimeGreen!12} +4.1 & 
        \cellcolor{LimeGreen!11} +3.6 & 
        94.2 & 
        \cellcolor{LimeGreen!14} +4.7 & 
        \cellcolor{LimeGreen!13} +4.5 &  
        94.4 & 
        \cellcolor{LimeGreen!1} +0.3 & 
        \cellcolor{LimeGreen!1} +0.2 &  
        93.3 & 
        \cellcolor{LimeGreen!6} +1.9 & 
        94.4 & 91.1 & 94.1 & 94.4 & 93.8 & 93.8 \\
    \midrule
    \multicolumn{18}{c}{DomainNet-Real} \\
    \midrule
        2 & 
        71.5 & 
        \cellcolor{LimeGreen!54} +12.6 & 
        \cellcolor{LimeGreen!75} +17.4 & 
        71.2 & 
        \cellcolor{LimeGreen!67} +15.4 & 
        \cellcolor{LimeGreen!90} +20.8 &  
        72.3 & 
        \cellcolor{LimeGreen!31} +7.1 & 
        \cellcolor{LimeGreen!45} +10.4 &  
        68.1 & 
        \cellcolor{LimeGreen!34} +7.8 & 
        65.3 & 58.8 & 65.2 & 71.4 & 72.4 & 71.3 \\
        
        4 & 
        79.4 & 
        \cellcolor{LimeGreen!60} +13.9 & 
        \cellcolor{LimeGreen!65} +14.9 & 
        78.8 & 
        \cellcolor{LimeGreen!84} +19.5 & 
        \cellcolor{LimeGreen!87} +20.1 &  
        79.1 & 
        \cellcolor{LimeGreen!28} +6.5 & 
        \cellcolor{LimeGreen!28} +6.6 &  
        78.0 & 
        \cellcolor{LimeGreen!38} +8.7 & 
        75.6 & 69.2 & 75.4 & 76.6 & 75.2 & 77.3 \\

        8 & 
        82.7 & 
        \cellcolor{LimeGreen!36} +8.3 & 
        \cellcolor{LimeGreen!37} +8.5 & 
        82.5 & 
        \cellcolor{LimeGreen!53} +12.2 & 
        \cellcolor{LimeGreen!53} +12.2 &  
        82.4 & 
        \cellcolor{LimeGreen!11} +2.6 & 
        \cellcolor{LimeGreen!11} +2.5 &  
        81.7 & 
        \cellcolor{LimeGreen!23} +5.4 & 
        80.7 & 76.0 & 80.7 & 78.7 & 77.0 & 80.4 \\

        16 & 
        84.6 & 
        \cellcolor{LimeGreen!19} +4.4 & 
        \cellcolor{LimeGreen!18} +4.2 & 
        84.5 & 
        \cellcolor{LimeGreen!25} +5.8 & 
        \cellcolor{LimeGreen!24} +5.5 &  
        84.4 & 
        \cellcolor{LimeGreen!3} +0.7 & 
        \cellcolor{LimeGreen!2} +0.4 &  
        84.5 & 
        \cellcolor{LimeGreen!14} +3.3 & 
        84.0 & 80.4 & 84.3 & 79.7 & 78.3 & 82.1 \\
    \bottomrule
    \end{tabular}
    \end{adjustbox}
    \label{tab:delta_diversity}
\end{table}

\subsection{Representative versus uncertainty sampling and the phase transition}\label{uncertain_rep}

Much prior art in AL has placed an emphasis on the distinction between sampling the most uncertain instances in $\mathcal{D}_U$ for the active learner and sampling diverse, representative instances for establishing well-defined decision boundaries. It is a well-known fact that querying diverse but relatively representative instances early on in the AL procedure allows for the classifier to learn the structure of the representation space early in order to make uncertainty estimates in later iterations more reliable \citep{typiclust}. This trade-off between diversity sampling in the low-data regime before uncertainty sampling is known as the \textit{phase transition} \citep{typiclust}.

However, when leveraging large foundation models as feature extractors, we observe a phenomenon that is a striking deviation from the existing literature. As demonstrated in Table \ref{tab:delta_diversity}, when controlling for random initialization, uncertainty sampling is actually competitive to representative sampling methods like \typiclust as early as the 2nd AL iteration (CIFAR100, DomainNet-Real) and even beat these methods in later iterations. Using foundation models in our diverse selection of datasets, we see no evidence of a phase transition and find no support for the notion that uncertainty sampling is ineffective in low-budget AL. Further, we witness that \typiclust underperforms the uncertainty-based queries in later iterations (8-16, Food101 \& DomainNet-Real).

\subsection{Leveraging unlabeled instances for training the active learner}\label{semisup}

Even with a good initial query, the classifier $g(\bm{z}_i; \bm{\theta})$ may perform poorly in the low-data regime since there are a limited number of training examples. In cases where the labeling budget is prohibitive, acquiring enough training instances for a classifier to establish good decision boundaries early on in AL may be difficult. This motivates the use of unlabeled instances in a principled way to ensure that the classifier maximizes performance even in the low-data regime. However, our experiments find that the efficacy of semi-supervised learning in this setting is questionable at best, with wide variation across query methods and datasets. While we do see an initial boost in performance, contrary to \cite{consistency} and \cite{rethinking}, the gap quickly narrows, and label propagation underperforms the reference query. Propagating labels from uncertain queried samples may cause points across the decision boundary to be assigned to incorrect classes, hampering performance in the long run. We point the reader to the Appendix for an in-depth analysis using a popular semi-supervised algorithm.

\subsection{Summary of results}
Our experiments with initial pool selection (we study the importance of intelligently selecting the initial labeled pool by comparing centroid-based initialization with random initialization) motivate the use of a centroid-based initialization strategy to overcome the cold start problem. Additionally, our experiments modifying existing uncertainty-based approaches to enforce diversity in query selection, which motivates the use of clustering to select diverse samples during active learning. Our findings also indicate that a clustering approach similar to \alfamix selects diverse candidates that span the representation space. Furthermore, our investigation into the previously identified phase transition implicates the need for an uncertainty-based query function instead of representative sampling throughout active learning. We revisited the notion that uncertainty-based queries are outperformed by representative sample selection methods such as \typiclust in the low-budget regime. We find that this is not the case, contradicting this existing notion. Finally, our inquiry into enhancing the active learner with unlabeled instances cautions against the use of semi-supervised learning as a complementary approach to active learning. 

Based on our findings, an effective AL strategy would initialize the labeled set with representative samples, employ an uncertainty-based query function for shortlisting unlabeled candidates, and select a diverse subset of these candidates.

\section{\dropout, a simple, effective active learning strategy}

The observations from the previous sections directly inform the principled construction of a new AL strategy leveraging the robust visual features of foundation models, which we refer to as \dropout. A detailed description of the query strategy is provided in Algorithm \ref{alg:dropquery}. Below, we briefly review the key results from section 3 that motivate the choice of components and the construction of \dropout.

\begin{itemize}
    \item \textbf{Centroid-based initial pool selection:} Our experimental results in Section \ref{init_pool} demonstrated the utility of intelligently selecting the initial pool of candidates to label. Informed by these results and given the semantic clusters characteristic to the latent spaces of foundation models, \dropout employs a centroid-based initial pool selection strategy to overcome the cold-start problem.
    \item \textbf{Uncertainy-based sampling approach:} Based on our investigation into the tradeoff between uncertainty-based and representative query sampling in Section \ref{uncertain_rep}, \dropout favors an uncertainty-based query strategy.  We experiment using dropout perturbations to measure uncertainty and select candidates for annotation. Given features $\bm{z}_i$ of an unlabeled instance, we produce $M$ dropout samples of these inputs ($\rho = 0.75$) to get $\{\bm{z}^1_i, \dots, \bm{z}^M_i\}$. The current classifier at $t$ then makes predictions on these $M$ samples and the original instance $\bm{z}_i$, and we measure the consistency of classifier predictions $\hat{y}_i = g(\bm{z}_i; \bm{\theta})$. If more than half of the $M$ predictions are inconsistent, we add $\bm{z}_i$ to the candidate set $Z_{c}$. In all of our experiments, we set $M=3$.
    \item \textbf{Selecting a diverse pool of candidates:} Our analysis in Section \ref{diversity} provided evidence in favor of choosing a diverse set of candidates to annotate. \dropout leverages this result by taking the candidate set $Z_c$ and clustering it into $B$ clusters, selecting the point closest to the center of each cluster. These constitute a diverse subset of points to be annotated.
    \item \textbf{Leveraging unlabeled instances:} The results from our investigation in Section \ref{semisup} did not provide conclusive evidence in favor of adding a semi-supervised learning component to \dropout, hence we choose to omit this particular variant of label propagation from our strategy.
\end{itemize}

\begin{minipage}{1\textwidth}
\begin{algorithm}[H]
\small
\caption{DropQuery}\label{alg:dropquery}
\begin{algorithmic}
\Require Given unlabeled instances $\bm{z}_i \in Z_U$, external oracle $\phi(.)$, budget $B$.
\Ensure Queried labels $Y = \{y_i : i\in 1, \dots, B\}$
\For{$\bm{z}_i \in Z$}
    \State $\{\bm{z}_i^1, \dots, \bm{z}_i^M\} \gets$ Dropout$(\bm{z}_i; \rho)$ \Comment Apply dropout to the input features
    \State $n_i \gets \sum\limits_{m=1}^{M} \mathds{1}[g(\bm{z}_i^m; \bm{\theta}) = g(\bm{z}_i; \bm{\theta})]$ \Comment Measure inconsistency among classifier predictions
\EndFor
\State $Z_{c} = \{\bm{z}_i | n_i > 0.5M \  \forall z_i \in Z_U\}$\Comment Use consistency as proxy for uncertainty
\State $\bm{C}_k \gets K$-means$(Z_c, B)$ where $k \in \{B\}$ \Comment Cluster points to enforce diversity
\State $S_k \gets \{\bm{z}_i : \argmin_{\bm{z}_i \in Z_c} \lVert \bm{z}_i - \bm{c}_k \rVert_2^2$ where $\bm{c}_k \in C_k\}$ 
\State $Y \gets \phi(S_k)$ \Comment Retrieve labels from samples closest to centroids
\end{algorithmic}
\end{algorithm}
\end{minipage}
\vspace{0.5em}

We take inspiration from works like \cite{batchwisedropout} and  \cite{powerensembles}, which leverage query-by-committee to distinguish informative samples. It is well-known that dropout on the weights of a classifier during inference simulates an ensemble of models \citep{empiricaldropout}. However, our strategy is a nuanced but critical change from these works. \dropout is more similar in spirit to ALFA-Mix, which interpolates unlabeled points with anchors to perturb them and tests the consistency of their class predictions. Applying dropout perturbs features, and those lying close to the decision boundary will have inconsistent predictions, making them good candidates for querying. Candidate clustering ensures the diversity of selected points, as shown in Table \ref{tab:dropout_query}. Since label propagation does not help queries in our experiments, we do not consider it a core component.

\begin{table}[h]
    \centering
    \caption{Test accuracy for our method (with random acquisition as a reference) utilizing dropout at inference time (DQ), with a centroid-based initial pool (DQ$_c$), and dropout with semi-supervised learning in the form of label propagation (DQ$_{ssl}$).}
    \vspace{0.5em}
    \begin{adjustbox}{width=1\textwidth}
    \begin{tabular}{c | c c c c | c c c c| c c c c | c c c c}
    \toprule 
        \rowcolor{gray!10} $t$ & Random & DQ & DQ$_c$ & DQ$_{ssl}$ & Random & DQ & DQ$_c$ & DQ$_{ssl}$ & Random & DQ & DQ$_c$ & DQ$_{ssl}$ & Random & DQ & DQ$_c$ & DQ$_{ssl}$\\
    \midrule
    & \multicolumn{4}{c}{CIFAR100} & \multicolumn{4}{c}{ImageNet-100} & \multicolumn{4}{c}{Food101} & \multicolumn{4}{c}{DomainNet-Real} \\
    \midrule
        1 & 48.1 & 48.1 & 72.6 & 53.3 & 54.7 & 54.7 & 79.1 & 57.8 & 47.4 & 47.4 & 71.3 & 49.7 & 44.8 & 44.8 & 68.0 & 46.8 \\
        2 & 64.5 & 81.0 & 83.5 & 79.2 & 76.5 & 88.8 & 89.4 & 85.9 & 64.4 & 72.7 & 76.1 & 70.6 & 61.8 & 73.0 & 75.0 & 68.9 \\
        4 & 78.7 & 87.4 & 87.7 & 87.1 & 86.8 & 91.7 & 91.5 & 90.9 & 77.2 & 81.8 & 82.2 & 80.9 & 73.1 & 79.1 & 78.7 & 76.7 \\
        8 & 86.1 & 89.8 & 90.1 & 89.6 & 91.0 & 93.2 & 93.0 & 92.8 & 83.8 & 86.3 & 86.3 & 84.7 & 79.2 & 82.4 & 82.2 & 80.0 \\
        16 & 89.2 & 91.4 & 91.5 & 91.0 & 93.3 & 94.1 & 94.1 & 93.7 & 87.5 & 89.5 & 89.4 & 86.7 & 82.1 & 84.8 & 84.7 & 82.1 \\
    \bottomrule
    \end{tabular}
    \end{adjustbox}
    \label{tab:dropout_query}
\end{table}

\section{Experimental results}

\begin{figure}[!h]
\centering
\begin{subfigure}{\textwidth}
    \centering
    \includegraphics[trim={0 0 0 0},clip,width=\linewidth]{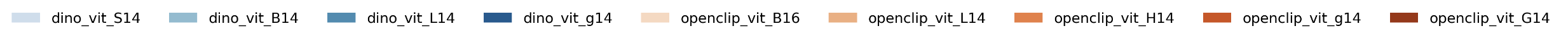}
\end{subfigure}

\begin{subfigure}{.245\textwidth}
    \centering
    \includegraphics[width=\linewidth]{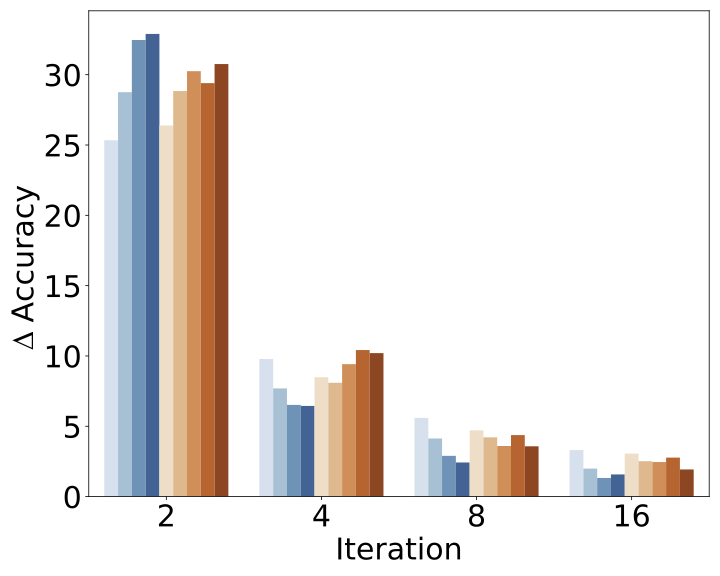}
    \caption{CIFAR100}
    \label{fig1cifar100}
\end{subfigure}
\begin{subfigure}{.245\textwidth}
    \centering
    \includegraphics[width=\linewidth]{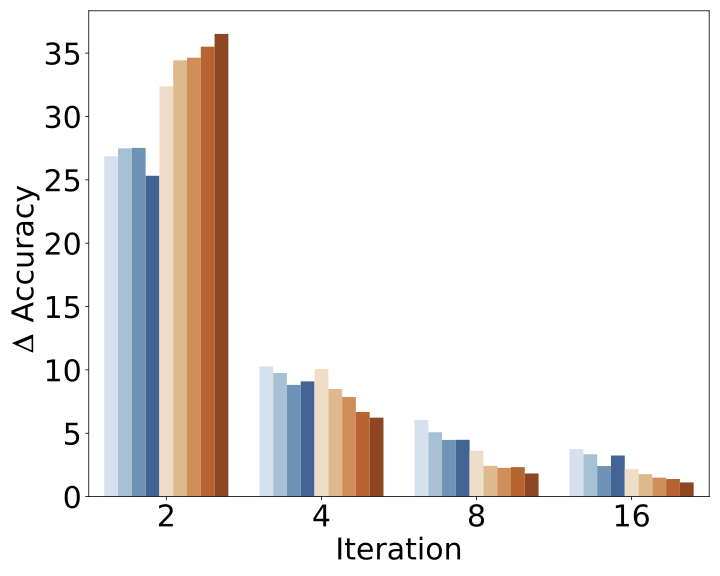}    
    \caption{Food101}
    \label{fig1food101}
\end{subfigure}
\begin{subfigure}{.245\textwidth}
    \centering
    \includegraphics[width=\linewidth]{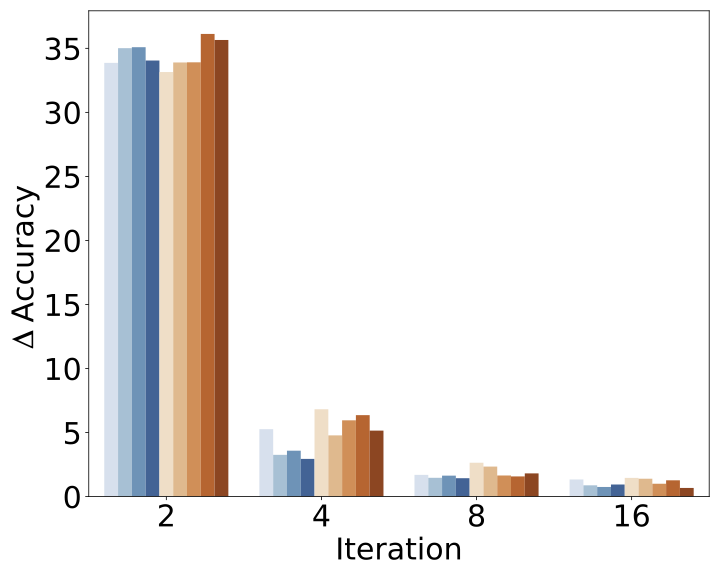} 
    \caption{ImageNet-100}
    \label{fig1imagenet100}
\end{subfigure}
\begin{subfigure}{.245\textwidth}
    \centering
    \includegraphics[width=\linewidth]{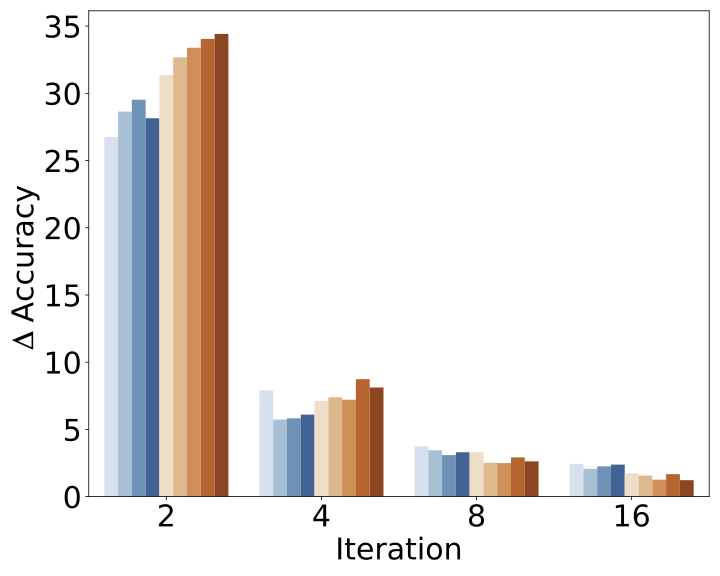} 
    \caption{DomainNet-Real}
    \label{fig1domainnetreal}
\end{subfigure}
\caption{Results of our AL strategy on different representation spaces (i.e. DINOv2 \citep{dinov2} and OpenCLIP \citep{openclip}). The y-axis is the delta in accuracy between iteration $i$ and $i/2$. In early iterations, the improvements to AL query performance are more pronounced for larger models.}
\label{fig:eval_models}
\end{figure}

Our strategy's effectiveness is demonstrated on several natural, out-of-domain biomedical, and large-scale image datasets. Initially, we assess \dropout's performance, maintaining the dataset constant while altering the underlying representation space (see Figure \ref{fig:eval_models}). Our experiments confirm that larger models generate more resilient visual features, thereby enhancing linear classifier performance in low-budget scenarios. An intriguing phenomenon, observed during iterations 4-16 in Figure \ref{fig:eval_models}, reveals that \textit{smaller} models produce embeddings that our AL strategy better accommodates in later iterations. The performance benefits of larger models in early iterations (limited data availability) diminish rapidly, likely due to their expressiveness compared to smaller models' features. Hence, dropout-motivated uncertainty sampling is not as advantageous. We further note that, on average, vision-language foundation models outperform their vision-only counterparts.

\subsection{Natural Image Datasets}
Our proposed AL strategy is evaluated through a comprehensive set of experiments on diverse natural image datasets sourced from the VTAB+ benchmark \citep{laion5b}. The VTAB+ benchmark, an expanded version of the VTAB \citep{vtab} benchmark, encompasses a range of challenging visual perception tasks including fine-grained image classification and is typically used for assessing zero-shot performance. A notable drawback in many prior AL studies is the excessive reliance on datasets such as CIFAR100 or TinyImageNet with small image sizes that are not representative of real-world scenarios where AL would be necessary. Therefore, our experiments include several larger, more complex natural image datasets that closely mirror real-world scenarios. We describe extensive implementation and training details in the Appendix. 

\begin{figure}[!t]
\centering
\begin{subfigure}{\textwidth}
    \centering
    \includegraphics[trim={0 6cm 0 0},clip,width=\linewidth]{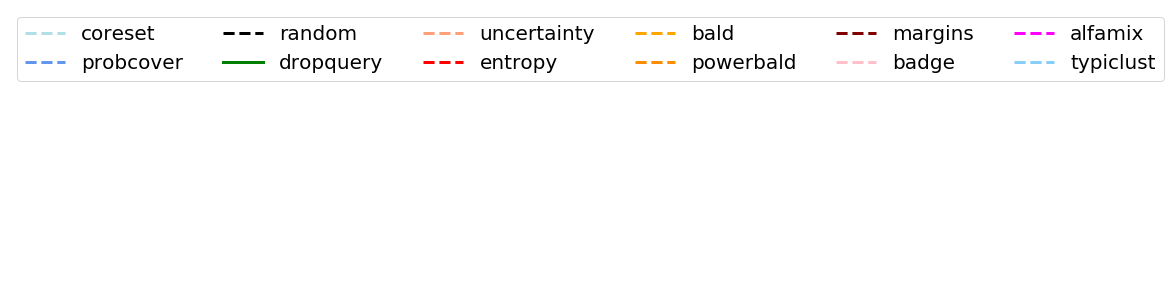}
\end{subfigure}

\begin{subfigure}{.245\textwidth}
    \centering
    \includegraphics[width=\linewidth]{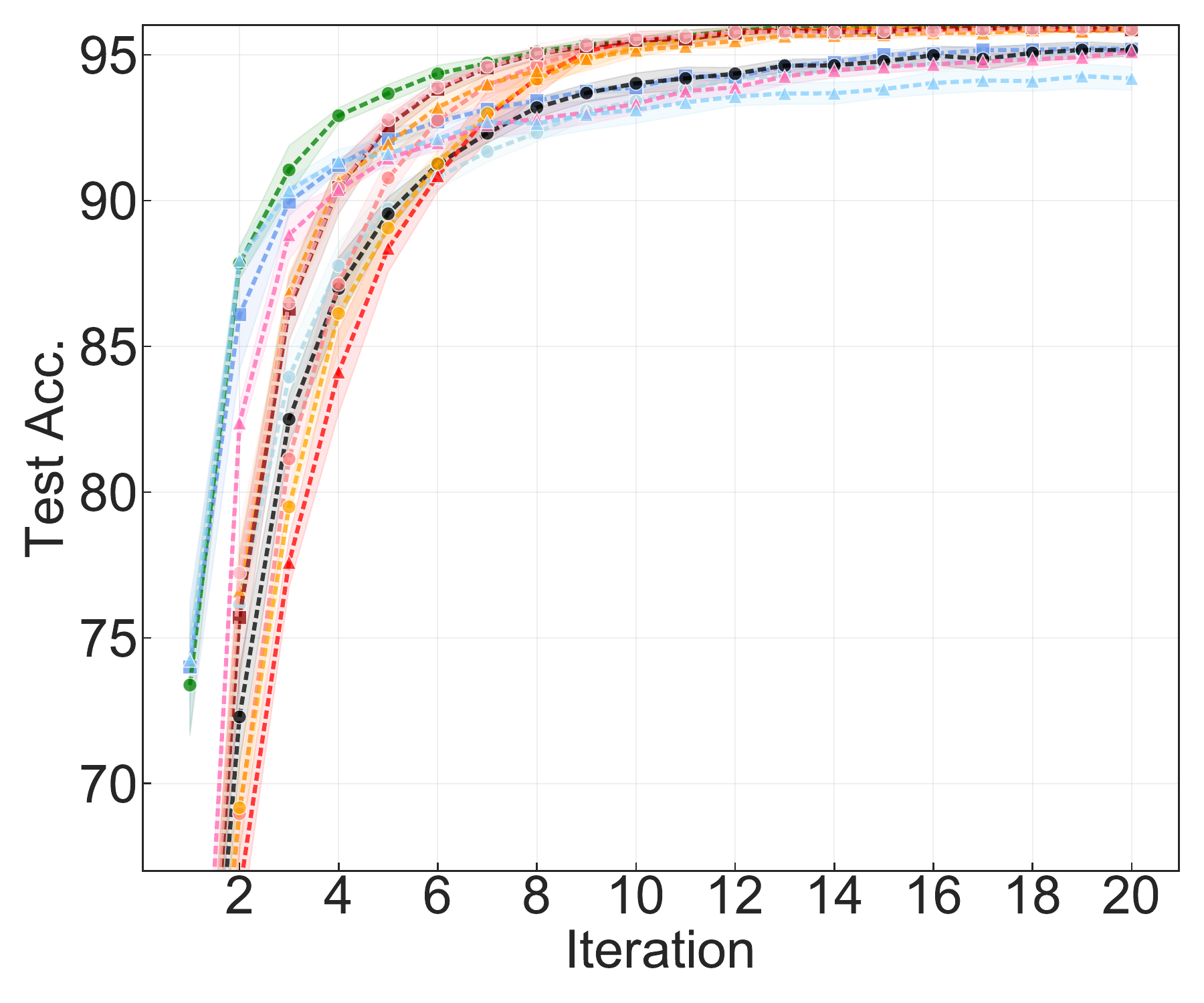}
    \caption{Cars}
    \label{fig2cars}
\end{subfigure}
\begin{subfigure}{.245\textwidth}
    \centering
    \includegraphics[width=\linewidth]{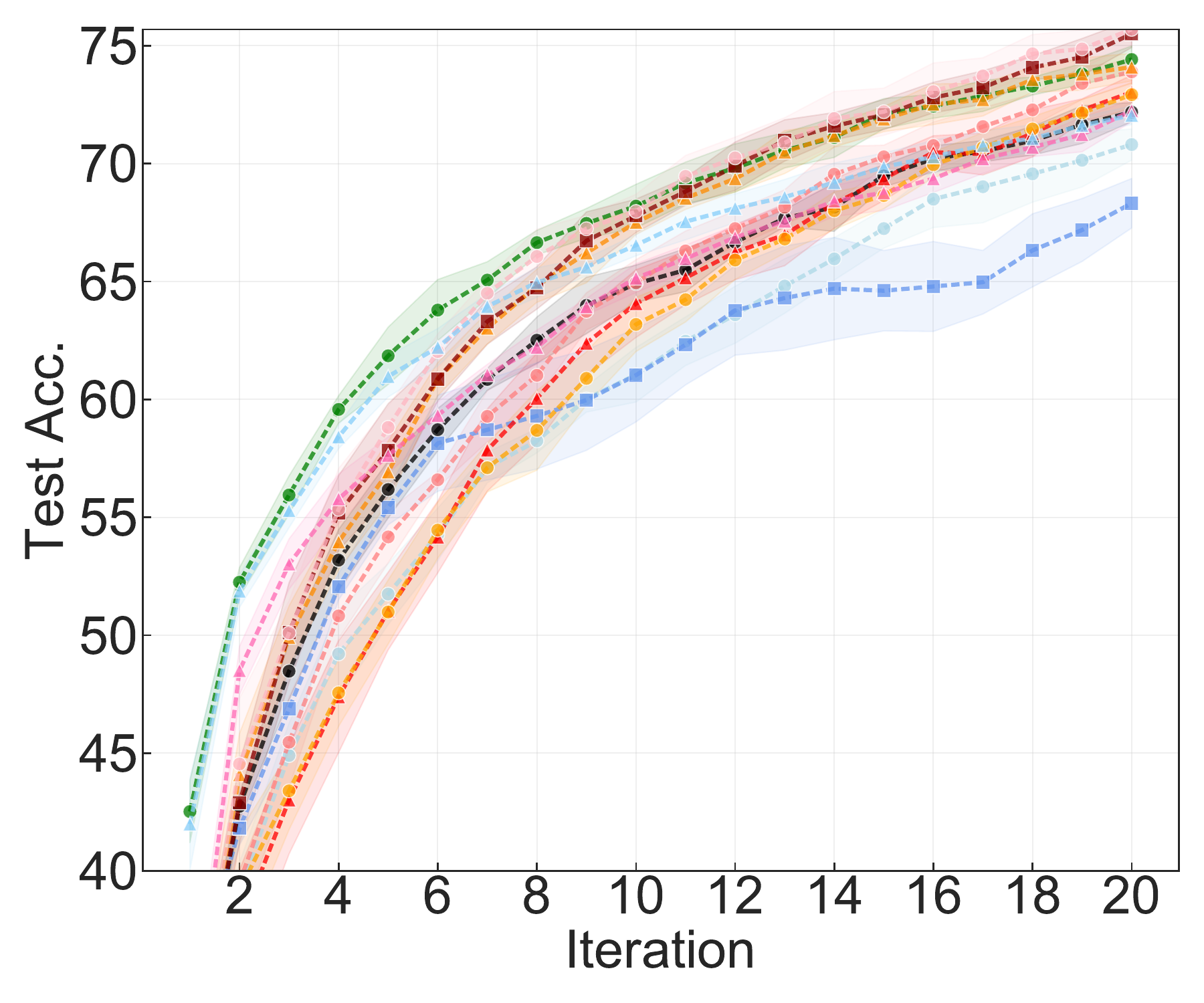}    
    \caption{Aircraft}
    \label{fig2aircraft}
\end{subfigure}
\begin{subfigure}{.245\textwidth}
    \centering
    \includegraphics[width=\linewidth]{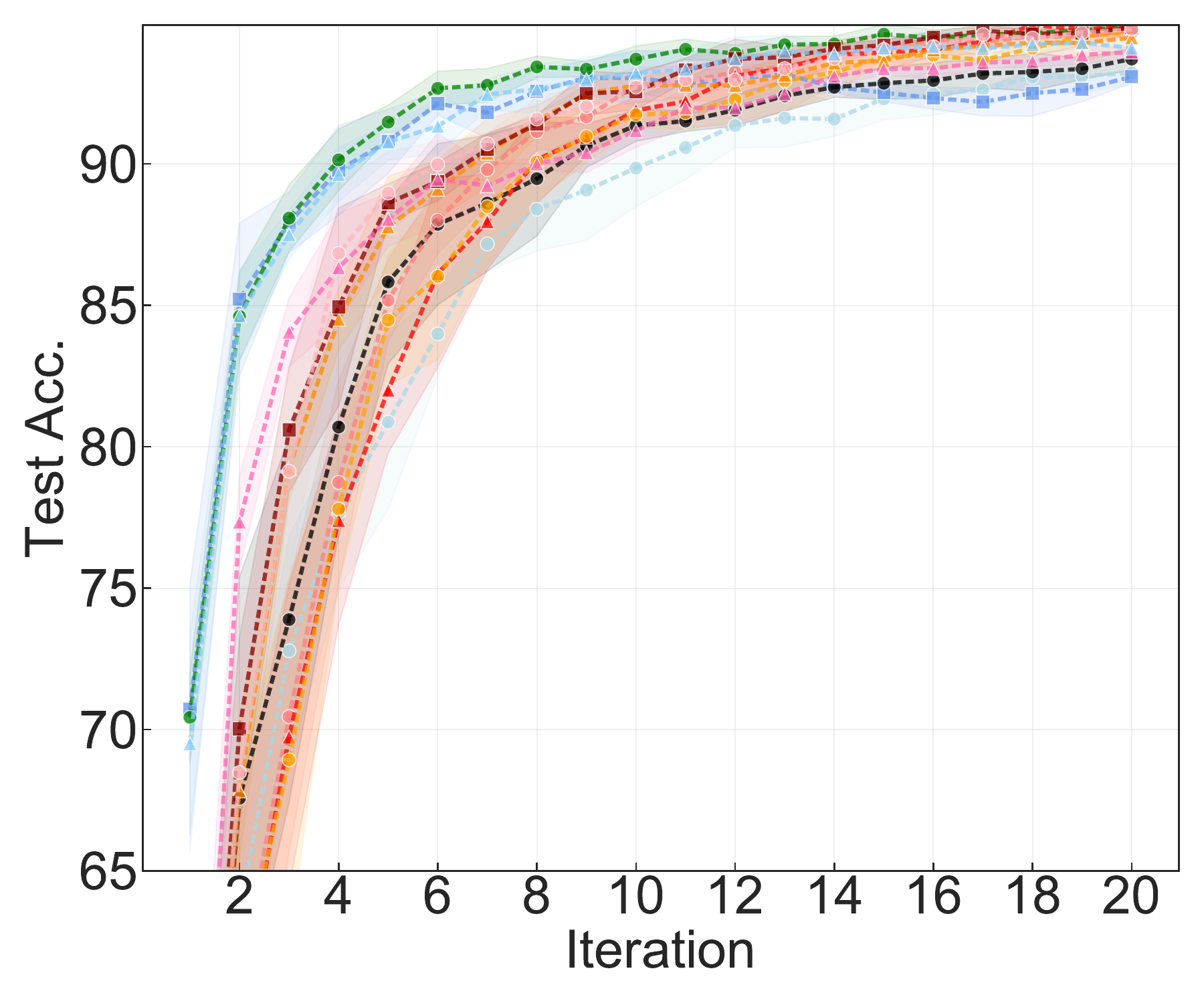} 
    \caption{Pets}
    \label{fig2pets}
\end{subfigure}
\begin{subfigure}{.245\textwidth}
    \centering
    \includegraphics[width=\linewidth]{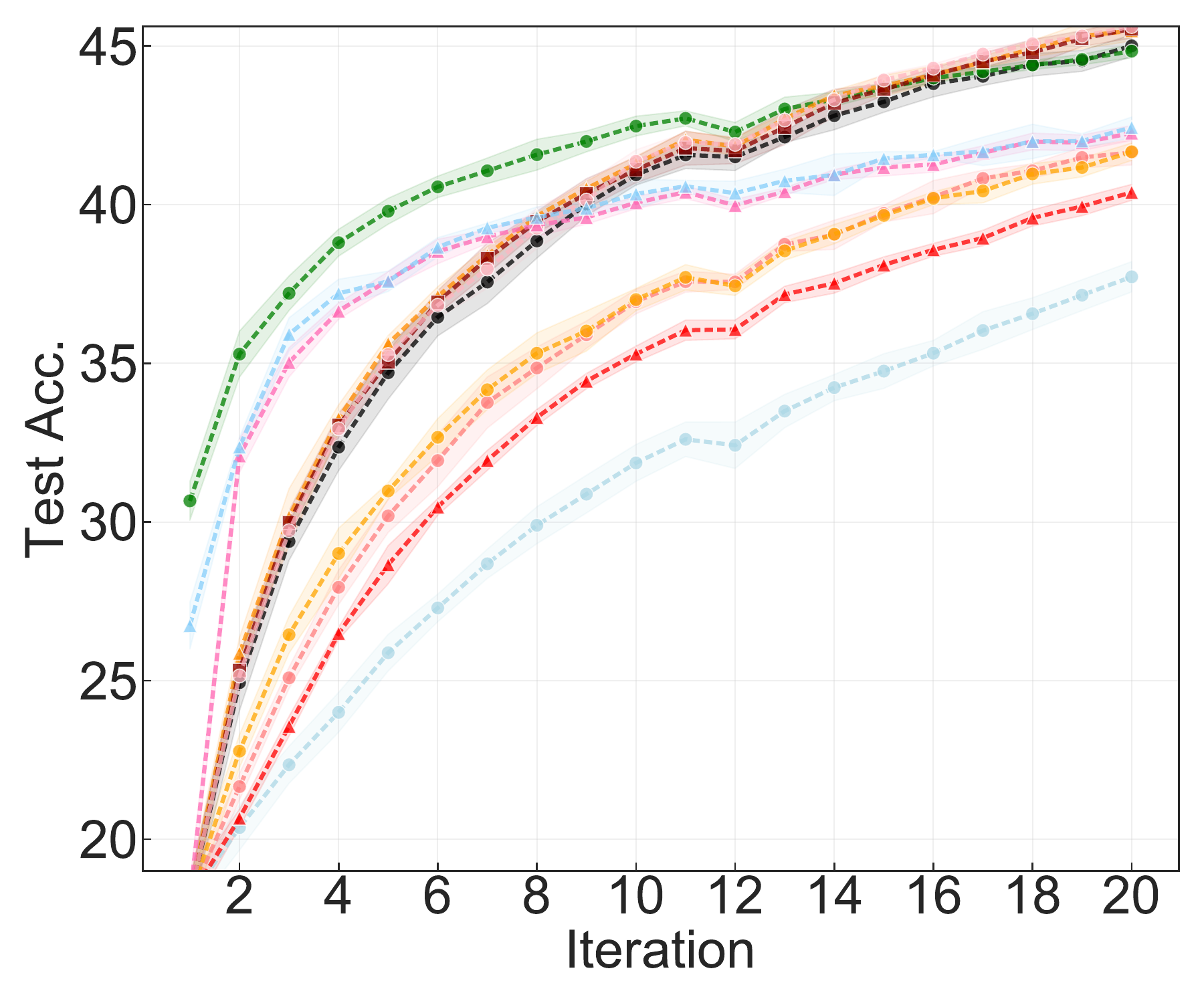} 
    \caption{Places365}
    \label{fig2places}
\end{subfigure}

\begin{subfigure}{.245\textwidth}
    \centering
    \includegraphics[width=\linewidth]{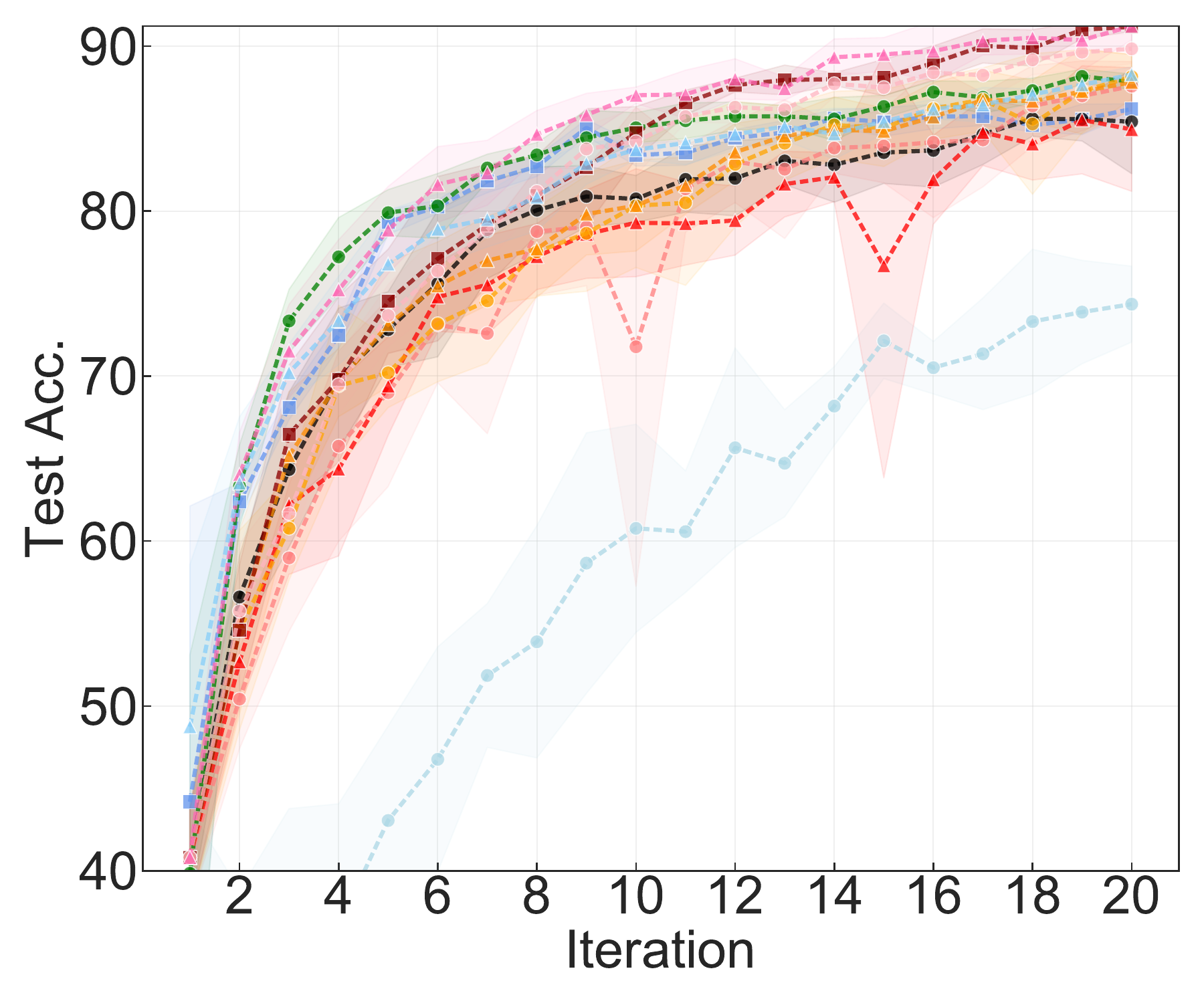}
    \caption{Blood Smear}
    \label{fig2bloodsmear}
\end{subfigure}
\begin{subfigure}{.245\textwidth}
    \centering
    \includegraphics[width=\linewidth]{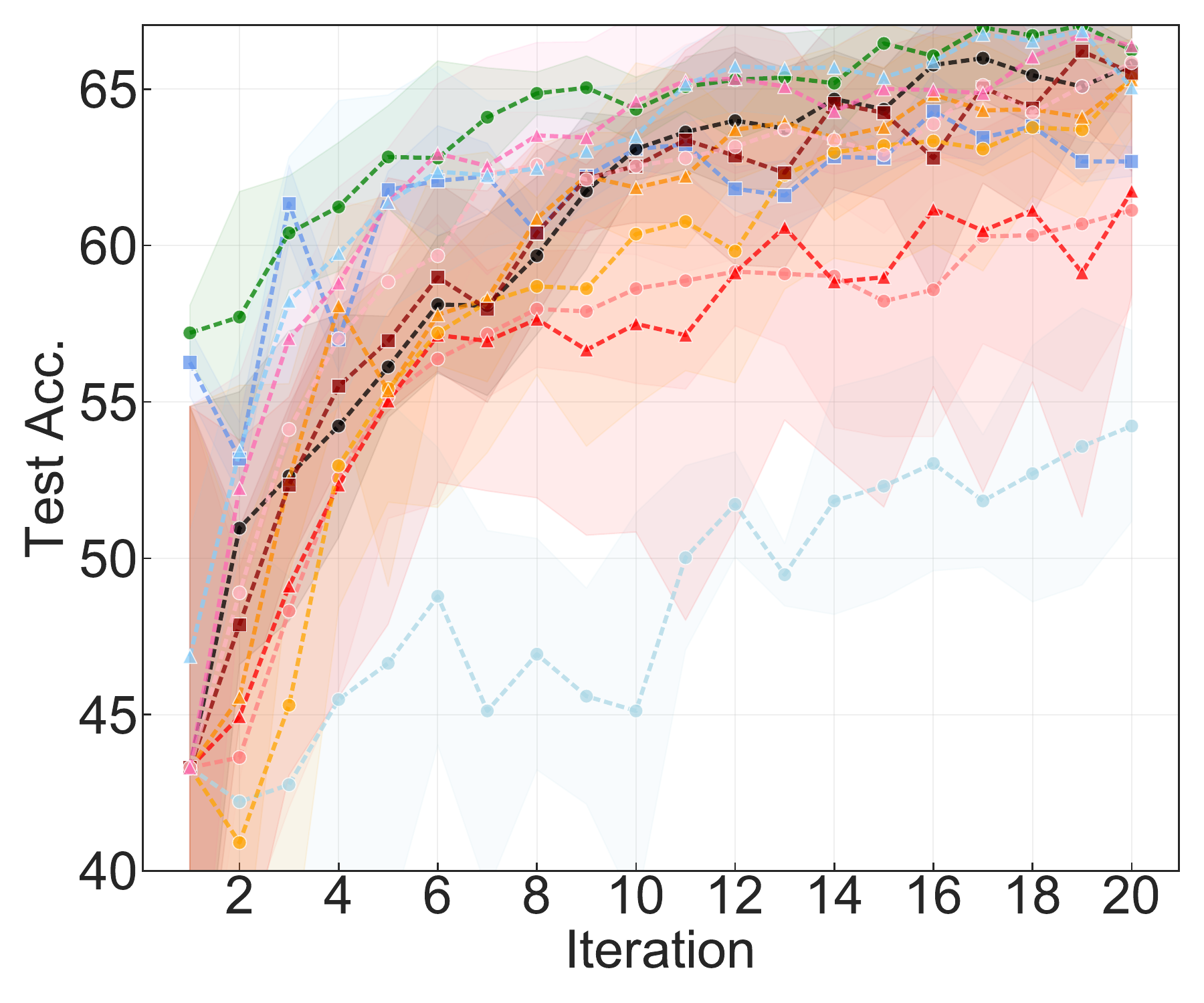}    
    \caption{Diabetic Retinopathy}
    \label{fig2diabetic}
\end{subfigure}
\begin{subfigure}{.245\textwidth}
    \centering
    \includegraphics[width=\linewidth]{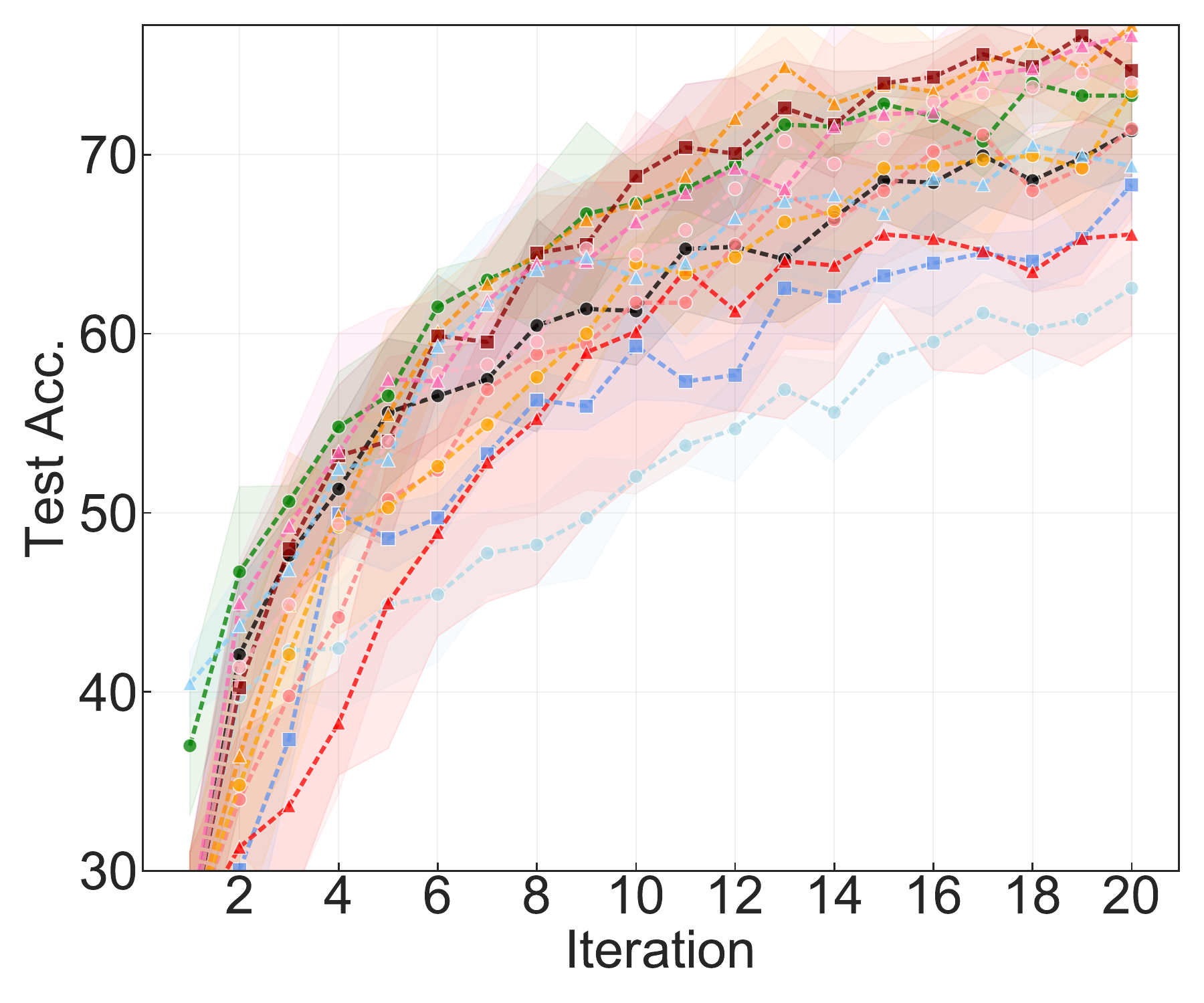} 
    \caption{IICBU Hela}
    \label{fig2iicbu}
\end{subfigure}
\begin{subfigure}{.245\textwidth}
    \centering
    \includegraphics[width=\linewidth]{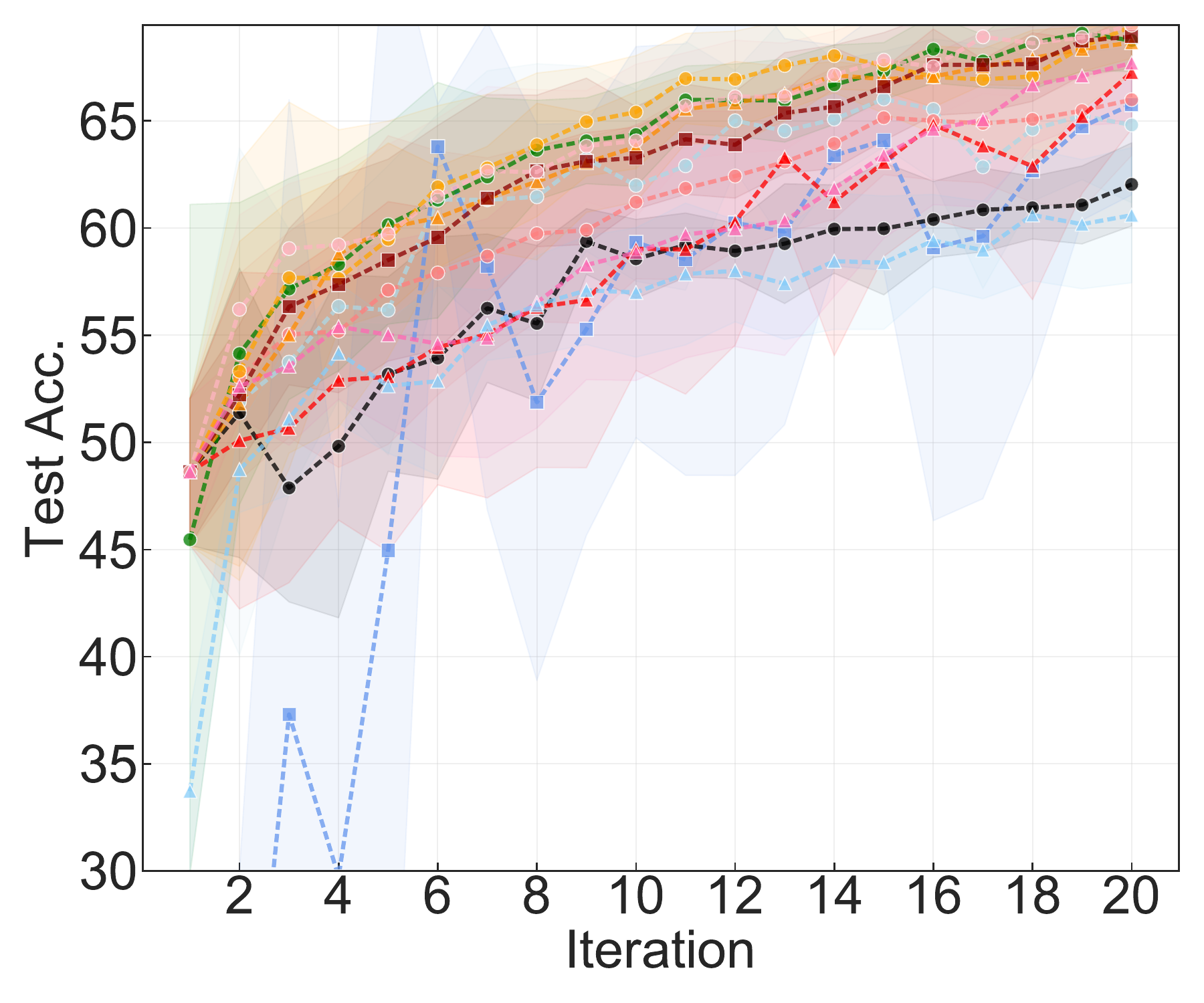} 
    \caption{HAM 10000}
    \label{fig2ham}
\end{subfigure}

\caption{(Top row) Results on fine-grained natural image classification tasks Stanford Cars \citep{stanfordcars}, FVGC Aircraft \citep{fvgcaircraft}, Oxford-IIIT Pets \citep{oxfordpets}, and the Places365 \citep{places365} datasets. (Bottom row) AL curves for biomedical datasets, including images of peripheral blood smears \citep{blood_smear}, retinal fundoscopy \citep{diabetic_retinopathy}, HeLa cell structures \citep{iicbu_hela}, and skin dermoscopy skin \citep{ham10000}, covering pathology, ophthalmology, cell biology, and dermatology domains using various imaging modalities. Additional biomedical datasets are explored in the Appendix.}
\label{fig:al_curves}
\end{figure}

Among the fine-grained natural image classification datasets, in Stanford Cars \citep{stanfordcars} and Oxford-IIIT Pets \citep{oxfordpets}, our \dropout outperforms all other AL queries in each iteration while also outperforming complex query approaches like \alfamix and \typiclust in the low-budget regime in FVGC Aircraft \citep{fvgcaircraft} (see Figure \ref{fig:al_curves}). Our approach, which is agnostic to dataset and model, outperforms the state-of-the-art AL queries, which often necessitate additional hyperparameter tuning given the underlying data distribution. We also test our method on a large-scale dataset with 365 classes, Places365 \citep{places365} (which contains approximately 1.8 million images), and our strategy beats all other modern AL queries. These results exemplify the scalability of our method on large datasets where AL would be used to probe the unlabeled data space for important samples.

\subsection{Out-of-domain datasets}
In addition, we conduct experiments on various biomedical datasets, which significantly deviate from the training distribution of the foundation models discussed in this study. The aim of these experiments is to underscore the effectiveness of AL queries when applied to datasets that are out-of-domain or underrepresented in pretraining. This kind of evaluation is especially pertinent in situations where a model trained in one institution or setting is deployed in another. An efficient AL query is helpful to obtain a minimum number of samples to label to fine-tune the model for the new task — a realistic scenario often neglected in previous studies. \ref{fig:al_curves} illustrates our strategy's performance on challenging out-of-domain data. In real-world contexts, \dropout excels as a practical, efficient method for querying hard-to-classify samples for improving the model.

\begin{figure}[!t]
\centering

\begin{subfigure}{.49\textwidth}
    \centering
    \includegraphics[trim={0 2cm 6cm 4cm},clip,width=\linewidth]{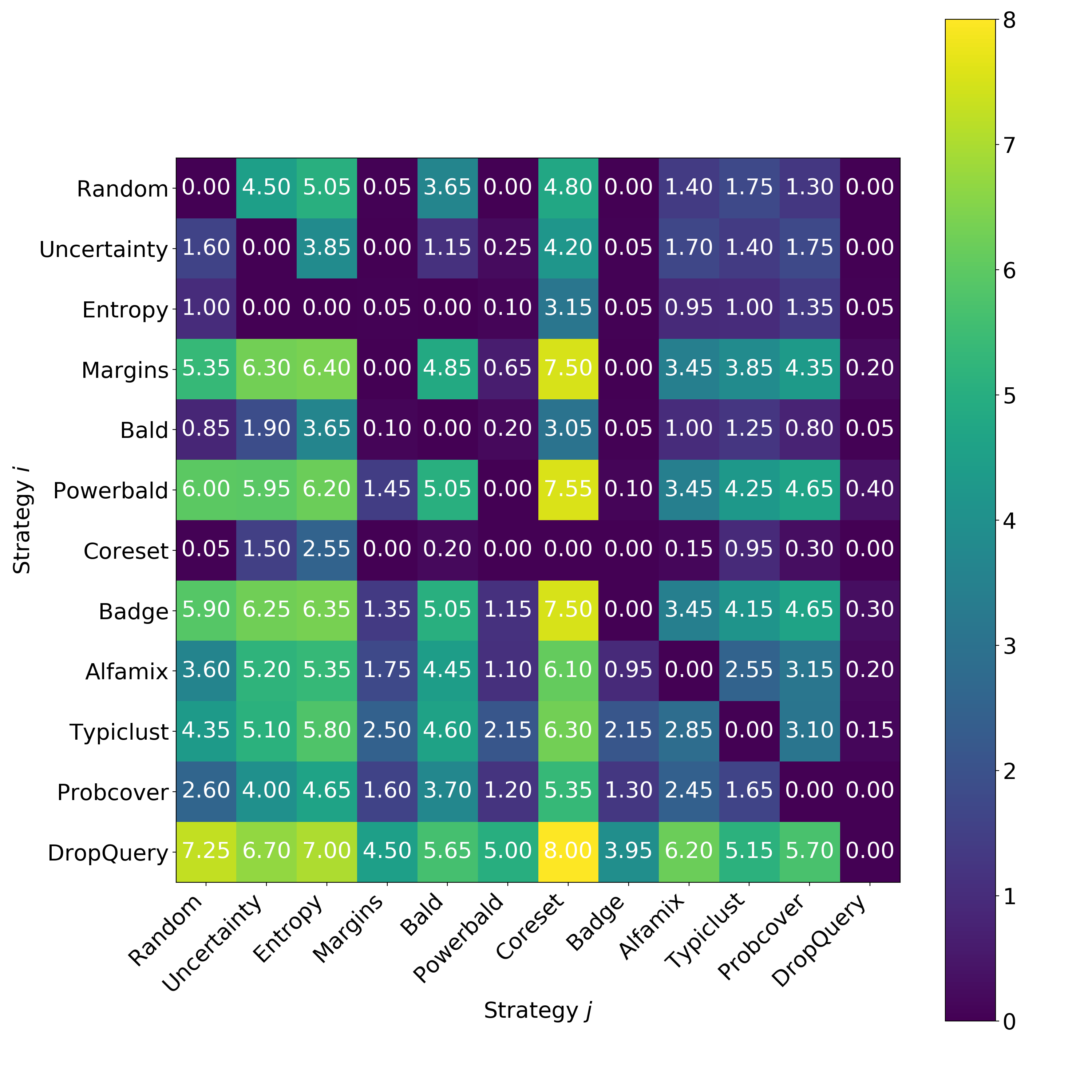}
    \caption{}
    \label{sig_a}
\end{subfigure}
\begin{subfigure}{.49\textwidth}
    \centering
    \includegraphics[trim={0 2cm 6cm 4cm},clip,width=\linewidth]{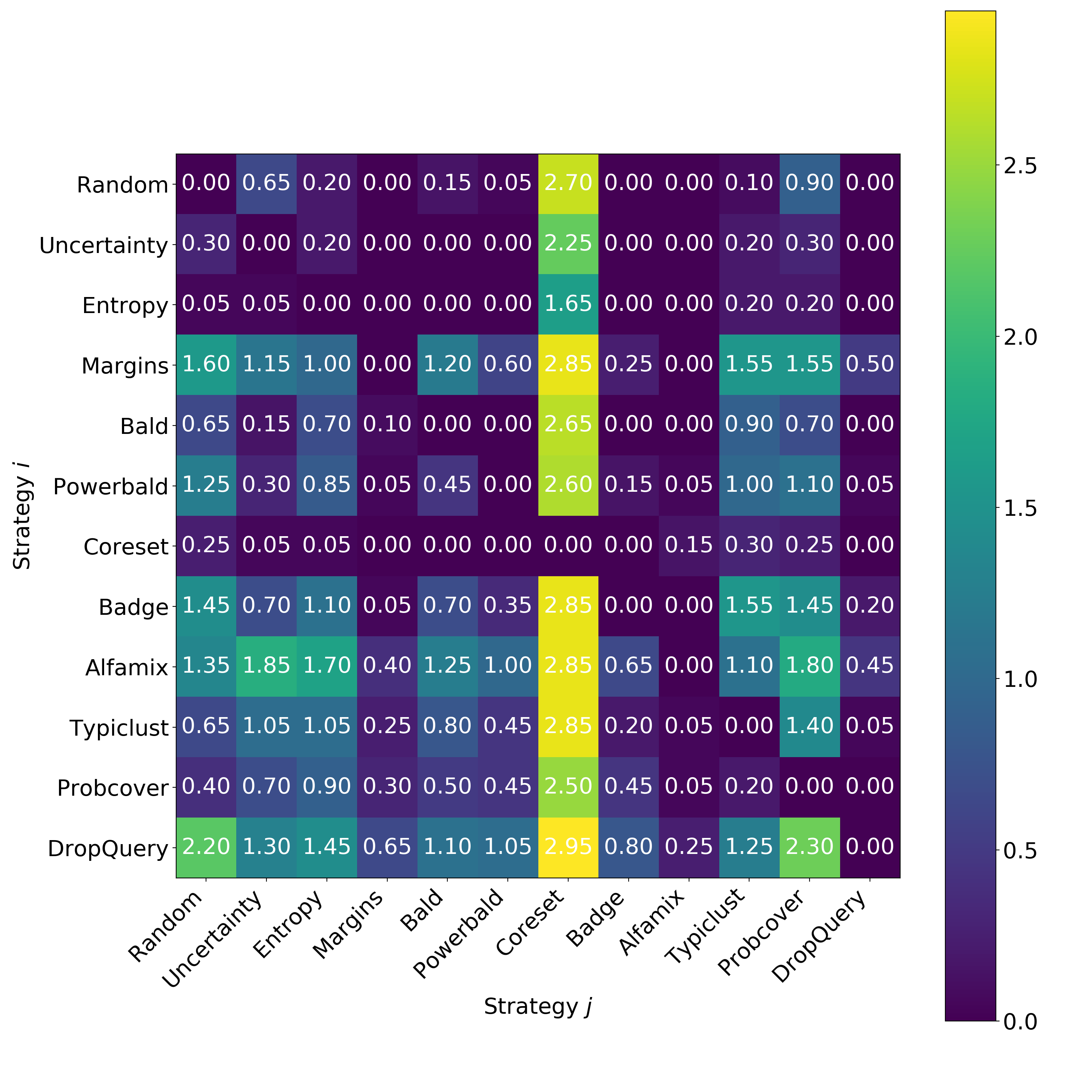}    
    \caption{}
    \label{sig_b}
\end{subfigure}

\caption{Win matrices for all AL strategies investigated in our study evaluated on natural image datasets and out-of-domain biomedical image datasets. \textbf{(a)} CIFAR100, Food101, Imagenet-100, and DomainNet-Real using DINOv2 VIT-g/14 features and Stanford Cars, FVGC Aircraft, Oxford-III Pets, and Places365 using OpenCLIP VIT-G/14 features (8 total settings). Due to computational costs, \probcover was not evaluated on Places365, so the max value of cells in the \probcover row/column is 7. \textbf{(b)} Blood Smear, Diabetic Retinopathy, IICBU Hela, and Skin cancer datasets using DINOv2 VIT-g/14 features (4 total settings). \dropout outperforms all other methods on the natural image datasets and is a strong competitor to all other methods on the biomedical image datasets with statistical significance.}
\label{fig:significance}
\end{figure}

\subsection{Statistical significance study}
In order to verify the empirical strength of \dropout, we furnish our results with statistical significance tests following the procedure in \cite{alfamix} and \cite{badge}. We conduct paired $t$-tests to determine whether a particular AL strategy $i$ outperforms another strategy $j$. We say that query strategy $i$ beats or surpasses strategy $j$ at iteration $r$ if 
$$c_{i,j}^r = \frac{\sqrt{5}\mu^r}{\sigma^r} > 2.776,\quad \mu^r = \frac{1}{5}\sum_{k=1}^5 (a_i^r - a_j^r),\quad \sigma^r = \sqrt{\frac{1}{5}\sum_{k=1}^5(a_i^r - a_j^r - \mu^r)^2}$$ 
where $a_i^r$ is the accuracy of strategy $i$ at iteration $r$. This corresponds to the $p = 0.05$ threshold for significance. We compute these $c_{i,j}^r$ for all iterations and sum the number of times strategy $i$ surpassed strategy $j$ at each iteration $r$ (divided by the total number of iterations) to indicate the score of $i$ beating $j$ in a specific dataset-model setting. Figure \ref{fig:significance} illustrates two “win” matrices illustrating the proportion of times strategy i has a significantly higher (p<0.05) mean accuracy compared to strategy j. For each pair, at each iteration, the mean accuracy for methods i and j are compared using a t-test with p<0.05, and the outcome is win, lose, or tie to yield a fraction of wins from (0, 1). This is repeated for all datasets considered and the values are summed to yield the proportion of wins. For example, in Figure \ref{sig_a}, \dropout beats random sampling in 7.25/8 fraction of total AL iterations across the 8 dataset settings.

\subsection{Ablations for \dropout}
\subsubsection{Impact of the dropout ratio}
We explore the influence of the dropout ratio on both the overall performance of the models and the number of selected candidates at each iteration step. We experiment with a range of values - [\textbf{0.15}, \textbf{0.30}, \textbf{0.45}, \textbf{0.60}] in addition to \textbf{0.75}, the setting used in our main experiments. Our findings, highlighted in Appendix Table 1, indicate that higher dropout ratios for the features result in slightly better-performing models, a consistent trend among datasets. This is perhaps unsurprising, as stronger regularization in the form of harsher dropout ratios would improve the generalizability of models in the low-budget regime. Interestingly, lower dropout ratios lead to fewer samples added to the candidate set. This may be a complementary factor affecting the performance of the model, as a smaller candidate set would likely be less diverse.

\subsubsection{Impact of $M$, the number of dropout iterations}
The number of dropout iterations, $M$, is a tunable hyper-parameter for our query function. We experiment with a range of values - [\textbf{5}, \textbf{7}, \textbf{9}] in addition to \textbf{3}, the value used in our main experiments, and present the results in Appendix Table 2. Our findings indicate that this hyper-parameter has a minimal impact on both the performance of the model and the number of selected candidates, with fewer iterations having a slight advantage. The robustness of our models' performance w.r.t $M$ is a desirable trait for an AL query function designed for the low-budget regime since the existence of a validation set for hyper-parameter tuning cannot be assumed due to the paucity of labels.

Over the course of our experiments, we found that other query functions could perform better with the right combination of hyperparameters. ProbCover performed better with lower purity levels, and ALFA-Mix performed better with a lower value of $\epsilon$. The maximum number of clusters was set arbitrarily for TypiClust, as was the number of neighbors used to calculate typicality. Ideally, these hyper-parameters would have been determined based on the distribution of features, the number of dimensions, and the number of samples in each dataset. These observations strengthen our belief that \dropout is a strong competitor to existing methods due to its simplicity and excellent performance without tuning $M$.
\vspace{-0.25em}
\section{Conclusions}
\vspace{-0.25em}
In this work, we systematically study four critical components of effective active learning in the context of foundation models. Given our observations, we propose a new AL strategy, named \dropout, that combines the benefits of strong initial pool selection by class centroids and a dropout-motivated uncertainty measure for querying unlabeled instances. This method outperforms other AL queries in a variety of natural image, biomedical, and large-scale datasets. It poses a paradigm shift to leveraging the properties of foundation model representations in AL. 
\vspace{-0.5em}
\paragraph{Limitations and Societal Impact}{Our work is a principled investigation focused on the interplay of AL and foundation models, and we recognize certain limitations. Firstly, our approach presumes the availability of public foundation models, which may not be universally accessible. Secondly, inherent biases and ethical dilemmas tied to these models could be reflected in our method, possibly leading to biased outcomes or the propagation of harmful stereotypes. Thirdly, our experiments focus on datasets with a relatively even distribution of labels, as has been the norm with most works exploring AL queries. Given that long-tailed datasets are quite common in the real world, there is a concern that our findings may not generalize to scenarios with heavily imbalanced class labels. This limitation warrants further investigation, particularly in the biomedical domain, where class imbalances and long-tailed distributions are widespread. Despite these issues, the growth of foundation models encourages their use in AL, even with out-of-domain datasets. Finally, we acknowledge that the semi-supervised learning method implemented in this study is not the only way to leverage unlabeled instances in this manner. Due to our experimental setup involving a fixed representation space generated by frozen foundation model backbones, we are constrained to a subset of SSL algorithms that operate in an offline mode. However, we encourage future work to investigate further along these lines and emphasize that our recommendation is limited to the use of a particular label propagation algorithm. We urge future research to address these concerns while considering the broader ethical implications.}

\newpage
\bibliography{main}
\bibliographystyle{tmlr}

\newpage
\appendix
\section{Appendix}

\subsection{Implementation details}

Our AL framework is implemented using PyTorch Lightning \citep{Falcon_PyTorch_Lightning_2019} for training and evaluation of the models and uses Faiss \citep{johnson2019billion} for GPU-accelerated clustering and nearest-neighbor searches. DINOv2 pre-trained models have been downloaded from \href{https://github.com/facebookresearch/dinov2}{https://github.com/facebookresearch/dinov2} and OpenCLIP pre-trained models are loaded using \href{https://github.com/mlfoundations/open_clip}{https://github.com/mlfoundations/open\_clip}. Our implementation is available at \href{https://github.com/sanketx/AL-foundation-models}{https://github.com/sanketx/AL-foundation-models}.

We conduct AL experiments for 20 iterations with 5 seeds - 1,10,100,1000,10000 for each combination of query, dataset, and model that we evaluate and report the mean accuracy averaged over all 5 seeds. The query budget at each iteration is set to $C$ where C is the number of classes in the dataset. All linear classifiers in our experiments are trained using the AdamW optimizer with a learning rate of 1e-2, weight decay of 1e-2, and dropout with $\rho=0.75$. In experiments involving label propagation, we set $\alpha=0.9$ and construct the graph using the 500 nearest neighbors.

Computations are carried out on A100 (40GB) GPUs, however we note that GPUs with 16-24GB of memory will also suffice for the majority of our experiments, with each individual run of 20 iterations taking approximately 15-30 minutes.

\subsubsection{AL query implementations}

All the AL queries tested in our experiments have been reimplemented to optimize them for compute and memory efficiency. BALD and PowerBALD both use 20 MC sampling iterations for uncertainty estimation. Core-Set is implemented using the greedy k-center approach. ALFA-Mix is implemented using the closed-form approximation of $
\alpha$ with $\epsilon=\frac{0.2}{\sqrt{D}}$ where $D$ is the dimensionality of $\alpha$. For TypiClust, we set the maximum number of clusters to 500 for all datasets except Places365 where we set it to 1000. Typicality is calculated using the 20 nearest neighbors. $\delta$ in ProbCover is estimated with a purity threshold of 0.95.

BADGE is difficult to scale to large datasets such as Places365 due to the size of the gradient embedding vectors. They have $N\times C \times D$ elements where $N$ is the number of unlabeled samples, $C$ is the number of classes, and $D$ is the dimensionality of the output of the penultimate layer, which in our case is feature vector derived from the foundation model. Fortunately, the K-Means++ initialization scheme used to pick diverse points only requires the computation of the squared distance between pairs of gradient embeddings, so the full embeddings themselves need not be computed.

The squared distance between a pair of gradient embeddings $G_i,G_j$ can be expressed in terms of their squared Frobenius norm $||G_i - G_j||_F^2$ where $G_i = Z_iP_i^T$. Here, $Z_i$ is the $D$ dimensional feature vector input to the linear classifier and $P_i = (p_k^{(i)} - I(\hat{y}^{(i)}=k))_{k=1}^C$, the difference of the predicted probability vector and the one hot encoded pseudo-label of the $i$th sample.

$||G_i - G_j||_F^2$ can be efficiently factorized as $Z_i^T(P_i^TP_i)Z_i + Z_j^T(P_j^TP_j)Z_j - 2Z_i^T(P_i^TP_j)Z_j$ which only requires $O(N\cdot(C+D))$ space instead of $O(N\cdot C\cdot D)$, enabling BADGE to scale to datasets with several million unlabeled examples.

\subsection{Utilizing unlabeled instances for improving the active learner}

In this section, we explore leveraging unlabeled instances in a semi-supervised fashion as a complementary approach to active learning. Prior art such as variants of \typiclust and \probcover use Flex-Match \citep{flexmatch}, a state-of-the-art semi-supervised learning algorithm based on consistency among augmented views of an image. However, this is challenging to implement in our context as repeated forward passes through the feature extractor would be computationally expensive. Since foundation models are typically trained in a self supervised fashion using a contrastive loss (or a variant) to encourage consistency, augmented views would map to similar points in the feature space.

Several works have suggested using state-of-the-art semi-supervised frameworks for label propagation from labeled instances \citep{consistency, rethinking}. We adopt an offline variant of a transductive label propagation method \citep{labelprop}. Our experimental setup accounted for the confidence of the propagated labels while training the classifier. We used equation 11 from \citep{labelprop} to weight each sample by using entropy as a measure of confidence. The weight was computed as $w_i = 1 - \frac{H(\hat{z}_i)}{log(c)}$. Since our representation space does not change, we only perform a cycle of label propagation when new labels are added to the pool, i.e. after each AL query. In theory, a semantically meaningful representation space from a Foundation Model would enable labels to propagate more effectively, thus increasing the efficacy of the active learner. However, our experimental results say otherwise.

\begin{figure*}[!t]
\centering
\begin{subfigure}{\textwidth}
    \centering
    \includegraphics[width=\linewidth]{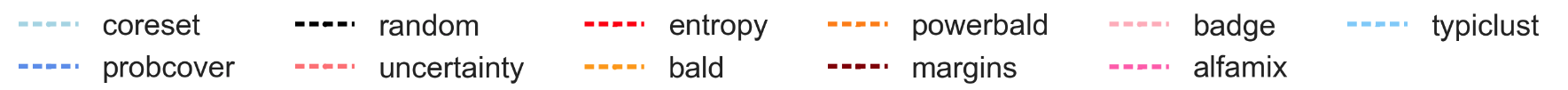}
\end{subfigure}

\begin{subfigure}{.245\textwidth}
    \centering
    \includegraphics[width=\linewidth]{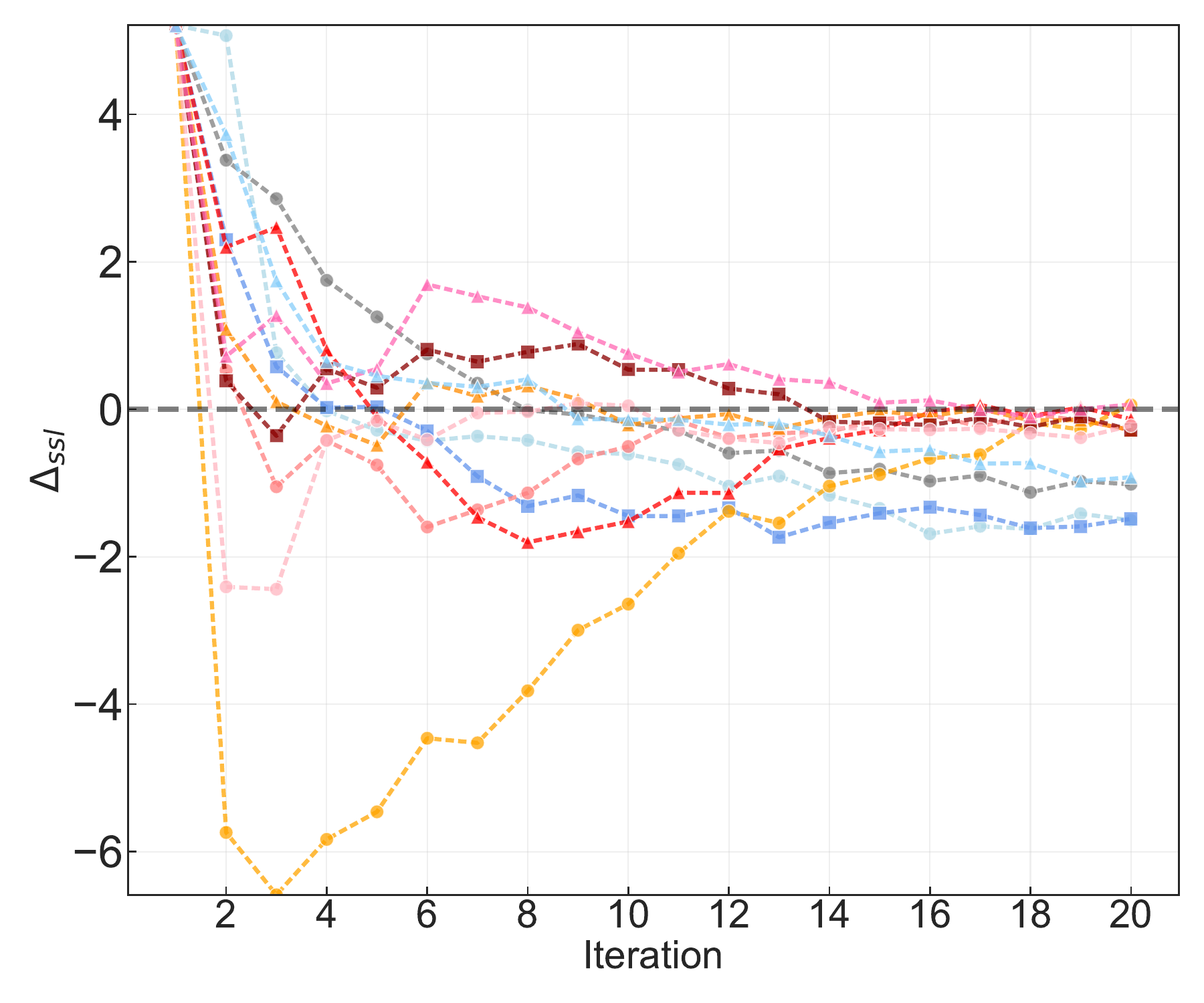}
    \caption{CIFAR100}
    \label{SUBFIGURE LABEL 1}
\end{subfigure}
\begin{subfigure}{.245\textwidth}
    \centering
    \includegraphics[width=\linewidth]{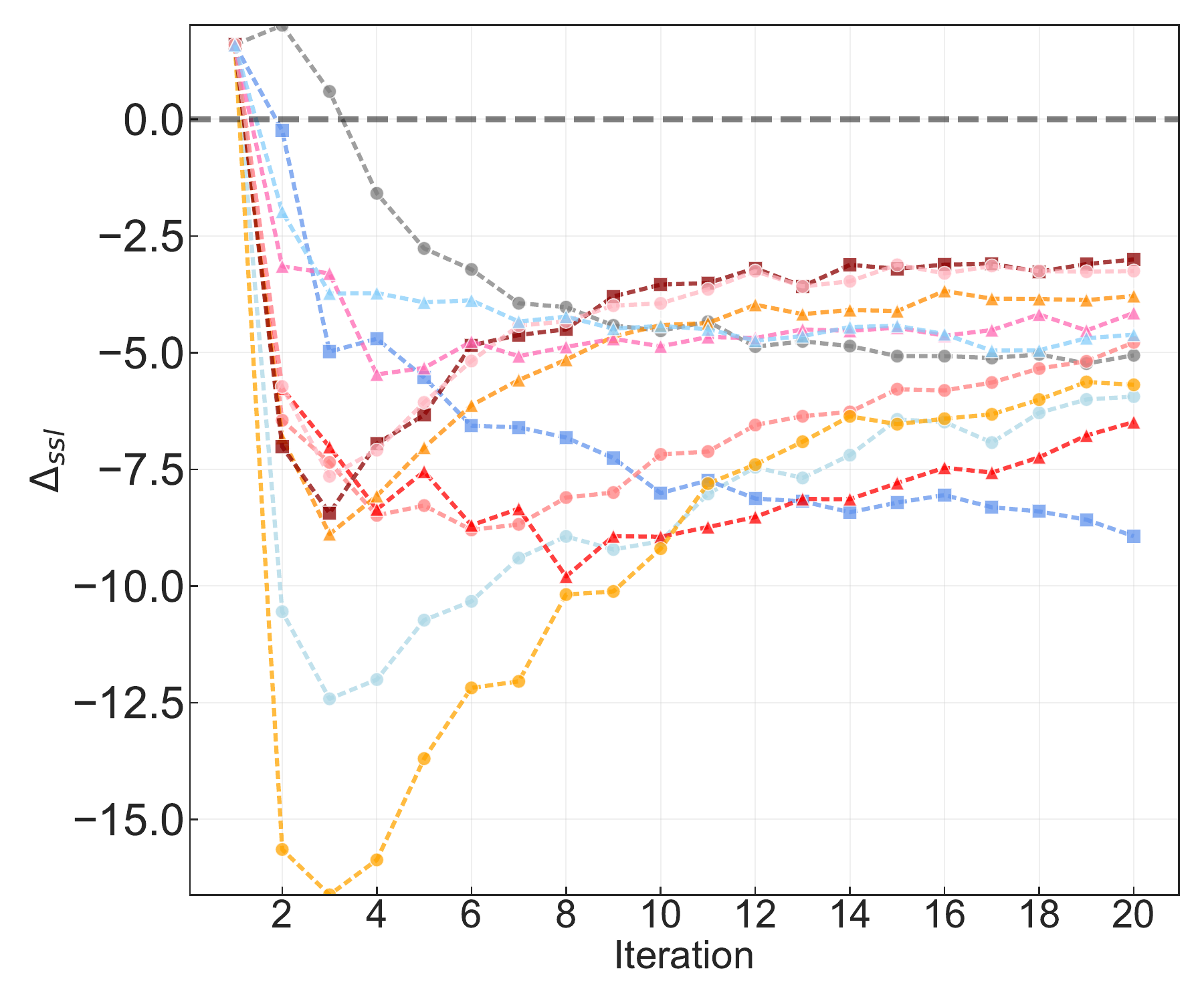}    
    \caption{Food101}
    \label{SUBFIGURE LABEL 2}
\end{subfigure}
\begin{subfigure}{.245\textwidth}
    \centering
    \includegraphics[width=\linewidth]{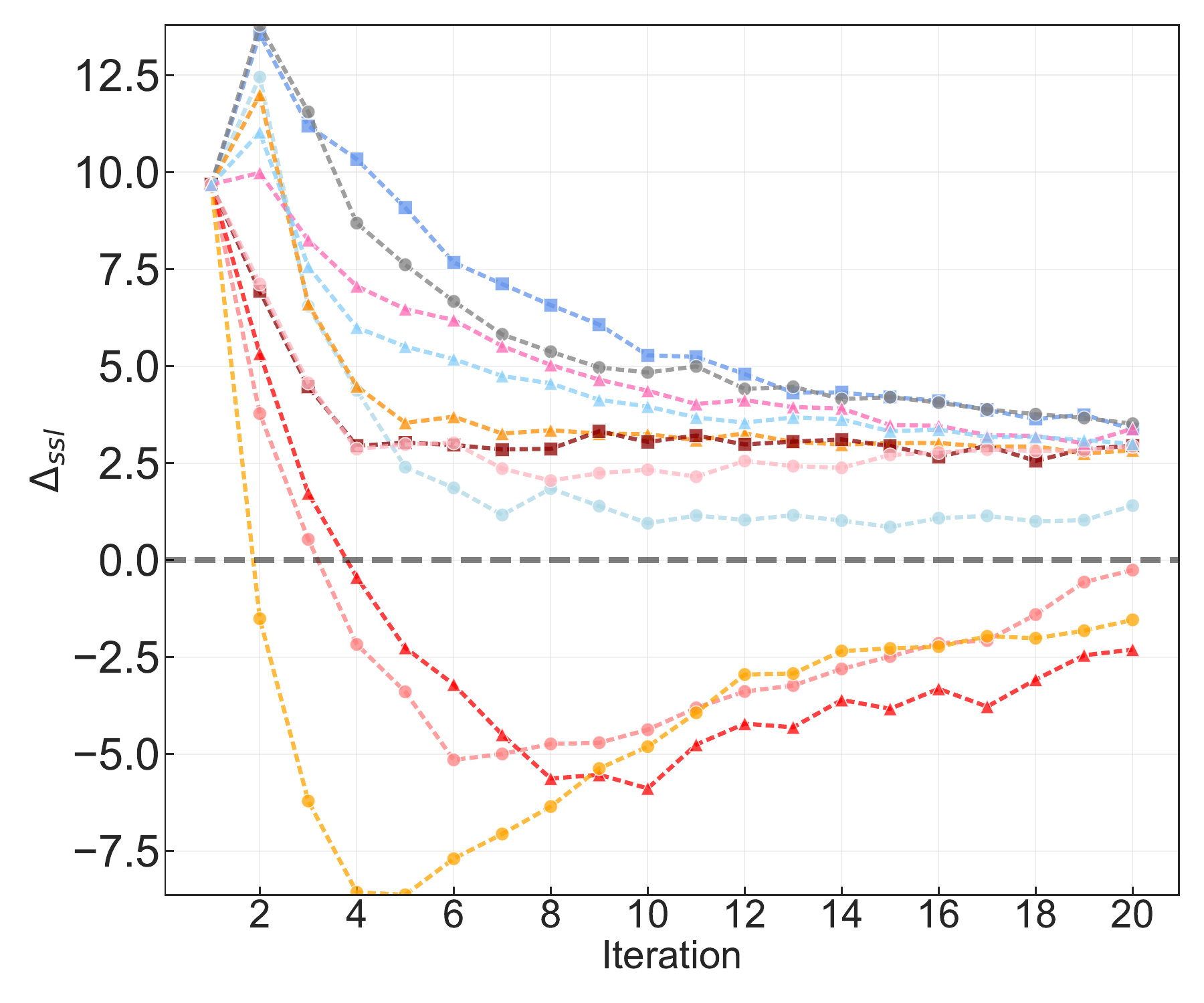} 
    \caption{ImageNet-100}
    \label{SUBFIGURE LABEL 4}
\end{subfigure}
\begin{subfigure}{.245\textwidth}
    \centering
    \includegraphics[width=\linewidth]{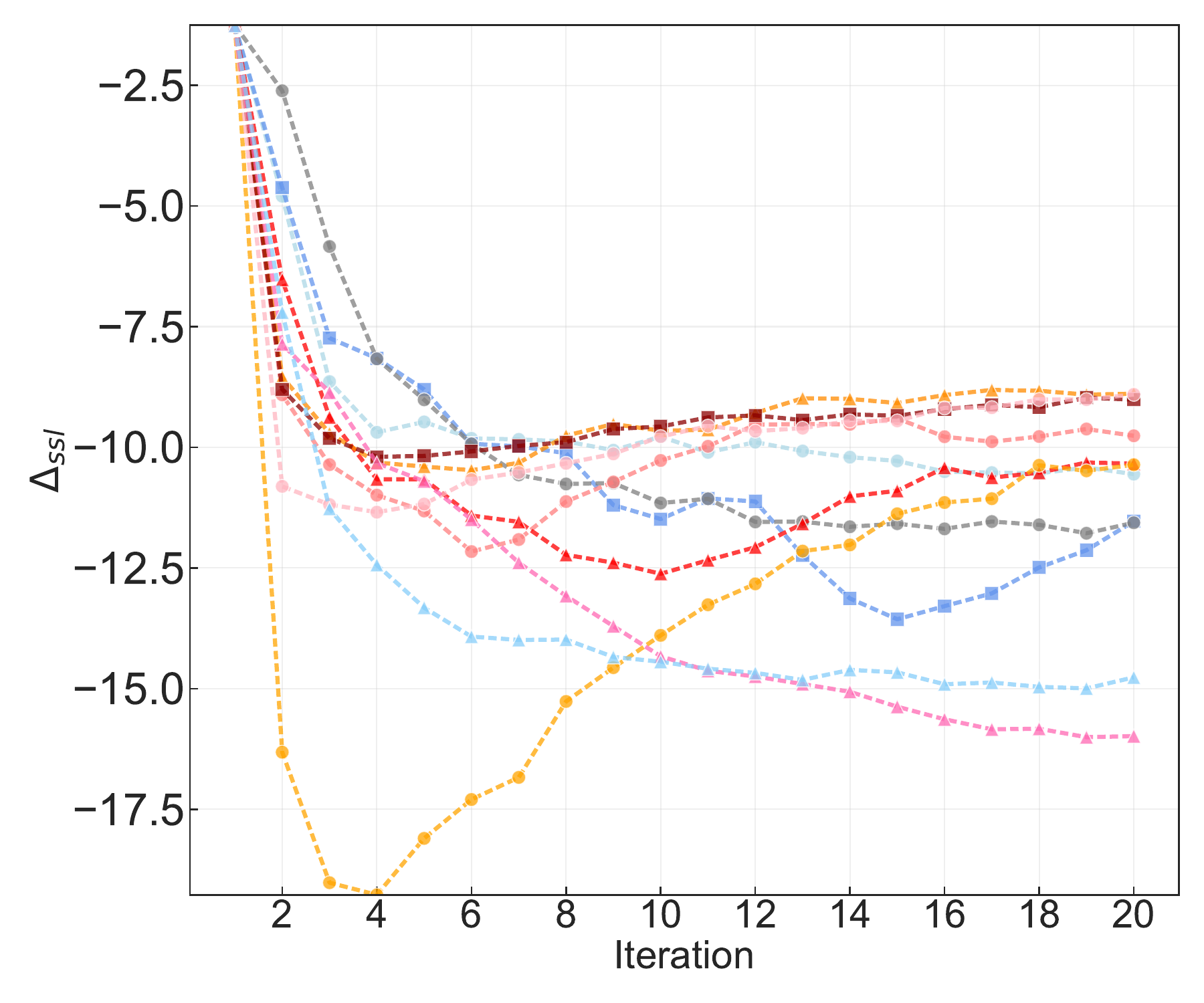} 
    \caption{DomainNet-Real}
    \label{SUBFIGURE LABEL 4}
\end{subfigure}
\caption{We illustrate the performance difference $\Delta_{ssl}$ between AL with and without label propagation for unlabeled instances. The results, averaged over 5 runs of 20 AL iterations on 4 natural image datasets, show that the suitability of foundation models for pseudo-label approaches is, although significant in the initial iterations of AL, hurts the performance of the active learner in later iterations.}
\label{fig:delta_ssl}
\end{figure*}

We show the effects of label propagation in Figure \ref{fig:delta_ssl} while using a randomly initialized labeled pool for all queries. We find that the efficacy of semi-supervised learning in this setting is questionable at best, with wide variation across query methods and datasets. While we do see an initial boost in performance, contrary to \citep{consistency, rethinking}, the gap quickly narrows and label propagation under performs the reference query. Propagating labels from uncertain queried samples may cause points across the decision boundary to be assigned to incorrect classes, hampering performance in the long run.

\subsection{Full results on natural image and out-of-domain data}
\begin{figure}[]
    \centering
    
\begin{subfigure}{\textwidth}
    \centering
    \includegraphics[trim={0 6cm 0 0},clip,width=\linewidth]{figures/al_curves_legend.png}
\end{subfigure}

    \begin{subfigure}{0.35\textwidth}
        \centering
        \includegraphics[width=\textwidth]{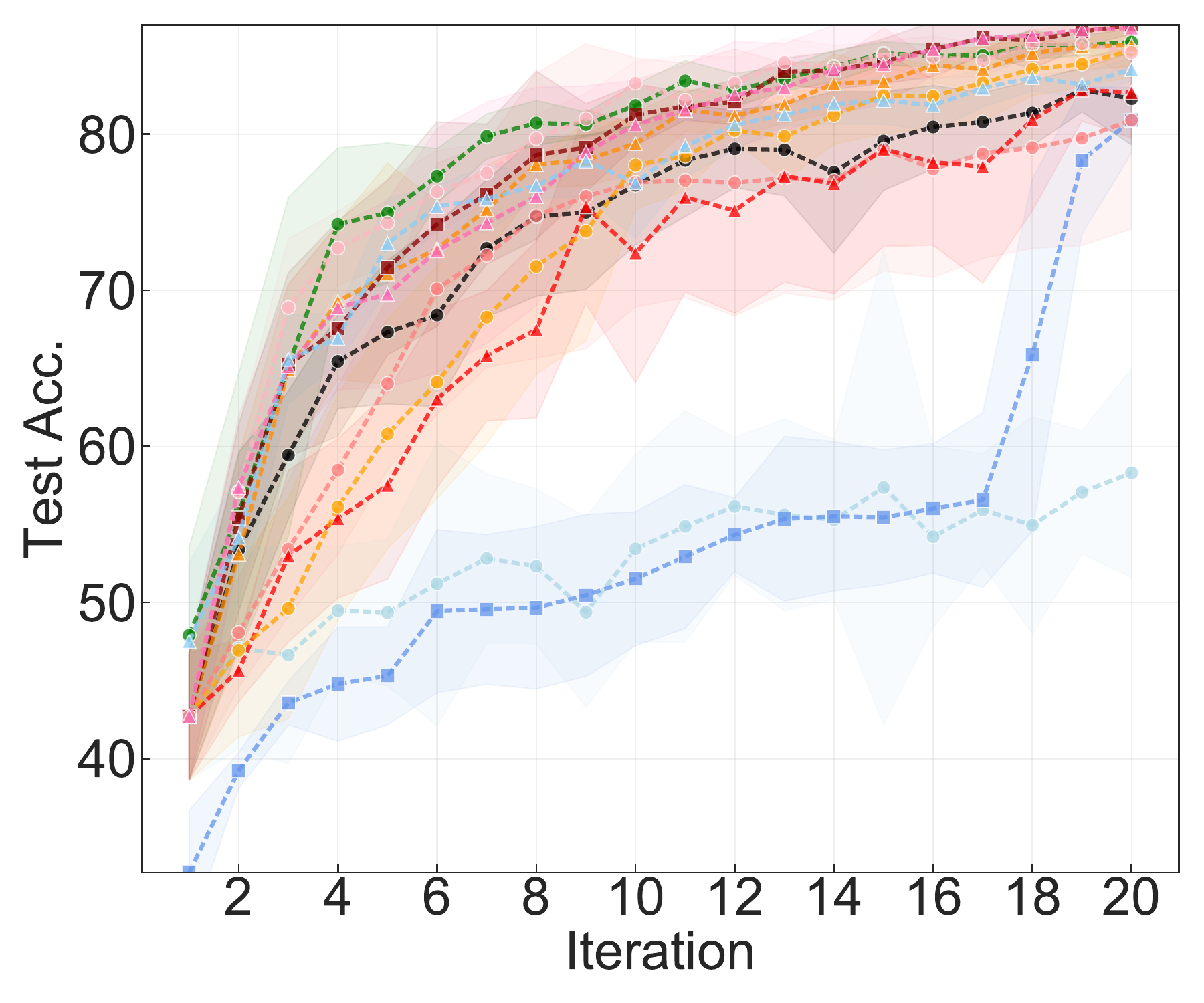}
        \caption{Colorectal Histology}
    \end{subfigure}
    \begin{subfigure}{0.35\textwidth}
        \centering
        \includegraphics[width=\textwidth]{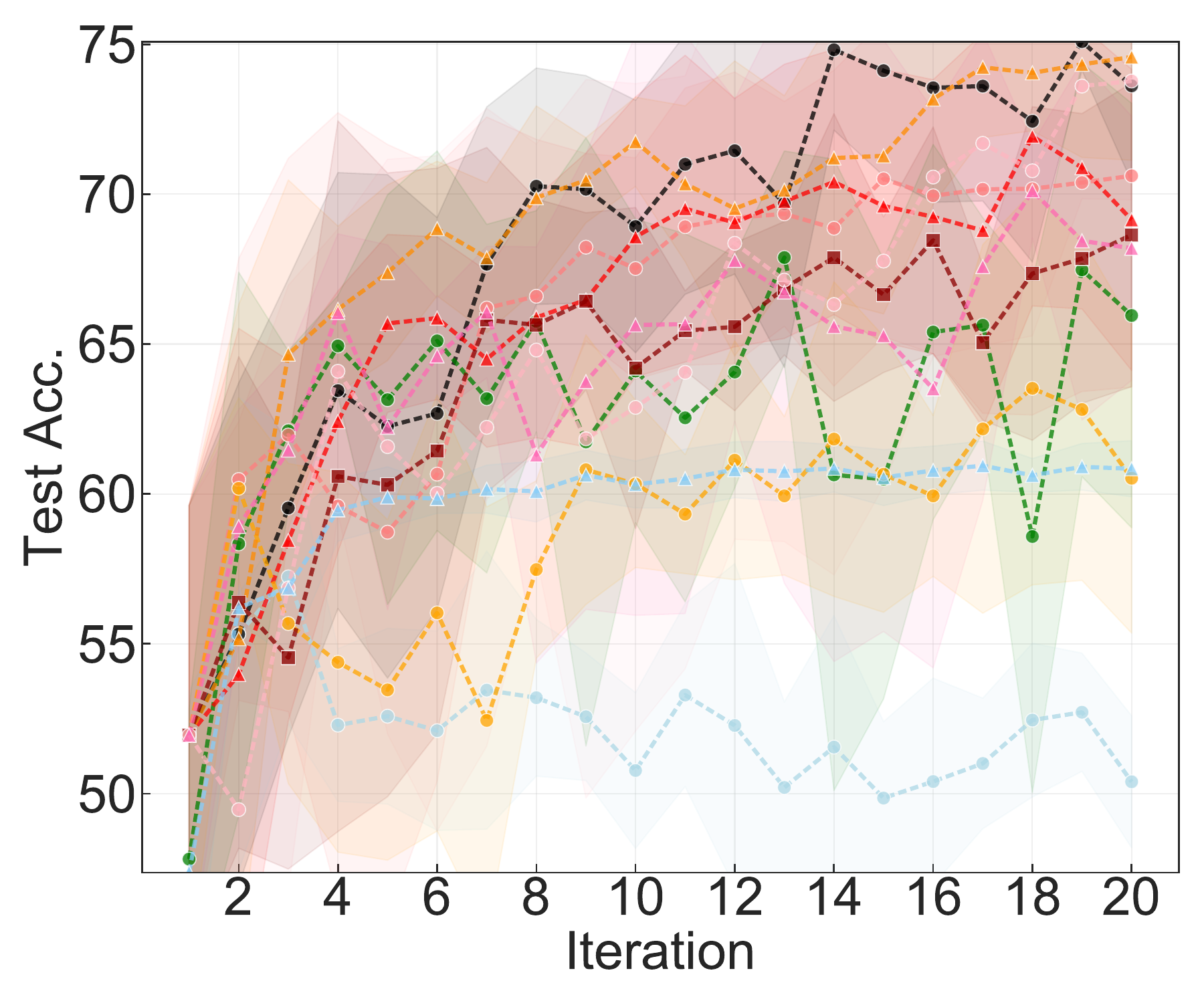}
        \caption{Patch Camelyon}
    \end{subfigure}
    \caption{Full AL curves for additional out-of-domain biomedical datasets.}
    \label{fig:bio_al_curves_dino_vit_g14}
\end{figure}

We also tabulate full AL empirical results in Table \ref{tab:full_results_nat1}, Table \ref{tab:full_results_nat2}, and Table \ref{tab:full_results_bio} for the datasets presented in the main manuscript (12 in total) and additional out-of-domain datasets like colorectal histology \citep{colorectal_histology} and Patch Camelyon \citep{patch_camelyon_1, patch_camelyon_2} (Figure \ref{fig:bio_al_curves_dino_vit_g14}). We repeated acquisition trajectories with 5 different random seeds for precise evaluation and report mean accuracy and standard deviation on the held out test set. \dropout performs competitively on the colorectal histology dataset, but underperforms others on the Patch Camelyon dataset. The abnormal spread in performances of AL strategies on Patch Camelyon suggests that Foundation Model features may not equally benefit all approaches on out-of-distribution data, and warrants further investigation of low-budget AL in biomedical imaging.

Biomedical datasets are ideal for testing active learning methods, given the time constraints and high costs of labeling by domain experts. Acevedo et al 2020's blood smear dataset \citep{blood_smear} consists of brightfield microscopy images of eight peripheral blood cell classes (imbalanced classes, minority/majority ratio = 0.36). The colorectal histology dataset \citep{colorectal_histology} includes H\&E stained brightfield microscopy images of 8 different classes of textures seen in colorectal cancer histology (balanced, min/Maj ratio = 1.0). The Kaggle Diabetic Retinopathy challenge dataset \citep{diabetic_retinopathy} presents retinal images categorized into five diabetic retinopathy severity levels (imbalanced, min/Maj ratio = 0.19). The Robert Murphy lab's IICBU 2008 HeLa dataset \citep{iicbu_hela} includes fluorescence microscopy images of HeLa cells, with ten different classes of labeled subcellular structures (min/Maj ratio = 0.74). The HAM10000 dataset \citep{ham10000} contains dermatoscopic images across seven skin lesion classes (imbalanced, min/Maj ratio = 0.01). Lastly, the patch camelyon dataset \citep{patch_camelyon_1} \citep{patch_camelyon_2} consists of 327,680 image patches from lymph notes with the goal of binary classification of the presence or absence of metastatic breast carcinoma cells (balanced, min/Maj ratio = 1.0). The patch camelyon train/val/test splits are 262,144/ 32,768/ 32,768 respectively. Together, these datasets span fields such as cell biology, cytology, dermatology, and ophthalmology, offering a robust out-of-distribution test for active learning strategies that utilize foundation models pre-trained on natural images.

\subsection{Additional pretraining strategies}

We study the impact of limited pretraining and a relatively weak feature extractor on the performance of leading AL strategies. In this experiment, we explore the use of Masked Autoencoders (MAE) \citep{he2022masked} pre-trained on ImageNet-1K \citep{deng2009imagenet} as the backbone model, specifically the largest variant: ViT-H/14. ImageNet is a considerably smaller dataset compared to LVD-142M used to train DINOv2 or LAION-2B used to train OpenCLIP models and is limited in terms of the diversity of images. Representations learned with limited pretraining data may not result in well structured latent spaces like those induced by foundation models trained on hundreds of millions of images, adversely impacting the performance of AL strategies which require robust representations. Furthermore, the MAE backbone is trained using a patch-level reconstruction objective instead of a contrastive method. This encourages the model to learn strong local features, but suboptimal global features which are necessary for instance discrimination or classification tasks such as the ones studied in our experimental setting \citep{huang2023contrastive}.

We analyze the best performing AL strategies in our previous experiments, \badge, \alfamix, \typiclust, \margins, and \powerbald and compare them with \dropout and a random sampling baseline on the CIFAR100, Food101, ImageNet-100, and DomainNet-Real datasets. We use the same experimental conditions and hyper-parameter settings as our previous experiments and report the results in Table \ref{tab:mae_result}. We observe a clear trend across datasets: AL strategies which rely on clustering images in the representation space are outperformed by those which are independent of the underlying latent structure. \badge performs well consistently, followed by \powerbald and \margins, while \typiclust, \alfamix, and \dropout demonstrate weaker performance. 

Unlike our experiments with foundation models, we do not see a significant impact of intelligent initial pool selection to overcome the cold start problem, and any advantage that \typiclust and \dropout may have is quickly eroded. Furthermore, we observe that the MAE features do not generalize well to CIFAR100 and Food101, with even the best AL methods not performing much better than random sampling. This is likely because ImageNet-100 is a subset of the MAE pretraining dataset and is compositionally similar to DomainNet-Real. Given that the MAE backbone typically requires domain-specific fine-tuning, it is unsurprising that these features perform poorly in the extremely low-budget AL regime, especially with a simple linear classifier. We conclude by re-iterating our fundamental premise: in the era of vision foundation models, we need to revisit previous findings in active learning to advance the development of strategies that can fully leverage the rich representations generated by these powerful backbones.

\begin{table}[h]
\centering
\tiny
\caption{Impact of the dropout ratio. We evaluate 5 different dropout ratios $\rho$ for on 4 natural image datasets - CIFAR100, Food101, ImageNet-100, and DomainNet-Real. We report the mean accuracy averaged over 5 runs along with the standard deviation at each AL iteration. \textbf{Bold} values represent the \textbf{first place} mean accuracy at iteration $t$ with the \underline{second place} value \underline{underlined}. We also report the corresponding fraction of the unlabeled examples added to the candidate set.}
\vspace{0.5em}
\begin{adjustbox}{width=1\textwidth}
\begin{tabular}{c|cc|cc|cc|cc|cc}
\toprule
 $t$ & \multicolumn{2}{c}{$\rho = 0.15$} & \multicolumn{2}{c}{$\rho = 0.3$} & \multicolumn{2}{c}{$\rho = 0.45$} & \multicolumn{2}{c}{$\rho = 0.6$} & \multicolumn{2}{c}{$\rho = 0.75$} \\ 
& Mean ± Std & Fraction & Mean ± Std & Fraction & Mean ± Std & Fraction & Mean ± Std & Fraction & Mean ± Std &Fraction \\
\midrule
\multicolumn{11}{c}{CIFAR100} \\
\midrule
1 & $\textbf{72.93} \pm 1.93$ & $0.0\%$ & $\underline{72.92} \pm 1.9$ & $0.0\%$ & $72.84 \pm 2.06$ & $0.0\%$ & $72.81 \pm 2.1$ & $0.0\%$ & $72.58 \pm 2.18$ & $0.0\%$ \\   
2 & $82.46 \pm 0.59$ & $4.05\%$ & $82.15 \pm 1.33$ & $6.7\%$ & $83.0 \pm 1.03$ & $9.9\%$ & $\textbf{83.56} \pm 1.31$ & $14.53\%$ & $\underline{83.52} \pm 0.76$ & $21.53\%$ \\   
3 & $85.13 \pm 0.48$ & $2.09\%$ & $85.4 \pm 1.18$ & $3.51\%$ & $85.49 \pm 0.69$ & $5.16\%$ & $\underline{85.88} \pm 0.94$ & $7.84\%$ & $\textbf{86.6} \pm 0.78$ & $11.7\%$ \\   
4 & $86.96 \pm 0.56$ & $1.76\%$ & $86.68 \pm 0.94$ & $2.84\%$ & $\underline{87.33} \pm 0.53$ & $4.25\%$ & $87.15 \pm 0.82$ & $6.14\%$ & $\textbf{87.67} \pm 0.57$ & $9.22\%$ \\   
5 & $87.81 \pm 0.44$ & $1.55\%$ & $88.07 \pm 0.64$ & $2.59\%$ & $\underline{88.43} \pm 0.4$ & $3.62\%$ & $88.02 \pm 0.75$ & $5.48\%$ & $\textbf{88.94} \pm 0.31$ & $8.33\%$ \\   
6 & $88.53 \pm 0.39$ & $1.49\%$ & $88.67 \pm 0.41$ & $2.41\%$ & $\underline{89.28} \pm 0.22$ & $3.39\%$ & $89.14 \pm 0.16$ & $5.11\%$ & $\textbf{89.35} \pm 0.16$ & $7.72\%$ \\   
7 & $88.99 \pm 0.15$ & $1.46\%$ & $89.07 \pm 0.41$ & $2.32\%$ & $\underline{89.59} \pm 0.08$ & $3.2\%$ & $89.54 \pm 0.17$ & $4.84\%$ & $\textbf{89.65} \pm 0.31$ & $7.37\%$ \\   
8 & $89.58 \pm 0.2$ & $1.42\%$ & $89.37 \pm 0.58$ & $2.18\%$ & $\underline{89.89} \pm 0.22$ & $3.04\%$ & $89.88 \pm 0.29$ & $4.6\%$ & $\textbf{90.12} \pm 0.14$ & $7.07\%$ \\   
9 & $89.76 \pm 0.22$ & $1.37\%$ & $89.82 \pm 0.49$ & $2.21\%$ & $\textbf{90.26} \pm 0.24$ & $3.03\%$ & $90.12 \pm 0.17$ & $4.41\%$ & $\underline{90.23} \pm 0.21$ & $6.88\%$ \\   
10 & $90.18 \pm 0.28$ & $1.31\%$ & $90.14 \pm 0.33$ & $2.09\%$ & $\underline{90.41} \pm 0.1$ & $2.92\%$ & $90.38 \pm 0.2$ & $4.36\%$ & $\textbf{90.48} \pm 0.22$ & $6.75\%$ \\   
11 & $90.24 \pm 0.22$ & $1.25\%$ & $90.25 \pm 0.27$ & $2.06\%$ & $\underline{90.62} \pm 0.1$ & $2.9\%$ & $\textbf{90.74} \pm 0.21$ & $4.16\%$ & $90.6 \pm 0.28$ & $6.68\%$ \\   
12 & $90.4 \pm 0.33$ & $1.24\%$ & $90.58 \pm 0.25$ & $2.03\%$ & $90.71 \pm 0.22$ & $2.84\%$ & $\textbf{90.86} \pm 0.26$ & $4.12\%$ & $\underline{90.8} \pm 0.27$ & $6.54\%$ \\   
13 & $90.58 \pm 0.2$ & $1.23\%$ & $90.81 \pm 0.22$ & $2.03\%$ & $90.91 \pm 0.14$ & $2.82\%$ & $\underline{90.94} \pm 0.24$ & $4.05\%$ & $\textbf{90.99} \pm 0.2$ & $6.37\%$ \\   
14 & $90.69 \pm 0.27$ & $1.2\%$ & $90.97 \pm 0.25$ & $1.93\%$ & $91.03 \pm 0.15$ & $2.77\%$ & $\underline{91.17} \pm 0.2$ & $4.0\%$ & $\textbf{91.26} \pm 0.25$ & $6.35\%$ \\   
15 & $90.92 \pm 0.25$ & $1.18\%$ & $91.13 \pm 0.26$ & $1.92\%$ & $\underline{91.2} \pm 0.18$ & $2.66\%$ & $91.13 \pm 0.19$ & $3.99\%$ & $\textbf{91.32} \pm 0.26$ & $6.27\%$ \\   
16 & $91.3 \pm 0.1$ & $1.19\%$ & $91.28 \pm 0.18$ & $1.86\%$ & $91.28 \pm 0.15$ & $2.69\%$ & $\underline{91.42} \pm 0.23$ & $3.97\%$ & $\textbf{91.47} \pm 0.16$ & $6.0\%$ \\      

\midrule
\multicolumn{11}{c}{Food101} \\
 \midrule
1 & $70.83 \pm 1.2$ & $0.0\%$ & $70.86 \pm 1.17$ & $0.0\%$ & $\underline{71.29} \pm 1.16$ & $0.0\%$ & $71.28 \pm 1.36$ & $0.0\%$ & $\textbf{71.33} \pm 1.38$ & $0.0\%$ \\   
2 & $74.24 \pm 2.46$ & $1.45\%$ & $\underline{75.18} \pm 1.64$ & $2.48\%$ & $75.16 \pm 1.72$ & $3.72\%$ & $74.8 \pm 1.75$ & $5.54\%$ & $\textbf{76.09} \pm 1.34$ & $8.49\%$ \\   
3 & $77.6 \pm 1.08$ & $1.61\%$ & $78.52 \pm 1.38$ & $2.73\%$ & $78.42 \pm 1.43$ & $3.88\%$ & $\underline{78.61} \pm 1.48$ & $5.87\%$ & $\textbf{79.76} \pm 0.88$ & $8.6\%$ \\   
4 & $79.98 \pm 1.2$ & $1.65\%$ & $\underline{81.35} \pm 1.25$ & $2.79\%$ & $80.83 \pm 1.19$ & $3.99\%$ & $81.32 \pm 0.91$ & $5.79\%$ & $\textbf{82.24} \pm 0.52$ & $8.75\%$ \\   
5 & $82.43 \pm 0.65$ & $1.75\%$ & $82.85 \pm 0.58$ & $2.84\%$ & $82.8 \pm 0.56$ & $4.06\%$ & $\underline{83.12} \pm 1.08$ & $5.87\%$ & $\textbf{83.75} \pm 0.84$ & $8.92\%$ \\   
6 & $83.17 \pm 0.46$ & $1.69\%$ & $\underline{84.64} \pm 0.52$ & $2.77\%$ & $84.46 \pm 0.25$ & $4.11\%$ & $84.35 \pm 0.64$ & $5.71\%$ & $\textbf{84.71} \pm 0.58$ & $8.58\%$ \\   
7 & $84.53 \pm 0.66$ & $1.7\%$ & $85.52 \pm 0.46$ & $2.75\%$ & $85.22 \pm 0.34$ & $4.01\%$ & $\underline{85.54} \pm 0.78$ & $5.6\%$ & $\textbf{85.72} \pm 0.57$ & $8.79\%$ \\   
8 & $85.37 \pm 0.78$ & $1.66\%$ & $86.12 \pm 0.54$ & $2.75\%$ & $85.92 \pm 0.39$ & $4.08\%$ & $\underline{86.27} \pm 0.71$ & $5.6\%$ & $\textbf{86.31} \pm 0.4$ & $8.75\%$ \\   
9 & $85.83 \pm 0.77$ & $1.72\%$ & $86.67 \pm 0.38$ & $2.78\%$ & $86.49 \pm 0.18$ & $4.12\%$ & $\textbf{86.89} \pm 0.71$ & $5.83\%$ & $\underline{86.87} \pm 0.29$ & $8.84\%$ \\   
10 & $86.64 \pm 0.65$ & $1.66\%$ & $87.27 \pm 0.37$ & $2.83\%$ & $86.88 \pm 0.15$ & $4.12\%$ & $\textbf{87.38} \pm 0.92$ & $5.78\%$ & $\underline{87.27} \pm 0.45$ & $8.96\%$ \\   
11 & $87.13 \pm 0.86$ & $1.64\%$ & $87.64 \pm 0.3$ & $2.77\%$ & $87.56 \pm 0.46$ & $4.12\%$ & $\textbf{87.77} \pm 0.73$ & $5.91\%$ & $\underline{87.72} \pm 0.55$ & $8.8\%$ \\   
12 & $87.4 \pm 0.75$ & $1.68\%$ & $88.02 \pm 0.49$ & $2.79\%$ & $88.05 \pm 0.43$ & $4.1\%$ & $\textbf{88.17} \pm 0.56$ & $5.86\%$ & $\underline{88.07} \pm 0.63$ & $8.77\%$ \\   
13 & $87.74 \pm 0.84$ & $1.7\%$ & $88.45 \pm 0.5$ & $2.8\%$ & $88.15 \pm 0.51$ & $4.07\%$ & $\underline{88.48} \pm 0.64$ & $5.93\%$ & $\textbf{88.51} \pm 0.59$ & $8.9\%$ \\   
14 & $88.03 \pm 0.74$ & $1.71\%$ & $\underline{88.89} \pm 0.3$ & $2.77\%$ & $88.5 \pm 0.34$ & $4.12\%$ & $88.82 \pm 0.49$ & $5.93\%$ & $\textbf{88.92} \pm 0.67$ & $8.94\%$ \\   
15 & $88.29 \pm 0.83$ & $1.73\%$ & $88.99 \pm 0.22$ & $2.8\%$ & $88.73 \pm 0.36$ & $4.17\%$ & $\textbf{89.09} \pm 0.56$ & $6.01\%$ & $\underline{89.01} \pm 0.54$ & $8.99\%$ \\   
16 & $88.82 \pm 0.6$ & $1.7\%$ & $89.22 \pm 0.13$ & $2.79\%$ & $88.9 \pm 0.37$ & $4.17\%$ & $\underline{89.25} \pm 0.53$ & $6.11\%$ & $\textbf{89.36} \pm 0.5$ & $9.03\%$ \\   

\midrule
\multicolumn{11}{c}{ImageNet-100} \\
 \midrule
1 & $\textbf{79.75} \pm 1.23$ & $0.0\%$ & $79.46 \pm 1.33$ & $0.0\%$ & $\underline{79.54} \pm 1.2$ & $0.0\%$ & $79.4 \pm 1.51$ & $0.0\%$ & $79.12 \pm 1.07$ & $0.0\%$ \\   
2 & $88.0 \pm 0.79$ & $3.92\%$ & $88.65 \pm 1.08$ & $6.19\%$ & $88.69 \pm 0.87$ & $8.61\%$ & $\underline{88.88} \pm 1.03$ & $11.1\%$ & $\textbf{89.4} \pm 0.78$ & $14.59\%$ \\   
3 & $89.94 \pm 0.77$ & $1.04\%$ & $90.09 \pm 0.72$ & $1.7\%$ & $90.24 \pm 0.88$ & $2.52\%$ & $\textbf{90.4} \pm 0.75$ & $3.63\%$ & $\underline{90.29} \pm 1.22$ & $5.49\%$ \\   
4 & $90.84 \pm 0.71$ & $0.78\%$ & $90.92 \pm 0.63$ & $1.29\%$ & $\textbf{91.52} \pm 0.58$ & $1.89\%$ & $\underline{91.46} \pm 0.48$ & $2.74\%$ & $91.42 \pm 0.53$ & $4.19\%$ \\   
5 & $91.66 \pm 0.53$ & $0.7\%$ & $91.75 \pm 0.41$ & $1.12\%$ & $\textbf{92.04} \pm 0.6$ & $1.62\%$ & $91.92 \pm 0.38$ & $2.38\%$ & $\underline{91.93} \pm 0.46$ & $3.64\%$ \\   
6 & $92.3 \pm 0.46$ & $0.65\%$ & $92.2 \pm 0.34$ & $1.0\%$ & $\textbf{92.37} \pm 0.48$ & $1.45\%$ & $\underline{92.31} \pm 0.29$ & $2.14\%$ & $92.18 \pm 0.45$ & $3.26\%$ \\   
7 & $92.53 \pm 0.34$ & $0.59\%$ & $92.55 \pm 0.36$ & $0.91\%$ & $\textbf{92.71} \pm 0.38$ & $1.36\%$ & $92.55 \pm 0.43$ & $2.06\%$ & $\underline{92.62} \pm 0.4$ & $3.16\%$ \\   
8 & $92.64 \pm 0.43$ & $0.56\%$ & $92.72 \pm 0.37$ & $0.86\%$ & $92.84 \pm 0.29$ & $1.29\%$ & $\underline{92.87} \pm 0.3$ & $1.93\%$ & $\textbf{92.94} \pm 0.2$ & $2.99\%$ \\   
9 & $92.82 \pm 0.41$ & $0.54\%$ & $93.02 \pm 0.18$ & $0.83\%$ & $\underline{93.12} \pm 0.39$ & $1.23\%$ & $93.06 \pm 0.23$ & $1.91\%$ & $\textbf{93.14} \pm 0.24$ & $2.92\%$ \\   
10 & $93.04 \pm 0.52$ & $0.51\%$ & $93.19 \pm 0.31$ & $0.83\%$ & $93.15 \pm 0.35$ & $1.21\%$ & $\textbf{93.32} \pm 0.27$ & $1.86\%$ & $\underline{93.24} \pm 0.35$ & $2.9\%$ \\   
11 & $93.07 \pm 0.35$ & $0.52\%$ & $\textbf{93.48} \pm 0.35$ & $0.81\%$ & $93.43 \pm 0.22$ & $1.23\%$ & $\underline{93.46} \pm 0.26$ & $1.83\%$ & $93.35 \pm 0.43$ & $2.88\%$ \\   
12 & $93.26 \pm 0.37$ & $0.49\%$ & $93.54 \pm 0.44$ & $0.83\%$ & $\underline{93.55} \pm 0.22$ & $1.18\%$ & $\textbf{93.57} \pm 0.17$ & $1.74\%$ & $93.54 \pm 0.47$ & $2.85\%$ \\   
13 & $93.4 \pm 0.41$ & $0.49\%$ & $93.66 \pm 0.41$ & $0.79\%$ & $\underline{93.7} \pm 0.2$ & $1.13\%$ & $93.62 \pm 0.17$ & $1.77\%$ & $\textbf{93.72} \pm 0.45$ & $2.75\%$ \\   
14 & $93.38 \pm 0.45$ & $0.5\%$ & $\underline{93.8} \pm 0.34$ & $0.8\%$ & $93.76 \pm 0.33$ & $1.12\%$ & $\textbf{94.04} \pm 0.4$ & $1.76\%$ & $93.71 \pm 0.32$ & $2.74\%$ \\   
15 & $93.57 \pm 0.53$ & $0.5\%$ & $93.83 \pm 0.53$ & $0.78\%$ & $93.9 \pm 0.13$ & $1.16\%$ & $\textbf{94.04} \pm 0.37$ & $1.77\%$ & $\underline{93.93} \pm 0.44$ & $2.68\%$ \\   
16 & $93.78 \pm 0.59$ & $0.49\%$ & $94.01 \pm 0.6$ & $0.76\%$ & $93.9 \pm 0.13$ & $1.15\%$ & $\textbf{94.22} \pm 0.42$ & $1.7\%$ & $\underline{94.06} \pm 0.32$ & $2.62\%$ \\   

\midrule
\multicolumn{11}{c}{DomainNet-Real} \\
 \midrule
1 & $\underline{68.31} \pm 0.39$ & $0.0\%$ & $\textbf{68.34} \pm 0.5$ & $0.0\%$ & $68.19 \pm 0.47$ & $0.0\%$ & $68.13 \pm 0.52$ & $0.0\%$ & $67.96 \pm 0.47$ & $0.0\%$ \\   
2 & $74.15 \pm 0.19$ & $3.16\%$ & $\underline{74.56} \pm 0.47$ & $5.09\%$ & $74.15 \pm 0.52$ & $7.63\%$ & $74.37 \pm 0.29$ & $10.57\%$ & $\textbf{74.97} \pm 0.09$ & $15.42\%$ \\   
3 & $76.4 \pm 0.33$ & $1.66\%$ & $76.9 \pm 0.44$ & $2.73\%$ & $76.96 \pm 0.31$ & $4.11\%$ & $\textbf{77.19} \pm 0.22$ & $5.95\%$ & $\underline{77.16} \pm 0.22$ & $9.27\%$ \\   
4 & $77.96 \pm 0.51$ & $1.57\%$ & $78.31 \pm 0.48$ & $2.52\%$ & $78.67 \pm 0.25$ & $3.69\%$ & $\textbf{78.88} \pm 0.26$ & $5.24\%$ & $\underline{78.68} \pm 0.39$ & $8.29\%$ \\   
5 & $79.24 \pm 0.47$ & $1.56\%$ & $79.55 \pm 0.38$ & $2.51\%$ & $79.85 \pm 0.3$ & $3.57\%$ & $\textbf{80.19} \pm 0.28$ & $5.15\%$ & $\underline{79.88} \pm 0.21$ & $8.08\%$ \\   
6 & $80.13 \pm 0.37$ & $1.56\%$ & $80.6 \pm 0.24$ & $2.52\%$ & $80.7 \pm 0.26$ & $3.55\%$ & $\textbf{81.11} \pm 0.27$ & $5.02\%$ & $\underline{80.8} \pm 0.31$ & $8.04\%$ \\   
7 & $80.94 \pm 0.4$ & $1.58\%$ & $81.35 \pm 0.34$ & $2.52\%$ & $81.51 \pm 0.11$ & $3.52\%$ & $\textbf{81.85} \pm 0.19$ & $5.07\%$ & $\underline{81.58} \pm 0.21$ & $7.96\%$ \\   
8 & $81.7 \pm 0.33$ & $1.59\%$ & $82.07 \pm 0.2$ & $2.52\%$ & $\underline{82.27} \pm 0.1$ & $3.57\%$ & $\textbf{82.28} \pm 0.12$ & $5.05\%$ & $82.21 \pm 0.13$ & $8.05\%$ \\   
9 & $82.25 \pm 0.3$ & $1.57\%$ & $82.65 \pm 0.13$ & $2.53\%$ & $\textbf{82.78} \pm 0.13$ & $3.53\%$ & $82.73 \pm 0.1$ & $5.12\%$ & $\underline{82.77} \pm 0.14$ & $8.1\%$ \\   
10 & $82.59 \pm 0.28$ & $1.56\%$ & $83.1 \pm 0.17$ & $2.55\%$ & $83.16 \pm 0.14$ & $3.49\%$ & $\underline{83.19} \pm 0.15$ & $5.14\%$ & $\textbf{83.27} \pm 0.16$ & $8.15\%$ \\   
11 & $83.05 \pm 0.28$ & $1.58\%$ & $83.43 \pm 0.18$ & $2.54\%$ & $\underline{83.6} \pm 0.19$ & $3.52\%$ & $83.53 \pm 0.07$ & $5.19\%$ & $\textbf{83.65} \pm 0.19$ & $8.38\%$ \\   
12 & $81.0 \pm 0.32$ & $1.53\%$ & $81.33 \pm 0.47$ & $2.55\%$ & $81.88 \pm 0.32$ & $3.48\%$ & $\underline{82.43} \pm 0.2$ & $5.19\%$ & $\textbf{82.79} \pm 0.27$ & $8.14\%$ \\   
13 & $83.27 \pm 0.26$ & $2.89\%$ & $83.27 \pm 0.17$ & $5.16\%$ & $\underline{83.5} \pm 0.22$ & $6.59\%$ & $83.48 \pm 0.17$ & $8.36\%$ & $\textbf{83.75} \pm 0.11$ & $13.12\%$ \\   
14 & $83.8 \pm 0.18$ & $1.88\%$ & $84.03 \pm 0.04$ & $3.28\%$ & $84.03 \pm 0.16$ & $4.89\%$ & $\underline{84.03} \pm 0.15$ & $7.55\%$ & $\textbf{84.12} \pm 0.2$ & $10.59\%$ \\   
15 & $84.15 \pm 0.1$ & $1.71\%$ & $\underline{84.35} \pm 0.13$ & $2.86\%$ & $84.3 \pm 0.12$ & $4.42\%$ & $84.32 \pm 0.13$ & $6.68\%$ & $\textbf{84.39} \pm 0.27$ & $9.86\%$ \\   
16 & $84.36 \pm 0.12$ & $1.64\%$ & $84.56 \pm 0.17$ & $2.73\%$ & $84.58 \pm 0.15$ & $4.32\%$ & $\underline{84.59} \pm 0.24$ & $6.33\%$ & $\textbf{84.72} \pm 0.17$ & $9.33\%$ \\   

\bottomrule
\end{tabular}
\end{adjustbox}
\label{tab:dropout_ratio}
\end{table}

\begin{table}[h]
\centering
\tiny
\caption{Impact of the number of dropout iterations. We evaluate 4 different settings of $M$ on 4 natural image datasets - CIFAR100, Food101, ImageNet-100, and DomainNet-Real. We report the mean accuracy averaged over 5 runs along with the standard deviation at each AL iteration. \textbf{Bold} values represent the \textbf{first place} mean accuracy at iteration $t$ with the \underline{second place} value \underline{underlined}. We also report the corresponding fraction of the unlabeled examples added to the candidate set.}
\vspace{0.5em}
\begin{adjustbox}{width=0.95\textwidth}
\begin{tabular}{c|cc|cc|cc|cc}
\toprule
 $t$ & \multicolumn{2}{c}{3} & \multicolumn{2}{c}{5} & \multicolumn{2}{c}{7} & \multicolumn{2}{c}{9} \\ 
  & Mean ± Std & Fraction & Mean ± Std & Fraction & Mean ± Std & Fraction & Mean ± Std & Fraction \\ 
\midrule
\multicolumn{9}{c}{CIFAR100} \\
\midrule
1 & $\textbf{72.58} \pm 2.18$ & $0.0\%$ & $72.58 \pm 2.18$ & $0.0\%$ & $\underline{72.58} \pm 2.18$ & $0.0\%$ & $72.58 \pm 2.18$ & $0.0\%$ \\    
2 & $83.52 \pm 0.76$ & $21.53\%$ & $\textbf{84.16} \pm 1.31$ & $21.64\%$ & $83.46 \pm 1.04$ & $21.7\%$ & $\underline{83.55} \pm 0.88$ & $21.72\%$ \\    
3 & $\textbf{86.6} \pm 0.78$ & $11.7\%$ & $85.9 \pm 0.94$ & $11.37\%$ & $85.82 \pm 0.69$ & $11.54\%$ & $\underline{86.2} \pm 0.55$ & $11.35\%$ \\    
4 & $\underline{87.67} \pm 0.57$ & $9.22\%$ & $\textbf{87.76} \pm 0.46$ & $9.12\%$ & $87.39 \pm 1.02$ & $8.81\%$ & $87.6 \pm 0.37$ & $8.78\%$ \\    
5 & $\textbf{88.94} \pm 0.31$ & $8.33\%$ & $\underline{88.72} \pm 0.38$ & $7.89\%$ & $88.21 \pm 0.73$ & $7.62\%$ & $88.43 \pm 0.48$ & $7.65\%$ \\    
6 & $\textbf{89.35} \pm 0.16$ & $7.72\%$ & $\underline{89.17} \pm 0.24$ & $7.26\%$ & $89.06 \pm 0.42$ & $7.17\%$ & $88.98 \pm 0.33$ & $7.01\%$ \\    
7 & $\textbf{89.65} \pm 0.31$ & $7.37\%$ & $89.53 \pm 0.2$ & $6.97\%$ & $89.53 \pm 0.33$ & $6.7\%$ & $\underline{89.6} \pm 0.18$ & $6.69\%$ \\    
8 & $\textbf{90.12} \pm 0.14$ & $7.07\%$ & $89.71 \pm 0.21$ & $6.62\%$ & $\underline{90.07} \pm 0.26$ & $6.42\%$ & $89.86 \pm 0.18$ & $6.3\%$ \\    
9 & $\textbf{90.23} \pm 0.21$ & $6.88\%$ & $\underline{90.11} \pm 0.38$ & $6.55\%$ & $89.99 \pm 0.25$ & $6.23\%$ & $90.02 \pm 0.21$ & $6.15\%$ \\    
10 & $\textbf{90.48} \pm 0.22$ & $6.75\%$ & $90.39 \pm 0.27$ & $6.36\%$ & $\underline{90.41} \pm 0.37$ & $6.09\%$ & $90.26 \pm 0.11$ & $6.02\%$ \\    
11 & $\textbf{90.6} \pm 0.28$ & $6.68\%$ & $90.5 \pm 0.13$ & $6.27\%$ & $\underline{90.59} \pm 0.36$ & $5.98\%$ & $90.52 \pm 0.19$ & $5.84\%$ \\    
12 & $\textbf{90.8} \pm 0.27$ & $6.54\%$ & $90.59 \pm 0.24$ & $6.24\%$ & $90.63 \pm 0.34$ & $5.86\%$ & $\underline{90.65} \pm 0.17$ & $5.63\%$ \\    
13 & $\textbf{90.99} \pm 0.2$ & $6.37\%$ & $90.81 \pm 0.14$ & $6.02\%$ & $\underline{90.96} \pm 0.27$ & $5.82\%$ & $90.84 \pm 0.21$ & $5.64\%$ \\    
14 & $\textbf{91.26} \pm 0.25$ & $6.35\%$ & $90.92 \pm 0.17$ & $6.02\%$ & $\underline{91.05} \pm 0.22$ & $5.7\%$ & $91.01 \pm 0.18$ & $5.59\%$ \\    
15 & $\textbf{91.32} \pm 0.26$ & $6.27\%$ & $91.05 \pm 0.06$ & $5.85\%$ & $\underline{91.13} \pm 0.33$ & $5.59\%$ & $91.01 \pm 0.12$ & $5.31\%$ \\    
16 & $\textbf{91.47} \pm 0.16$ & $6.0\%$ & $91.21 \pm 0.15$ & $5.73\%$ & $\underline{91.37} \pm 0.19$ & $5.42\%$ & $91.19 \pm 0.21$ & $5.21\%$ \\    

\midrule
\multicolumn{9}{c}{Food101} \\
 \midrule
1 & $\textbf{71.33} \pm 1.38$ & $0.0\%$ & $71.33 \pm 1.38$ & $0.0\%$ & $\underline{71.33} \pm 1.38$ & $0.0\%$ & $71.33 \pm 1.38$ & $0.0\%$ \\    
2 & $\textbf{76.09} \pm 1.34$ & $8.49\%$ & $75.65 \pm 1.37$ & $7.81\%$ & $\underline{75.74} \pm 1.74$ & $7.43\%$ & $75.62 \pm 2.14$ & $7.22\%$ \\    
3 & $\textbf{79.76} \pm 0.88$ & $8.6\%$ & $79.15 \pm 1.48$ & $8.16\%$ & $\underline{79.73} \pm 1.55$ & $7.72\%$ & $78.55 \pm 2.43$ & $7.36\%$ \\    
4 & $\textbf{82.24} \pm 0.52$ & $8.75\%$ & $81.6 \pm 0.86$ & $8.05\%$ & $\underline{81.88} \pm 1.06$ & $7.44\%$ & $81.27 \pm 1.39$ & $7.41\%$ \\    
5 & $\textbf{83.75} \pm 0.84$ & $8.92\%$ & $83.01 \pm 0.8$ & $8.23\%$ & $\underline{83.32} \pm 1.18$ & $7.64\%$ & $82.87 \pm 1.08$ & $7.44\%$ \\    
6 & $\underline{84.71} \pm 0.58$ & $8.58\%$ & $\textbf{84.73} \pm 0.78$ & $8.22\%$ & $84.33 \pm 1.07$ & $7.59\%$ & $84.2 \pm 1.06$ & $7.51\%$ \\    
7 & $\textbf{85.72} \pm 0.57$ & $8.79\%$ & $85.3 \pm 0.68$ & $8.15\%$ & $\underline{85.56} \pm 0.88$ & $7.69\%$ & $85.05 \pm 0.88$ & $7.47\%$ \\    
8 & $\textbf{86.31} \pm 0.4$ & $8.75\%$ & $86.08 \pm 0.49$ & $8.09\%$ & $\underline{86.3} \pm 0.59$ & $7.95\%$ & $85.72 \pm 0.67$ & $7.44\%$ \\    
9 & $\underline{86.87} \pm 0.29$ & $8.84\%$ & $\textbf{87.04} \pm 0.44$ & $8.49\%$ & $86.77 \pm 0.69$ & $7.97\%$ & $86.3 \pm 0.7$ & $7.73\%$ \\    
10 & $87.27 \pm 0.45$ & $8.96\%$ & $\textbf{87.41} \pm 0.75$ & $8.28\%$ & $\underline{87.31} \pm 0.64$ & $8.02\%$ & $87.03 \pm 0.76$ & $7.72\%$ \\    
11 & $\underline{87.72} \pm 0.55$ & $8.8\%$ & $\textbf{87.8} \pm 0.88$ & $8.47\%$ & $87.63 \pm 0.66$ & $7.88\%$ & $87.55 \pm 0.68$ & $7.8\%$ \\    
12 & $\underline{88.07} \pm 0.63$ & $8.77\%$ & $88.06 \pm 0.72$ & $8.85\%$ & $\textbf{88.1} \pm 0.64$ & $7.92\%$ & $87.84 \pm 0.77$ & $7.86\%$ \\    
13 & $\textbf{88.51} \pm 0.59$ & $8.9\%$ & $88.44 \pm 0.51$ & $8.58\%$ & $\underline{88.51} \pm 0.8$ & $8.03\%$ & $88.11 \pm 0.61$ & $7.75\%$ \\    
14 & $\textbf{88.92} \pm 0.67$ & $8.94\%$ & $88.59 \pm 0.73$ & $8.48\%$ & $\underline{88.86} \pm 0.74$ & $7.95\%$ & $88.61 \pm 0.64$ & $7.97\%$ \\    
15 & $\underline{89.01} \pm 0.54$ & $8.99\%$ & $88.9 \pm 0.55$ & $8.55\%$ & $\textbf{89.04} \pm 0.85$ & $7.99\%$ & $88.89 \pm 0.77$ & $7.86\%$ \\    
16 & $\textbf{89.36} \pm 0.5$ & $9.03\%$ & $89.28 \pm 0.7$ & $8.39\%$ & $\underline{89.29} \pm 0.73$ & $7.92\%$ & $89.29 \pm 0.73$ & $7.82\%$ \\    

\midrule
\multicolumn{9}{c}{ImageNet-100} \\
 \midrule
1 & $\textbf{79.12} \pm 1.07$ & $0.0\%$ & $79.12 \pm 1.07$ & $0.0\%$ & $\underline{79.12} \pm 1.07$ & $0.0\%$ & $79.12 \pm 1.07$ & $0.0\%$ \\    
2 & $\textbf{89.4} \pm 0.78$ & $14.59\%$ & $\underline{89.38} \pm 1.01$ & $14.78\%$ & $89.06 \pm 1.07$ & $14.87\%$ & $88.79 \pm 1.1$ & $14.92\%$ \\    
3 & $90.29 \pm 1.22$ & $5.49\%$ & $\underline{90.48} \pm 0.79$ & $5.37\%$ & $\textbf{90.49} \pm 1.18$ & $5.26\%$ & $90.23 \pm 0.82$ & $5.24\%$ \\    
4 & $\underline{91.42} \pm 0.53$ & $4.19\%$ & $\textbf{91.47} \pm 0.42$ & $4.02\%$ & $91.24 \pm 0.63$ & $3.93\%$ & $91.14 \pm 0.54$ & $3.86\%$ \\    
5 & $\textbf{91.93} \pm 0.46$ & $3.64\%$ & $91.89 \pm 0.43$ & $3.44\%$ & $\underline{91.9} \pm 0.36$ & $3.36\%$ & $91.84 \pm 0.56$ & $3.31\%$ \\    
6 & $92.18 \pm 0.45$ & $3.26\%$ & $\underline{92.43} \pm 0.56$ & $3.08\%$ & $92.17 \pm 0.48$ & $2.96\%$ & $\textbf{92.47} \pm 0.61$ & $2.96\%$ \\    
7 & $92.62 \pm 0.4$ & $3.16\%$ & $\underline{92.68} \pm 0.58$ & $2.85\%$ & $92.67 \pm 0.4$ & $2.76\%$ & $\textbf{92.72} \pm 0.57$ & $2.75\%$ \\    
8 & $\underline{92.94} \pm 0.2$ & $2.99\%$ & $92.79 \pm 0.61$ & $2.74\%$ & $92.72 \pm 0.38$ & $2.63\%$ & $\textbf{92.95} \pm 0.26$ & $2.6\%$ \\    
9 & $\underline{93.14} \pm 0.24$ & $2.92\%$ & $\textbf{93.2} \pm 0.33$ & $2.68\%$ & $92.91 \pm 0.26$ & $2.58\%$ & $92.97 \pm 0.31$ & $2.51\%$ \\    
10 & $\underline{93.24} \pm 0.35$ & $2.9\%$ & $\textbf{93.37} \pm 0.21$ & $2.6\%$ & $93.09 \pm 0.18$ & $2.53\%$ & $93.14 \pm 0.31$ & $2.45\%$ \\    
11 & $93.35 \pm 0.43$ & $2.88\%$ & $\underline{93.38} \pm 0.22$ & $2.54\%$ & $\textbf{93.39} \pm 0.17$ & $2.42\%$ & $93.34 \pm 0.18$ & $2.37\%$ \\    
12 & $\textbf{93.54} \pm 0.47$ & $2.85\%$ & $\underline{93.5} \pm 0.22$ & $2.53\%$ & $93.42 \pm 0.18$ & $2.35\%$ & $93.41 \pm 0.21$ & $2.29\%$ \\    
13 & $\textbf{93.72} \pm 0.45$ & $2.75\%$ & $\underline{93.51} \pm 0.22$ & $2.48\%$ & $93.47 \pm 0.2$ & $2.31\%$ & $93.38 \pm 0.14$ & $2.26\%$ \\    
14 & $\textbf{93.71} \pm 0.32$ & $2.74\%$ & $\underline{93.65} \pm 0.17$ & $2.43\%$ & $93.56 \pm 0.14$ & $2.23\%$ & $93.54 \pm 0.2$ & $2.23\%$ \\    
15 & $\textbf{93.93} \pm 0.44$ & $2.68\%$ & $93.65 \pm 0.14$ & $2.39\%$ & $93.62 \pm 0.12$ & $2.22\%$ & $\underline{93.79} \pm 0.27$ & $2.18\%$ \\    
16 & $\textbf{94.06} \pm 0.32$ & $2.62\%$ & $93.69 \pm 0.2$ & $2.34\%$ & $93.63 \pm 0.24$ & $2.23\%$ & $\underline{93.83} \pm 0.18$ & $2.16\%$ \\    

\midrule
\multicolumn{9}{c}{DomainNet-Real} \\
 \midrule
1 & $\textbf{67.96} \pm 0.47$ & $0.0\%$ & $67.96 \pm 0.47$ & $0.0\%$ & $\underline{67.96} \pm 0.47$ & $0.0\%$ & $67.96 \pm 0.47$ & $0.0\%$ \\    
2 & $\underline{74.97} \pm 0.09$ & $15.42\%$ & $\textbf{74.97} \pm 0.24$ & $15.36\%$ & $74.85 \pm 0.31$ & $15.29\%$ & $74.94 \pm 0.38$ & $15.26\%$ \\    
3 & $\textbf{77.16} \pm 0.22$ & $9.27\%$ & $\underline{77.16} \pm 0.26$ & $8.94\%$ & $77.06 \pm 0.16$ & $8.84\%$ & $77.09 \pm 0.21$ & $8.63\%$ \\    
4 & $78.68 \pm 0.39$ & $8.29\%$ & $78.67 \pm 0.15$ & $7.89\%$ & $\textbf{78.74} \pm 0.35$ & $7.63\%$ & $\underline{78.71} \pm 0.26$ & $7.48\%$ \\    
5 & $\underline{79.88} \pm 0.21$ & $8.08\%$ & $79.85 \pm 0.2$ & $7.56\%$ & $79.88 \pm 0.19$ & $7.36\%$ & $\textbf{79.95} \pm 0.21$ & $7.17\%$ \\    
6 & $80.8 \pm 0.31$ & $8.04\%$ & $80.81 \pm 0.17$ & $7.55\%$ & $\textbf{80.86} \pm 0.22$ & $7.23\%$ & $\underline{80.82} \pm 0.16$ & $7.04\%$ \\    
7 & $\textbf{81.58} \pm 0.21$ & $7.96\%$ & $81.47 \pm 0.21$ & $7.52\%$ & $\underline{81.58} \pm 0.34$ & $7.25\%$ & $81.5 \pm 0.19$ & $7.0\%$ \\    
8 & $\underline{82.21} \pm 0.13$ & $8.05\%$ & $82.18 \pm 0.3$ & $7.68\%$ & $\textbf{82.3} \pm 0.24$ & $7.22\%$ & $82.07 \pm 0.24$ & $7.1\%$ \\    
9 & $82.77 \pm 0.14$ & $8.1\%$ & $\underline{82.79} \pm 0.28$ & $7.65\%$ & $\textbf{82.8} \pm 0.27$ & $7.22\%$ & $82.59 \pm 0.22$ & $7.19\%$ \\    
10 & $\textbf{83.27} \pm 0.16$ & $8.15\%$ & $83.19 \pm 0.25$ & $7.71\%$ & $\underline{83.2} \pm 0.3$ & $7.39\%$ & $83.07 \pm 0.33$ & $7.27\%$ \\    
11 & $\textbf{83.65} \pm 0.19$ & $8.38\%$ & $\underline{83.55} \pm 0.15$ & $7.83\%$ & $83.53 \pm 0.17$ & $7.44\%$ & $83.53 \pm 0.28$ & $7.11\%$ \\    
12 & $\underline{82.79} \pm 0.27$ & $8.14\%$ & $\textbf{82.79} \pm 0.24$ & $7.56\%$ & $82.49 \pm 0.21$ & $7.39\%$ & $82.6 \pm 0.2$ & $7.22\%$ \\    
13 & $\textbf{83.75} \pm 0.11$ & $13.12\%$ & $83.65 \pm 0.23$ & $12.41\%$ & $\underline{83.73} \pm 0.16$ & $12.19\%$ & $83.65 \pm 0.15$ & $11.76\%$ \\    
14 & $84.12 \pm 0.2$ & $10.59\%$ & $84.12 \pm 0.24$ & $9.88\%$ & $\textbf{84.31} \pm 0.11$ & $9.85\%$ & $\underline{84.22} \pm 0.07$ & $9.75\%$ \\    
15 & $84.39 \pm 0.27$ & $9.86\%$ & $84.44 \pm 0.19$ & $9.1\%$ & $\textbf{84.6} \pm 0.1$ & $8.78\%$ & $\underline{84.47} \pm 0.1$ & $8.64\%$ \\    
16 & $84.72 \pm 0.17$ & $9.33\%$ & $84.67 \pm 0.17$ & $8.65\%$ & $\textbf{84.81} \pm 0.11$ & $8.45\%$ & $\underline{84.76} \pm 0.04$ & $8.28\%$ \\    

\bottomrule
\end{tabular}
\end{adjustbox}
\label{tab:iterations}
\end{table}

\begin{table}[h]
\centering
\caption{Descriptions of the biomedical datasets utilized in our study, which span a wide range of imaging modalities, biomedical disciplines, and task difficulties. Abbreviations: binary classification (BC), hematoxylin and eosin (H\&E), immunofluorescence (IF), multiclass classification (MC). All train/ val/ test splits are 0.7/ 0.1/ 0.2 of the total unless noted otherwise in the text.}
\vspace{0.5em}
\begin{adjustbox}{width=1\textwidth}
\begin{tabular}{ccccccccccccc}
\toprule
\rowcolor{Gray!25} Name & Modality & Domain & Stain & Task (Classes) & Image size & Total \\ 
\midrule
Blood Smear \citep{blood_smear} & Brightfield & Cytology & Giemsa & MC (8) & (360, 363)  & 17092 \\ 
Colorectal histology \citep{colorectal_histology} &  Brightfield & Pathology & H\&E & MC (8) & (150, 150) & 5000 \\ 
Diabetic Retinopathy \citep{diabetic_retinopathy} & Fundoscopy & Ophthalmology & - & MC (5) & (256, 256)  & 2750 \\ 
IICBU 2008 HeLa\citep{iicbu_hela} & Fluorescence & Cell Biology & IF & MC (10) & (382, 382) & 862 \\ 
Skin cancer \citep{ham10000} &  Dermoscopy & Dermatology & - & MC (7) & (450, 600) &  10015 \\ 
Patch Camelyon \citep{patch_camelyon_1}\citep{patch_camelyon_2} &  Brightfield & Pathology & H\&E & BC (2) & (96, 96)  & 327,680 \\
\bottomrule
\end{tabular}
\end{adjustbox}
\end{table}

\begin{table}[h]
    \centering
    \caption{Test set accuracy with standard deviations (average of 5 random seeds) at certain iterations $t$ during AL on CIFAR100 \citep{cifar100}, Food101 \citep{food101}, ImageNet-100 \citep{imagenet100}, and DomainNet-Real \citep{domainnetreal} (from top to bottom) with DINOv2 ViT-g14 as the feature extractor $f$. Typiclust, ProbCover, and our proposed \dropout use their own respective initialization strategies for the cold-start. Cells are color coded according to the magnitude of improvement in mean accuracy over the Random baseline. Green cells are positive with better performance than random whereas red cells are negative with worse performance than random. \textbf{Bold} values represent the \textbf{first place} mean accuracy at iteration $t$ with the \underline{second place} value \underline{underlined}.}
    \vspace{0.5em}
    \begin{adjustbox}{width=1\textwidth}
\begin{tabular}{c | c | c c c | c c | c c c | c c | c}
    \toprule 
        $t$ & Random & Uncertainty & Entropy & Margins & BALD & pBALD & Coreset & Typiclust & ProbCover & BADGE & Alfamix & DropQuery (Ours)\\
    \midrule
    \multicolumn{12}{c}{CIFAR100} \\
    \midrule
1 & 
    $ 48.1 \pm2.1 $ & 
    \cellcolor{LimeGreen!0!white} $48.1 \pm{2.1}$ & 
    \cellcolor{LimeGreen!0!white} $48.1 \pm{2.1}$ & 
    \cellcolor{LimeGreen!0!white} $48.1 \pm{2.1}$ & 
    \cellcolor{LimeGreen!0!white} $48.1 \pm{2.1}$ & 
    \cellcolor{LimeGreen!0!white} $48.1 \pm{2.1}$ & 
    \cellcolor{LimeGreen!0!white} $48.1 \pm{2.1}$ & 
    \cellcolor{LimeGreen!66!white} $\underline{64.6 \pm{2.0}}$ & 
    \cellcolor{LimeGreen!57!white} $62.3 \pm{1.1}$ & 
    \cellcolor{LimeGreen!0!white} $48.1 \pm{2.1}$ & 
    \cellcolor{LimeGreen!0!white} $48.1 \pm{2.1}$ & 
    \cellcolor{LimeGreen!98!white} $\bm{72.6 \pm{2.2}}$ \\

2 & 
    $ 64.5 \pm2.7 $ & 
    \cellcolor{red!8!white} $62.6 \pm{1.8}$ & 
    \cellcolor{red!26!white} $58.0 \pm{2.3}$ & 
    \cellcolor{LimeGreen!21!white} $69.9 \pm{2.6}$ & 
    \cellcolor{LimeGreen!11!white} $67.4 \pm{1.7}$ & 
    \cellcolor{LimeGreen!27!white} $71.2 \pm{1.3}$ & 
    \cellcolor{red!0!white} $64.5 \pm{2.2}$ & 
    \cellcolor{LimeGreen!65!white} $\underline{80.8 \pm{0.2}}$ & 
    \cellcolor{LimeGreen!47!white} $76.2 \pm{1.9}$ & 
    \cellcolor{LimeGreen!32!white} $72.5 \pm{2.0}$ & 
    \cellcolor{LimeGreen!54!white} $78.0 \pm{1.1}$ & 
    \cellcolor{LimeGreen!76!white} $\bm{83.5 \pm{0.8}}$ \\
\bottomrule
    \end{tabular}
    \end{adjustbox}
    \label{tab:full_results_nat1}
\end{table}

\begin{table}[h]
    \centering
    \caption{Test set accuracy with standard deviations (average of 5 random seeds) at certain iterations $t$ during AL on other VTAB+ \citep{vtab} datasets including Stanford Cars \citep{stanfordcars}, FVGC Aircraft \citep{fvgcaircraft}, Oxford-IIIT Pets \citep{oxfordpets}, and the large Places365 \citep{places365} dataset (from top to bottom) with OpenCLIP ViT-G14 as the feature extractor $f$. Cells are color coded according to the magnitude of improvement in mean accuracy over the Random baseline. Probcover was intractable on Places365 with the resources available, and thus not performed (-). \textbf{Bold} values represent the \textbf{first place} mean accuracy at iteration $t$ with the \underline{second place} value \underline{underlined}.}
    \vspace{0.5em}
    \begin{adjustbox}{width=1\textwidth}
\begin{tabular}{c | c | c c c | c c | c c c | c c | c}
    \toprule 
        $t$ & Random & Uncertainty & Entropy & Margins & BALD & pBALD & Coreset & Typiclust & ProbCover & BADGE & Alfamix & DropQuery (Ours)\\
    \midrule
    \multicolumn{12}{c}{Stanford Cars} \\
    \midrule
   1 & 
    $ 51.9 \pm1.1 $ & 
    \cellcolor{LimeGreen!0!white} $51.9 \pm{1.1}$ & 
    \cellcolor{LimeGreen!0!white} $51.9 \pm{1.1}$ & 
    \cellcolor{LimeGreen!0!white} $51.9 \pm{1.1}$ & 
    \cellcolor{LimeGreen!1!white} $52.1 \pm{1.2}$ & 
    \cellcolor{LimeGreen!0!white} $51.9 \pm{1.1}$ & 
    \cellcolor{LimeGreen!0!white} $51.9 \pm{1.1}$ & 
    \cellcolor{LimeGreen!89!white} $\bm{74.2 \pm{1.6}}$ & 
    \cellcolor{LimeGreen!88!white} $\underline{74.0 \pm{2.2}}$ & 
    \cellcolor{LimeGreen!0!white} $51.9 \pm{1.1}$ & 
    \cellcolor{LimeGreen!0!white} $51.9 \pm{1.1}$ & 
    \cellcolor{LimeGreen!86!white} $73.4 \pm{1.7}$ \\

2 & 
    $ 72.3 \pm1.7 $ & 
    \cellcolor{red!13!white} $69.0 \pm{1.1}$ & 
    \cellcolor{red!24!white} $66.3 \pm{1.5}$ & 
    \cellcolor{LimeGreen!14!white} $75.7 \pm{2.1}$ & 
    \cellcolor{red!12!white} $69.2 \pm{1.6}$ & 
    \cellcolor{LimeGreen!17!white} $76.6 \pm{1.5}$ & 
    \cellcolor{LimeGreen!15!white} $76.1 \pm{0.4}$ & 
    \cellcolor{LimeGreen!63!white} $\bm{87.9 \pm{0.6}}$ & 
    \cellcolor{LimeGreen!55!white} $86.1 \pm{1.9}$ & 
    \cellcolor{LimeGreen!20!white} $77.2 \pm{1.6}$ & 
    \cellcolor{LimeGreen!40!white} $82.4 \pm{0.6}$ & 
    \cellcolor{LimeGreen!62!white} $\underline{87.8 \pm{0.5}}$ \\
\bottomrule
    \end{tabular}
    \end{adjustbox}
    \label{tab:full_results_nat2}
\end{table}

\begin{table}[h]
    \centering
    \caption{Test set accuracy with standard deviations (average of 5 random seeds) at certain iterations $t$ during AL on out-of-domain biomedical datasets including Blood Smear \citep{blood_smear}, Diabetic Retinopathy \citep{diabetic_retinopathy}, IICBU HeLa \citep{iicbu_hela}, and Skin cancer \citep{ham10000} (from top to bottom) with DINOv2 ViT-g14 as the feature extractor $f$. Probcover was intractable on Patch Camelyon with the resources available, and thus not performed (-). Cells are color coded according to the magnitude of improvement in mean accuracy over the Random baseline.}
    \vspace{0.5em}
    \begin{adjustbox}{width=1\textwidth}
\begin{tabular}{c | c | c c c | c c | c c c | c c | c}
    \toprule 
        $t$ & Random & Uncertainty & Entropy & Margins & BALD & pBALD & Coreset & Typiclust & ProbCover & BADGE & Alfamix & DropQuery (Ours)\\
    \midrule
    \multicolumn{12}{c}{Blood Smear} \\
    \midrule
   1 & 
    $ 40.8 \pm3.0 $ & 
    \cellcolor{LimeGreen!0!white} $40.8 \pm{3.0}$ & 
    \cellcolor{LimeGreen!0!white} $40.8 \pm{3.0}$ & 
    \cellcolor{LimeGreen!0!white} $40.8 \pm{3.0}$ & 
    \cellcolor{LimeGreen!0!white} $40.8 \pm{3.0}$ & 
    \cellcolor{LimeGreen!0!white} $40.8 \pm{3.0}$ & 
    \cellcolor{LimeGreen!0!white} $40.8 \pm{3.0}$ & 
    \cellcolor{LimeGreen!32!white} $\bm{48.7 \pm{9.8}}$ & 
    \cellcolor{LimeGreen!14!white} $\underline{44.2 \pm{17.9}}$ & 
    \cellcolor{LimeGreen!0!white} $40.8 \pm{3.0}$ & 
    \cellcolor{LimeGreen!0!white} $40.8 \pm{3.0}$ & 
    \cellcolor{red!4!white} $39.9 \pm{13.3}$ \\

2 & 
    $ 56.6 \pm2.5 $ & 
    \cellcolor{red!25!white} $50.4 \pm{3.0}$ & 
    \cellcolor{red!16!white} $52.7 \pm{2.8}$ & 
    \cellcolor{red!8!white} $54.6 \pm{4.0}$ & 
    \cellcolor{red!8!white} $54.6 \pm{6.0}$ & 
    \cellcolor{red!6!white} $55.1 \pm{5.6}$ & 
    \cellcolor{red!94!white} $33.2 \pm{5.6}$ & 
    \cellcolor{LimeGreen!28!white} $\underline{63.5 \pm{4.0}}$ & 
    \cellcolor{LimeGreen!23!white} $62.4 \pm{1.1}$ & 
    \cellcolor{red!3!white} $55.7 \pm{4.1}$ & 
    \cellcolor{LimeGreen!30!white} $\bm{64.0 \pm{2.6}}$ & 
    \cellcolor{LimeGreen!27!white} $63.3 \pm{2.5}$ \\
\bottomrule
    \end{tabular}
    \end{adjustbox}
    \label{tab:full_results_bio}
\end{table}

\begin{table}[h]
    \centering
    \caption{Test set accuracy with standard deviations (average of 5 random seeds) at certain iterations $t$ during AL on CIFAR100 \citep{cifar100}, Food101 \citep{food101}, ImageNet-100 \citep{imagenet100}, and DomainNet-Real \citep{domainnetreal} (from top to bottom) with an ImageNet-1K pretrained Masked AutoEncoder ViT-H/14 \citep{he2022masked} as the feature extractor $f$. Typiclust, and our proposed \dropout use their own respective initialization strategies for the cold-start. Cells are color coded according to the magnitude of improvement in mean accuracy over the Random baseline. Green cells are positive with better performance than random whereas red cells are negative with worse performance than random. \textbf{Bold} values represent the \textbf{first place} mean accuracy at iteration $t$ with the \underline{second place} value \underline{underlined}.}
    \vspace{0.5em}
    \begin{adjustbox}{width=0.8\textwidth}
    \begin{tabular}{c | c | c c c | c c | c }
    \toprule 
    $t$ & Random & Margins & pBALD & BADGE & Typiclust & Alfamix &  DropQuery (Ours)\\
    \midrule
    \multicolumn{8}{c}{CIFAR100} \\
    \midrule

1 &
     $\underline{14.2 \pm{1.3}}$ & 
     $\underline{14.2 \pm{1.3}}$ & 
     $\underline{14.2 \pm{1.3}}$ & 
     $\underline{14.2 \pm{1.3}}$ & 
    \cellcolor{LimeGreen!8!white} $\bm{16.2 \pm{0.6}}$ & 
     $\underline{14.2 \pm{1.3}}$ & 
    \cellcolor{red!4!white} $13.0 \pm{0.6}$ \\ 

2 &
     $21.0 \pm{1.5}$ & 
    \cellcolor{red!3!white} $20.2 \pm{1.3}$ & 
    \cellcolor{LimeGreen!1!white} $21.4 \pm{1.1}$ & 
    \cellcolor{LimeGreen!2!white} $21.8 \pm{0.6}$ & 
    \cellcolor{LimeGreen!2!white} $21.7 \pm{0.6}$ & 
    \cellcolor{LimeGreen!10!white} $\bm{23.7 \pm{0.9}}$ & 
    \cellcolor{LimeGreen!7!white} $\underline{22.9 \pm{1.5}}$ \\ 

4 &
     $31.0 \pm{1.3}$ & 
    \cellcolor{red!2!white} $30.4 \pm{0.8}$ & 
    \cellcolor{red!1!white} $30.6 \pm{0.9}$ & 
    \cellcolor{LimeGreen!1!white} $31.3 \pm{1.5}$ & 
    \cellcolor{red!7!white} $29.2 \pm{0.8}$ & 
    \cellcolor{LimeGreen!3!white} $\underline{31.9 \pm{0.9}}$ & 
    \cellcolor{LimeGreen!8!white} $\bm{33.1 \pm{0.9}}$ \\ 

6 &
     $37.5 \pm{1.0}$ & 
     $37.6 \pm{0.9}$ & 
    \cellcolor{LimeGreen!4!white} $\underline{38.5 \pm{1.2}}$ & 
    \cellcolor{LimeGreen!4!white} $\bm{38.6 \pm{1.0}}$ & 
    \cellcolor{red!9!white} $35.0 \pm{0.8}$ & 
     $37.4 \pm{0.9}$ & 
    \cellcolor{LimeGreen!3!white} $38.4 \pm{0.8}$ \\ 

8 &
     $42.1 \pm{1.2}$ & 
    \cellcolor{LimeGreen!2!white} $42.6 \pm{0.9}$ & 
    \cellcolor{LimeGreen!2!white} $\underline{42.8 \pm{1.0}}$ & 
    \cellcolor{LimeGreen!5!white} $\bm{43.4 \pm{0.9}}$ & 
    \cellcolor{red!11!white} $39.1 \pm{0.7}$ & 
    \cellcolor{red!4!white} $40.8 \pm{1.4}$ & 
    \cellcolor{LimeGreen!1!white} $42.5 \pm{0.7}$ \\ 

10 &
     $45.8 \pm{0.8}$ & 
     $45.9 \pm{1.0}$ & 
    \cellcolor{LimeGreen!2!white} $\underline{46.3 \pm{0.5}}$ & 
    \cellcolor{LimeGreen!4!white} $\bm{46.9 \pm{0.9}}$ & 
    \cellcolor{red!18!white} $41.3 \pm{0.9}$ & 
    \cellcolor{red!5!white} $44.4 \pm{2.0}$ & 
    \cellcolor{red!4!white} $44.7 \pm{0.3}$ \\ 

12 &
     $48.7 \pm{0.8}$ & 
    \cellcolor{red!2!white} $48.1 \pm{0.7}$ & 
    \cellcolor{LimeGreen!1!white} $\underline{49.2 \pm{0.8}}$ & 
    \cellcolor{LimeGreen!4!white} $\bm{49.7 \pm{0.6}}$ & 
    \cellcolor{red!19!white} $43.8 \pm{0.8}$ & 
    \cellcolor{red!6!white} $46.9 \pm{1.0}$ & 
    \cellcolor{red!6!white} $47.0 \pm{0.3}$ \\ 

14 &
     $50.9 \pm{0.6}$ & 
    \cellcolor{red!2!white} $50.4 \pm{0.9}$ & 
    \cellcolor{LimeGreen!2!white} $\underline{51.6 \pm{0.6}}$ & 
    \cellcolor{LimeGreen!4!white} $\bm{51.9 \pm{0.3}}$ & 
    \cellcolor{red!24!white} $44.8 \pm{0.7}$ & 
    \cellcolor{red!6!white} $49.1 \pm{1.3}$ & 
    \cellcolor{red!9!white} $48.6 \pm{0.3}$ \\ 

16 &
     $51.9 \pm{0.9}$ & 
     $52.1 \pm{0.4}$ & 
    \cellcolor{LimeGreen!5!white} $\underline{53.3 \pm{1.3}}$ & 
    \cellcolor{LimeGreen!6!white} $\bm{53.5 \pm{0.6}}$ & 
    \cellcolor{red!20!white} $46.7 \pm{0.4}$ & 
    \cellcolor{red!4!white} $50.9 \pm{1.5}$ & 
    \cellcolor{red!6!white} $50.3 \pm{0.4}$ \\ 

18 &
     $53.2 \pm{0.3}$ & 
    \cellcolor{LimeGreen!3!white} $54.1 \pm{0.3}$ & 
    \cellcolor{LimeGreen!4!white} $\underline{54.3 \pm{1.0}}$ & 
    \cellcolor{LimeGreen!8!white} $\bm{55.3 \pm{0.4}}$ & 
    \cellcolor{red!21!white} $47.8 \pm{0.6}$ & 
    \cellcolor{red!3!white} $52.2 \pm{0.8}$ & 
    \cellcolor{red!7!white} $51.4 \pm{0.8}$ \\ 

20 &
     $55.1 \pm{0.4}$ & 
     $55.2 \pm{0.4}$ & 
    \cellcolor{LimeGreen!2!white} $\underline{55.9 \pm{0.4}}$ & 
    \cellcolor{LimeGreen!3!white} $\bm{56.1 \pm{0.5}}$ & 
    \cellcolor{red!23!white} $49.3 \pm{0.6}$ & 
    \cellcolor{red!5!white} $53.9 \pm{0.8}$ & 
    \cellcolor{red!12!white} $52.0 \pm{0.9}$ \\

    \midrule
    \multicolumn{8}{c}{Food101} \\
    \midrule

1 &
     $\bm{6.6 \pm{0.8}}$ & 
     $\bm{6.6 \pm{0.8}}$ & 
     $\bm{6.6 \pm{0.8}}$ & 
     $\bm{6.6 \pm{0.8}}$ & 
     $\underline{6.4 \pm{0.4}}$ & 
     $\bm{6.6 \pm{0.8}}$ & 
    \cellcolor{red!1!white} $6.1 \pm{0.4}$ \\ 

2 &
     $10.3 \pm{1.0}$ & 
    \cellcolor{LimeGreen!1!white} $10.6 \pm{1.1}$ & 
    \cellcolor{LimeGreen!2!white} $11.0 \pm{0.6}$ & 
     $10.5 \pm{0.6}$ & 
    \cellcolor{red!2!white} $9.8 \pm{0.3}$ & 
    \cellcolor{LimeGreen!8!white} $\bm{12.4 \pm{0.8}}$ & 
    \cellcolor{LimeGreen!6!white} $\underline{12.0 \pm{0.7}}$ \\ 

4 &
     $17.6 \pm{0.6}$ & 
     $17.8 \pm{0.8}$ & 
    \cellcolor{LimeGreen!1!white} $17.9 \pm{0.6}$ & 
    \cellcolor{LimeGreen!1!white} $17.9 \pm{0.8}$ & 
    \cellcolor{red!15!white} $13.8 \pm{0.5}$ & 
    \cellcolor{LimeGreen!3!white} $\underline{18.5 \pm{0.6}}$ & 
    \cellcolor{LimeGreen!3!white} $\bm{18.5 \pm{0.3}}$ \\ 

6 &
     $23.2 \pm{0.4}$ & 
     $23.0 \pm{0.8}$ & 
    \cellcolor{LimeGreen!1!white} $\bm{23.6 \pm{0.7}}$ & 
    \cellcolor{LimeGreen!1!white} $\underline{23.5 \pm{1.1}}$ & 
    \cellcolor{red!12!white} $20.1 \pm{0.3}$ & 
    \cellcolor{LimeGreen!1!white} $23.4 \pm{0.6}$ & 
     $23.0 \pm{0.2}$ \\ 

8 &
     $27.5 \pm{1.2}$ & 
    \cellcolor{red!1!white} $27.2 \pm{0.8}$ & 
     $\bm{27.6 \pm{0.4}}$ & 
     $\underline{27.6 \pm{1.0}}$ & 
    \cellcolor{red!14!white} $24.0 \pm{0.5}$ & 
    \cellcolor{red!2!white} $26.8 \pm{0.7}$ & 
    \cellcolor{red!3!white} $26.6 \pm{0.5}$ \\ 

10 &
     $\underline{31.0 \pm{0.6}}$ & 
    \cellcolor{red!2!white} $30.4 \pm{0.4}$ & 
     $30.9 \pm{1.3}$ & 
     $\bm{31.1 \pm{1.4}}$ & 
    \cellcolor{red!16!white} $27.0 \pm{1.1}$ & 
    \cellcolor{red!6!white} $29.3 \pm{0.9}$ & 
    \cellcolor{red!6!white} $29.4 \pm{0.5}$ \\ 

12 &
     $33.7 \pm{0.7}$ & 
    \cellcolor{red!2!white} $33.1 \pm{0.7}$ & 
    \cellcolor{LimeGreen!1!white} $\bm{34.1 \pm{1.0}}$ & 
    \cellcolor{LimeGreen!1!white} $\underline{34.0 \pm{0.7}}$ & 
    \cellcolor{red!20!white} $28.6 \pm{1.0}$ & 
    \cellcolor{red!6!white} $32.1 \pm{0.7}$ & 
    \cellcolor{red!7!white} $31.8 \pm{0.4}$ \\ 

14 &
     $36.3 \pm{0.5}$ & 
    \cellcolor{red!1!white} $36.0 \pm{0.6}$ & 
    \cellcolor{LimeGreen!2!white} $\bm{37.0 \pm{0.6}}$ & 
     $\underline{36.4 \pm{0.5}}$ & 
    \cellcolor{red!19!white} $31.4 \pm{0.8}$ & 
    \cellcolor{red!9!white} $33.9 \pm{1.1}$ & 
    \cellcolor{red!10!white} $33.7 \pm{0.2}$ \\ 

16 &
     $38.3 \pm{0.4}$ & 
     $38.2 \pm{0.6}$ & 
    \cellcolor{LimeGreen!2!white} $\bm{38.8 \pm{1.0}}$ & 
    \cellcolor{LimeGreen!1!white} $\underline{38.5 \pm{0.7}}$ & 
    \cellcolor{red!20!white} $33.1 \pm{0.8}$ & 
    \cellcolor{red!9!white} $35.9 \pm{0.6}$ & 
    \cellcolor{red!14!white} $34.7 \pm{1.0}$ \\ 

18 &
     $40.1 \pm{0.3}$ & 
    \cellcolor{red!2!white} $39.3 \pm{1.4}$ & 
    \cellcolor{LimeGreen!2!white} $\bm{40.7 \pm{0.5}}$ & 
    \cellcolor{LimeGreen!2!white} $\underline{40.7 \pm{0.5}}$ & 
    \cellcolor{red!20!white} $34.8 \pm{0.9}$ & 
    \cellcolor{red!9!white} $37.7 \pm{0.7}$ & 
    \cellcolor{red!12!white} $36.9 \pm{0.5}$ \\ 

20 &
     $41.5 \pm{0.3}$ & 
    \cellcolor{red!2!white} $40.9 \pm{0.9}$ & 
     $\underline{41.5 \pm{1.0}}$ & 
    \cellcolor{LimeGreen!2!white} $\bm{42.2 \pm{0.8}}$ & 
    \cellcolor{red!24!white} $35.3 \pm{0.8}$ & 
    \cellcolor{red!9!white} $39.2 \pm{0.9}$ & 
    \cellcolor{red!15!white} $37.6 \pm{0.7}$ \\ 

    \midrule
    \multicolumn{8}{c}{ImageNet-100} \\
    \midrule

1 &
     $20.4 \pm{1.1}$ & 
     $20.4 \pm{1.1}$ & 
     $20.4 \pm{1.1}$ & 
     $20.4 \pm{1.1}$ & 
    \cellcolor{LimeGreen!22!white} $\bm{25.9 \pm{1.1}}$ & 
     $20.4 \pm{1.1}$ & 
    \cellcolor{LimeGreen!6!white} $\underline{22.0 \pm{1.3}}$ \\ 

2 &
     $31.4 \pm{2.7}$ & 
    \cellcolor{red!2!white} $30.7 \pm{1.3}$ & 
    \cellcolor{red!6!white} $29.8 \pm{0.8}$ & 
    \cellcolor{red!5!white} $30.0 \pm{1.2}$ & 
    \cellcolor{LimeGreen!18!white} $36.0 \pm{0.8}$ & 
    \cellcolor{LimeGreen!20!white} $\underline{36.5 \pm{0.8}}$ & 
    \cellcolor{LimeGreen!21!white} $\bm{36.8 \pm{0.9}}$ \\ 

4 &
     $45.5 \pm{2.2}$ & 
    \cellcolor{red!1!white} $45.2 \pm{1.6}$ & 
     $45.7 \pm{1.1}$ & 
    \cellcolor{LimeGreen!4!white} $46.5 \pm{1.9}$ & 
    \cellcolor{LimeGreen!6!white} $\underline{47.1 \pm{1.9}}$ & 
    \cellcolor{LimeGreen!4!white} $46.6 \pm{1.3}$ & 
    \cellcolor{LimeGreen!19!white} $\bm{50.3 \pm{1.4}}$ \\ 

6 &
     $54.6 \pm{1.3}$ & 
    \cellcolor{LimeGreen!9!white} $56.9 \pm{2.0}$ & 
    \cellcolor{LimeGreen!2!white} $55.3 \pm{1.0}$ & 
    \cellcolor{LimeGreen!10!white} $\underline{57.3 \pm{1.4}}$ & 
    \cellcolor{LimeGreen!2!white} $55.2 \pm{0.7}$ & 
    \cellcolor{red!3!white} $53.8 \pm{2.7}$ & 
    \cellcolor{LimeGreen!15!white} $\bm{58.5 \pm{0.5}}$ \\ 

8 &
     $60.1 \pm{1.4}$ & 
    \cellcolor{LimeGreen!10!white} $62.7 \pm{0.8}$ & 
    \cellcolor{LimeGreen!4!white} $61.2 \pm{1.6}$ & 
    \cellcolor{LimeGreen!12!white} $\bm{63.3 \pm{0.8}}$ & 
    \cellcolor{red!5!white} $58.7 \pm{1.1}$ & 
    \cellcolor{red!5!white} $58.8 \pm{4.2}$ & 
    \cellcolor{LimeGreen!12!white} $\underline{63.1 \pm{0.4}}$ \\ 

10 &
     $64.1 \pm{0.9}$ & 
    \cellcolor{LimeGreen!10!white} $\underline{66.6 \pm{1.9}}$ & 
    \cellcolor{LimeGreen!9!white} $66.4 \pm{0.9}$ & 
    \cellcolor{LimeGreen!14!white} $\bm{67.8 \pm{1.4}}$ & 
    \cellcolor{red!9!white} $61.7 \pm{1.0}$ & 
    \cellcolor{red!3!white} $63.2 \pm{4.0}$ & 
    \cellcolor{LimeGreen!9!white} $66.4 \pm{0.9}$ \\ 

12 &
     $67.3 \pm{0.5}$ & 
    \cellcolor{LimeGreen!12!white} $\underline{70.4 \pm{1.1}}$ & 
    \cellcolor{LimeGreen!8!white} $69.4 \pm{0.8}$ & 
    \cellcolor{LimeGreen!12!white} $\bm{70.5 \pm{1.2}}$ & 
    \cellcolor{red!11!white} $64.3 \pm{1.0}$ & 
    \cellcolor{red!4!white} $66.2 \pm{2.8}$ & 
    \cellcolor{LimeGreen!6!white} $68.8 \pm{0.9}$ \\ 

14 &
     $69.2 \pm{0.4}$ & 
    \cellcolor{LimeGreen!12!white} $\underline{72.4 \pm{0.9}}$ & 
    \cellcolor{LimeGreen!10!white} $71.9 \pm{0.4}$ & 
    \cellcolor{LimeGreen!14!white} $\bm{72.8 \pm{1.1}}$ & 
    \cellcolor{red!13!white} $65.9 \pm{0.5}$ & 
     $69.0 \pm{1.8}$ & 
    \cellcolor{LimeGreen!4!white} $70.4 \pm{0.5}$ \\ 

16 &
     $71.3 \pm{0.6}$ & 
    \cellcolor{LimeGreen!10!white} $\underline{73.9 \pm{1.0}}$ & 
    \cellcolor{LimeGreen!6!white} $72.8 \pm{1.2}$ & 
    \cellcolor{LimeGreen!12!white} $\bm{74.4 \pm{0.8}}$ & 
    \cellcolor{red!16!white} $67.1 \pm{0.5}$ & 
    \cellcolor{red!1!white} $70.9 \pm{1.6}$ & 
    \cellcolor{LimeGreen!2!white} $72.0 \pm{0.4}$ \\ 

18 &
     $72.5 \pm{0.6}$ & 
    \cellcolor{LimeGreen!10!white} $\underline{75.1 \pm{0.8}}$ & 
    \cellcolor{LimeGreen!7!white} $74.5 \pm{0.7}$ & 
    \cellcolor{LimeGreen!13!white} $\bm{75.8 \pm{1.3}}$ & 
    \cellcolor{red!16!white} $68.3 \pm{0.4}$ & 
    \cellcolor{LimeGreen!2!white} $73.0 \pm{1.3}$ & 
    \cellcolor{LimeGreen!2!white} $73.2 \pm{0.9}$ \\ 

20 &
     $73.5 \pm{0.9}$ & 
    \cellcolor{LimeGreen!11!white} $\underline{76.4 \pm{0.5}}$ & 
    \cellcolor{LimeGreen!9!white} $75.7 \pm{0.6}$ & 
    \cellcolor{LimeGreen!14!white} $\bm{77.2 \pm{0.9}}$ & 
    \cellcolor{red!19!white} $68.7 \pm{0.8}$ & 
    \cellcolor{LimeGreen!3!white} $74.4 \pm{1.2}$ & 
    \cellcolor{LimeGreen!3!white} $74.3 \pm{0.6}$ \\ 

    \midrule
    \multicolumn{8}{c}{DomainNet-Real} \\
    \midrule

1 &
     $16.0 \pm{0.7}$ & 
     $16.0 \pm{0.7}$ & 
     $16.0 \pm{0.7}$ & 
     $16.0 \pm{0.7}$ & 
    \cellcolor{LimeGreen!27!white} $\bm{22.9 \pm{0.6}}$ & 
     $16.0 \pm{0.7}$ & 
    \cellcolor{LimeGreen!7!white} $\underline{17.9 \pm{0.6}}$ \\ 

2 &
     $25.2 \pm{0.6}$ & 
    \cellcolor{red!2!white} $24.5 \pm{0.9}$ & 
    \cellcolor{LimeGreen!2!white} $25.8 \pm{0.7}$ & 
    \cellcolor{LimeGreen!2!white} $25.7 \pm{0.7}$ & 
    \cellcolor{LimeGreen!16!white} $29.2 \pm{0.7}$ & 
    \cellcolor{LimeGreen!19!white} $\underline{30.1 \pm{0.4}}$ & 
    \cellcolor{LimeGreen!20!white} $\bm{30.4 \pm{0.4}}$ \\ 

4 &
     $37.7 \pm{1.1}$ & 
    \cellcolor{LimeGreen!2!white} $38.3 \pm{0.6}$ & 
    \cellcolor{LimeGreen!3!white} $38.5 \pm{0.6}$ & 
    \cellcolor{LimeGreen!4!white} $38.9 \pm{0.9}$ & 
    \cellcolor{red!7!white} $35.8 \pm{0.5}$ & 
    \cellcolor{LimeGreen!6!white} $\underline{39.2 \pm{0.5}}$ & 
    \cellcolor{LimeGreen!17!white} $\bm{41.9 \pm{0.2}}$ \\ 

6 &
     $45.6 \pm{0.6}$ & 
    \cellcolor{LimeGreen!3!white} $46.4 \pm{0.7}$ & 
    \cellcolor{LimeGreen!3!white} $46.4 \pm{1.0}$ & 
    \cellcolor{LimeGreen!5!white} $\underline{47.0 \pm{0.3}}$ & 
    \cellcolor{red!20!white} $40.3 \pm{0.4}$ & 
    \cellcolor{red!6!white} $43.9 \pm{0.5}$ & 
    \cellcolor{LimeGreen!8!white} $\bm{47.7 \pm{0.4}}$ \\ 

8 &
     $50.4 \pm{0.6}$ & 
    \cellcolor{LimeGreen!4!white} $51.5 \pm{0.3}$ & 
    \cellcolor{LimeGreen!4!white} $\underline{51.5 \pm{0.8}}$ & 
    \cellcolor{LimeGreen!7!white} $\bm{52.2 \pm{0.9}}$ & 
    \cellcolor{red!26!white} $43.8 \pm{0.3}$ & 
    \cellcolor{red!14!white} $46.9 \pm{0.5}$ & 
    \cellcolor{LimeGreen!4!white} $51.5 \pm{0.3}$ \\ 

10 &
     $54.1 \pm{0.4}$ & 
    \cellcolor{LimeGreen!4!white} $\underline{55.2 \pm{0.3}}$ & 
    \cellcolor{LimeGreen!4!white} $55.2 \pm{0.5}$ & 
    \cellcolor{LimeGreen!6!white} $\bm{55.8 \pm{0.4}}$ & 
    \cellcolor{red!30!white} $46.5 \pm{0.6}$ & 
    \cellcolor{red!18!white} $49.4 \pm{0.2}$ & 
    \cellcolor{LimeGreen!1!white} $54.3 \pm{0.3}$ \\ 

12 &
     $\bm{47.8 \pm{0.7}}$ & 
    \cellcolor{red!18!white} $43.2 \pm{0.9}$ & 
    \cellcolor{red!7!white} $\underline{45.9 \pm{1.8}}$ & 
    \cellcolor{red!15!white} $43.9 \pm{1.2}$ & 
    \cellcolor{red!26!white} $41.1 \pm{0.8}$ & 
    \cellcolor{red!15!white} $43.9 \pm{1.7}$ & 
    \cellcolor{red!21!white} $42.4 \pm{1.0}$ \\ 

14 &
     $57.6 \pm{0.2}$ & 
    \cellcolor{LimeGreen!3!white} $58.5 \pm{0.4}$ & 
    \cellcolor{LimeGreen!4!white} $\underline{58.8 \pm{0.1}}$ & 
    \cellcolor{LimeGreen!6!white} $\bm{59.3 \pm{0.1}}$ & 
    \cellcolor{red!29!white} $50.2 \pm{0.6}$ & 
    \cellcolor{red!20!white} $52.5 \pm{0.1}$ & 
    \cellcolor{red!3!white} $56.8 \pm{0.3}$ \\ 

16 &
     $59.8 \pm{0.2}$ & 
    \cellcolor{LimeGreen!3!white} $60.6 \pm{0.3}$ & 
    \cellcolor{LimeGreen!3!white} $\underline{60.7 \pm{0.3}}$ & 
    \cellcolor{LimeGreen!6!white} $\bm{61.4 \pm{0.3}}$ & 
    \cellcolor{red!30!white} $52.0 \pm{0.4}$ & 
    \cellcolor{red!23!white} $54.0 \pm{0.3}$ & 
    \cellcolor{red!4!white} $58.7 \pm{0.3}$ \\ 

18 &
     $61.2 \pm{0.1}$ & 
    \cellcolor{LimeGreen!3!white} $62.0 \pm{0.2}$ & 
    \cellcolor{LimeGreen!4!white} $\underline{62.2 \pm{0.3}}$ & 
    \cellcolor{LimeGreen!6!white} $\bm{62.8 \pm{0.2}}$ & 
    \cellcolor{red!29!white} $53.9 \pm{0.3}$ & 
    \cellcolor{red!23!white} $55.4 \pm{0.2}$ & 
    \cellcolor{red!4!white} $60.2 \pm{0.1}$ \\ 

20 &
     $62.3 \pm{0.2}$ & 
    \cellcolor{LimeGreen!4!white} $\underline{63.4 \pm{0.2}}$ & 
    \cellcolor{LimeGreen!4!white} $63.4 \pm{0.3}$ & 
    \cellcolor{LimeGreen!7!white} $\bm{64.1 \pm{0.2}}$ & 
    \cellcolor{red!28!white} $55.1 \pm{0.3}$ & 
    \cellcolor{red!22!white} $56.7 \pm{0.2}$ & 
    \cellcolor{red!4!white} $61.2 \pm{0.2}$ \\ 

    \bottomrule
    \end{tabular}
    \end{adjustbox}
    \label{tab:mae_result}
\end{table}

\end{document}

%% file: math_commands.tex
\usepackage{amsmath,amsfonts,bm}

\def\eqref#1{equation~\ref{#1}}

\def\1{\bm{1}}

\DeclareMathAlphabet{\mathsfit}{\encodingdefault}{\sfdefault}{m}{sl}
\SetMathAlphabet{\mathsfit}{bold}{\encodingdefault}{\sfdefault}{bx}{n}

\DeclareMathOperator*{\argmin}{arg\,min}